\documentclass[%
 reprint,
 amsmath,amssymb,
 aps,
]{revtex4-2}

\usepackage{graphicx}
\usepackage{dcolumn}
\usepackage{bm}
\usepackage{booktabs} 
\usepackage{amsfonts}
\usepackage{multirow}
\usepackage{amsmath}
\usepackage{caption}
\usepackage{subcaption}
\usepackage{hyperref}
\usepackage{natbib}

\begin{document}

\preprint{APS/123-QED}

\title{Port-Hamiltonian Neural Networks \\ for Learning Explicit Time-Dependent Dynamical Systems}

\author{Shaan A. Desai}
 \altaffiliation[Also at ]{John A. Paulson School of Engineering and Applied Sciences, Harvard University}
 \email{shaan@robots.ox.ac.uk}
\author{Stephen J. Roberts}%
 
\affiliation{%
 Machine Learning Research Group, University of Oxford \\
 Eagle House, Oxford OX26ED, United Kingdom
}%


\author{Marios Mattheakis}
\author{David Sondak}
\author{Pavlos Protopapas}
\affiliation{
John A. Paulson School of Engineering and Applied Sciences, Harvard University \\
Cambridge, Massachusetts 02138, United States
}%

\date{\today}

\begin{abstract}
Accurately learning the temporal behavior of dynamical systems requires models with well-chosen learning biases. Recent innovations embed the Hamiltonian and Lagrangian formalisms into neural networks and demonstrate a significant improvement over other approaches in predicting trajectories of physical systems. These methods generally tackle autonomous systems that depend implicitly on time or systems for which a control signal is known apriori. Despite this success, many real world dynamical systems are non-autonomous, driven by time-dependent forces and experience energy dissipation. In this study, we address the challenge of learning from such non-autonomous systems by embedding the port-Hamiltonian formalism into neural networks, a versatile framework that can capture energy dissipation and time-dependent control forces. We show that the proposed \emph{port-Hamiltonian neural network} can efficiently learn the dynamics of nonlinear physical systems of practical interest and accurately recover the underlying stationary Hamiltonian, time-dependent force, and dissipative coefficient. A promising outcome of our network is its ability to learn and predict chaotic systems such as the Duffing equation, for which the trajectories are typically hard to learn.
\end{abstract}

\maketitle


\section{\label{sec:level1}Introduction}

Neural networks (NNs), as universal function approximators \cite{hornik_multilayer_1989}, have shown resounding success across a host of domains including image segmentation \cite{he_mask_2018}, machine translation \cite{devlin_bert_2019}, and material property predictions \cite{toussaint_differentiable_2018,yao_tensormol-01_2018}. 
However, their performance in learning and generalizing the long-term behaviour of dynamic systems governed by known physical laws from state data has often been limited \cite{greydanus_hamiltonian_2019,pukrittayakamee_simultaneous_2009}. New research in \textit{scientific machine learning}, a field that tackles scientific problems with domain-specific machine learning methods, is paving a way to address this challenge. Concretely, it has been shown that physically-informed learning biases embedded in networks, such as Hamiltonian mechanics \cite{mattheakis_hamiltonian_2020, greydanus_hamiltonian_2019}, Lagrangians \cite{cranmer_lagrangian_2020, lutter_deep_2019}, Ordinary Differential Equations (ODEs) \cite{chen_neural_2018}, physics-informed networks \cite{raissi_physics_2017, HNN_PRE2020}, generative networks \cite{HNN_iclr2020},  and Graph Neural Networks \cite{battaglia_interaction_2016,sanchez-gonzalez_hamiltonian_2019} can significantly improve learning and generalization over vanilla neural networks in complex physical domains. 
The performance uplift arises primarily because learning biases are able to constrain networks to learn physically meaningful representations from data that are crucial to generalization.

Despite extensive research on learning biases, there is yet no method that accounts for non-autonomous systems i.e. systems with explicit time dependence. Non-autonomous dynamics feature prominently in settings with externally driven or controlled time-dependent forces as well as in systems with energy dissipation, for example, interacting materials, forced oscillators and charge-discharge cycles. Defining a network to accurately learn and predict the dynamics of such systems from position and momentum data is therefore of critical practical interest. We address this challenge by embedding the port-Hamiltonian formalism \cite{sanz-sole_port-hamiltonian_2007,ORTEGA2002585,acosta_interconnection_2005,zheng_time-varying_2018,cherifi_overview_2020} into neural networks. We show that the structure of this formulation can be used to uncover the underlying Hamiltonian, force and damping terms given position and momentum data and as such, can be used to accurately predict the long-range trajectories of many forced/damped systems. We extensively benchmark our network on a range of tasks including a simple mass-spring system with damping and external force, a Duffing system in both the non-chaotic and chaotic regimes, and a relativistic Duffing system. Our proposed network consistently outperforms other approaches while accurately recovering both the driving force and the damping coefficient. Furthermore, using minimal data, we show that our network can visually recover the Poincar\'e section of the Duffing system in a chaotic regime, emphasizing how our network can be used to identify and understand chaotic trajectories.

\section{\label{sec:level1} Background}

\subsection{Hamiltonian Neural Networks}
Recently, the authors of \cite{greydanus_hamiltonian_2019} demonstrated that the dynamics of an energy conserving autonomous system can be accurately learned by guiding a neural network to predict a Hamiltonian - an important representation of a dynamical system.
Considering a dynamical system of $M$ objects, the Hamiltonian $\mathcal{H}$ is a scalar function of a position vector $\mathbf{q}(t) = (q_1(t),q_2(t),....,q_M(t))$ and momentum vector $\mathbf{p}(t) = (p_1(t),p_2(t),....,p_M(t))$ that obeys Hamilton's equations,
\begin{align}
\dot{\mathbf{q}}= \frac{\partial \mathcal{H}}{\partial \mathbf{p}}, \qquad 
\dot{\mathbf{p}}= -\frac{\partial \mathcal{H}}{\partial \mathbf{q}},
\label{eqn.hamiltonian}
\end{align}
where $\dot{\mathbf{q}}=\frac{d\mathbf{q}}{dt}$ and  $\dot{\mathbf{p}}=\frac{d\mathbf{p}}{dt}$. 

Using the Hamiltonian formalism, \cite{greydanus_hamiltonian_2019} showed that a NN with parameters $\theta$ can be used to learn a Hamiltonian $\mathcal{H}_{\theta}(\mathbf{q},\mathbf{p})$ given $\mathbf{q}$ and $\mathbf{p}$ as inputs to the network. The time derivatives are recovered from Eq.~\eqref{eqn.hamiltonian} by differentiating $\mathcal{H}_{\theta}$ with respect to its inputs using automatic differentiation. The resulting system has the form,
\begin{align}
  \dot{\mathbf{x}} = s(\mathbf{x}) \label{eq:system},
\end{align}
where $\mathbf{x} = \left(\mathbf{q}, \mathbf{p}\right)$ and $s$ is determined by Eq.~\eqref{eqn.hamiltonian} after differentiating the trained NN. Eq.~\eqref{eq:system} can be discretized using a time-integrator to determine the trajectory of an initial state. Moreover, since $s$ is the symplectic gradient of the Hamiltonian, energy conservation is embedded into the method by construction. Given these advantages, Hamiltonian Neural Networks (HNNs) outperform traditional approaches that directly predict the state time derivatives from the input state. However, this formulation does not readily generalize to damped or forced time-varying systems.

\subsection{Port-Hamiltonian framework}

The port-Hamiltonian \cite{sanz-sole_port-hamiltonian_2007,ORTEGA2002585,acosta_interconnection_2005,zheng_time-varying_2018,cherifi_overview_2020} is a well studied formalism that generalizes Hamilton's equations to incorporate energy dissipation and an external control input to a dynamical system. Hamilton's equations in the  port-Hamiltonian framework are represented as:
\begin{align}
\resizebox{0.9\linewidth}{!}{$%
\begin{bmatrix}
  \dot{\mathbf{q}} \\
  \dot{\mathbf{p}}
\end{bmatrix}
=
\left(
\begin{bmatrix}
  \mathbf{0}  & \mathbf{I} \\
  -\mathbf{I} & \mathbf{0}
\end{bmatrix} +
\mathbf{D}\left(\mathbf{q}\right)
\right)
\begin{bmatrix}
  \dfrac{\partial \mathcal{H}}{\partial \mathbf{q}} \\[1.0em]
  \dfrac{\partial \mathcal{H}}{\partial \mathbf{p}}
\end{bmatrix}
+
\begin{bmatrix}
  \mathbf{0} \\
  \mathbf{G}\left(\mathbf{q}\right)
\end{bmatrix}
\mathbf{u},
$%
}%
\label{eqn.pham}
\end{align}

where $\mathbf{D}(\mathbf{q})$ is a damping matrix, $\mathbf{u}$ is a temporal control input, $\mathbf{G}(\mathbf{q})$ is a non-linear scaling of the position vector, $\mathbf{I}$ is the identity matrix, and $\mathbf{0}$ the zero matrix. The damping matrix is semi-positive definite. This general formalism readily reduces to the standard Hamiltonian system when $\mathbf{D}=\mathbf{0}$ and $\mathbf{u}=\mathbf{0}$. The port Hamiltonian has been used in control applications where explicit knowledge of the control term $\mathbf{u}$ is known and was recently shown to reveal promising results in NNs \cite{zhong_dissipative_2020}. Note that in \cite{zhong_dissipative_2020}, the vector $[\mathbf{q},\mathbf{p},\mathbf{u}]$ is provided as input to the network. However, in many applications the control force $\mathbf{u}$ is unknown and it is therefore of interest to uncover the underlying forcing term from the data of the state vector, where no explicit knowledge of the control input and damping term is available.

\subsection{Related Work}

While Hamiltonian mechanics presents one way to address learning dynamical systems, numerous recent methods highlight how incorporating other physically-informed inductive biases into neural networks can improve learning.

Functional priors, for example, embed the full functional form of an equation into the NN. Physics-informed neural networks (PINNs) \cite{raissi_physics_2017,raissi_physics-informed_2019} and Hamiltonian networks \cite{mattheakis_hamiltonian_2020} are two such approaches that look at directly embedding the equations of motion into the loss function. While PINNs are data-driven approaches that rely on Autograd \cite{maclaurin_autograd_nodate} to compute partial-derivatives of a hidden state, Hamiltonian networks are data-independent approaches that pre-specify the full functional form of a system-specific Hamiltonian in the loss function.

Many recent methods have also sought to embed integrators into the training process. Indeed doing so induces an effectively continuous depth neural network able to perform large time-step predictions. NeuralODE \cite{chen_neural_2018} presents one way to tackle back-propagating through this continuous depth network more efficiently.

More recent work has looked at generalizing this approach to different and more complex data structures and topologies that standard NeuralODEs cannot represent \cite{dupont_augmented_2019}. Other work, such as \citet{zhu_deep_2020}, theoretically shows the importance of using symplectic integrators over Runge-Kutta methods to evolve Hamiltonian systems in NNs.

In \cite{battaglia_interaction_2016}, the authors detailed how a Graph Neural Network, designed to capture the relational structure of systems, can be used to learn dynamics of interacting particles. This work has been exploited in numerous advances \cite{sanchez-gonzalez_graph_2018,sanchez-gonzalez_learning_2020,cranmer_lagrangian_2020} and emphasizes how a relational inductive bias can significantly improve learning.

In  \cite{cranmer_lagrangian_2020} and \cite{greydanus_hamiltonian_2019} it has been shown that by learning the Hamiltonian or the Lagrangian of a system, it is possible to accurately predict the temporal dynamics and conserve energy. The work of \cite{lutter_deep_2019} also showed that by exploiting the Euler-Lagrange equation, it is possible to predict a controlled double pendulum - a system pertinent for controlled robots. 
%
%
A recent advance shows that Hamiltonian and Lagrangian NNs can be drastically improved if they are optimized over Cartesian coordinates with holonomic constraints \cite{finzi_generalizing_2020}. 

Despite the significant breakthroughs, there is no existing method that investigates explicit time-dependence and damping in dynamical systems, two elements that are often found in real world problems. As such, we outline a novel technique to address this challenge.

\section{Method}
\subsection{Theory}
In this section we introduce port-Hamiltonian Neural Networks (pHNNs). We begin by illustrating how a NN takes the form of the port-Hamiltonian formulation of Eq. (\ref{eqn.pham}) with two modifications. 
First, our approach exploits the fact that many damped systems consist of a non-zero, state-independent damping term in the lower right quadrant, so we replace the damping matrix $\mathbf{D}(\mathbf{q})$ with a state independent matrix for which only the lower right term is non-zero and represented by $\mathbf{N}$.
Secondly, in order to generalize to time dependent forcing, we replace $\mathbf{G}(\mathbf{q})\mathbf{u}$ with the force field $\mathbf{F}(t)$. The resulting representation, 
\begin{equation}
\resizebox{0.9\linewidth}{!}{$%
\begin{bmatrix}
\dot{\mathbf{q}} \\
\dot{\mathbf{p}}
\end{bmatrix}
=
\Bigg(\begin{bmatrix}
\mathbf{0} & \mathbf{I} \\
-\mathbf{I} & \mathbf{0}
\end{bmatrix} +
\begin{bmatrix}
\mathbf{0} & \mathbf{0} \\
\mathbf{0} & \mathbf{N}
\end{bmatrix}
 \Bigg)
 \begin{bmatrix}
\dfrac{\partial \mathcal{H}}{\partial \mathbf{q}} \\[1.0em]
\dfrac{\partial \mathcal{H}}{\partial \mathbf{p}}
\end{bmatrix}
+
\begin{bmatrix}
\mathbf{0} \\
\mathbf{F}(t)
\end{bmatrix},
$%
}%
\label{eqn.pham1}
\end{equation}
is general enough to handle many well-known forced systems but also specific enough to tackle learning in physical domains of practical importance such as  Duffing equation.
%
\begin{figure}[h]
\centering
\includegraphics[width=0.5\textwidth]{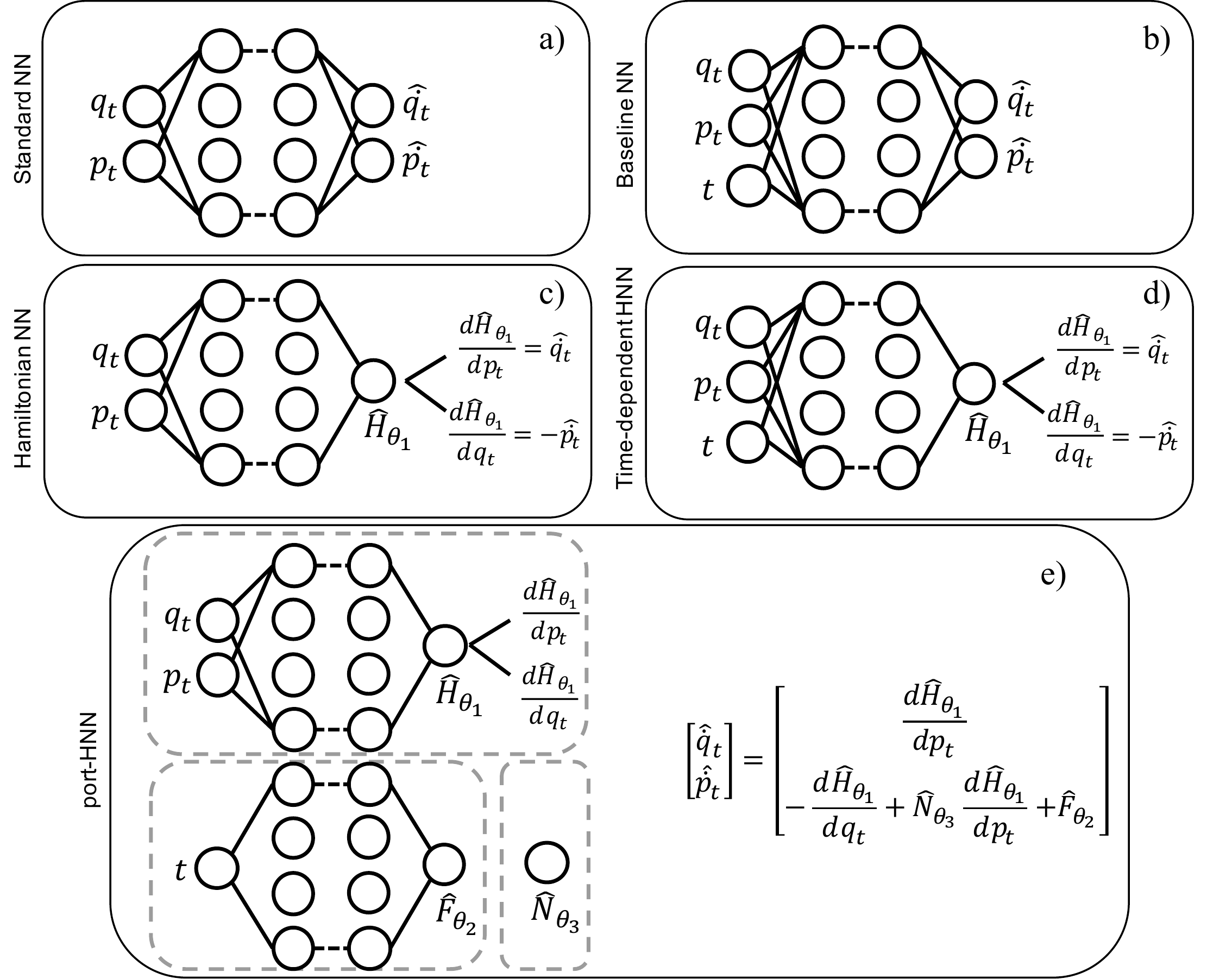}
\caption{Architectures used in this study to learn dynamical systems.  The naive extension of a standard feed forward NN (outlined in (a)) to incorporate time as an additional variable is shown in (b) and considered as the baseline network. The standard HNN in (c) is extended to receive time as an input and demonstrated by (d). 
 Our innovation is presented in (e), which exploits port-Hamiltonians and explicitly learns the force $F_{\theta_2}$, the damping term $N_{\theta_3}$ and the Hamiltonian $\mathcal{H}_{\theta_1}$ to predict the state-time derivatives. 
}
\label{fig.architecture}
\end{figure}

The architecture for the proposed pHNN model and  a summary of comparable existing approaches are shown in Fig. \ref{fig.architecture}. A standard feed forward NN in Fig.~\ref{fig.architecture}(a) and an HNN in Fig.~\ref{fig.architecture}(c) \cite{greydanus_hamiltonian_2019} take $q$ and $p$ as inputs and are trained to yield the time derivatives of the input, with the HNN learning an intermediate Hamiltonian and employing backpropagation to compute the final output. A natural way to extend these architectures for time-varying non-autonomous systems is to include time as an additional input. This gives rise to the baseline network (Baseline NN) represented by Fig.~\ref{fig.architecture}(b) and a time-dependent HNN (TDHNN) shown by Fig. \ref{fig.architecture}(d).

Although the Baseline NN and TDHNN incorporate time, they do not provide information for the underlying dynamics of the system. On the other hand, the pHNN is able to extract and provide information about the stationary Hamiltonian, the driving force, and the damping term. 
Moreover, the pHNN consistently outperforms all the network architectures shown in Figs. \ref{fig.architecture}(a-d) across the applications investigated in this study. 

\subsection{Network optimization}
The training of the pHNN consists of feeding the inputs $[\mathbf{q},\mathbf{p},t]$ into the model. 
The first component, $\mathcal{H}_{\theta_1}$, consists of three hidden layers designed to predict a stationary $\hat{\mathcal{H}_{\theta_1}}$ from $[\mathbf{q},\mathbf{p}]$ input data. The second component, $\hat{\mathbf{N}}_{\theta_3}$, consists of a single weight parameter (i.e. $\theta_3$ is a single node) designed to learn the damping. The third neural-network unit $\hat{\mathbf{F}}_{\theta_2}$ solely depends on $t$ and consists of three hidden layers designed to predict a time-varying force. The output of each component is transformed through Eq. (\ref{eqn.pham1}) to obtain predictions of the state time derivatives $[\hat{\dot{\mathbf{q}}},\hat{\dot{\mathbf{p}}}]$.
Using these predicted quantities we construct the loss function for optimizing the pHNN:
\begin{align}
\label{eqn.loss}
   \mathcal{L} = || \hat{\dot{\mathbf{q}}}_t - \dot{\mathbf{q}}_t ||_2^2 &+
 || \hat{\dot{\mathbf{p}}}_t -\dot{\mathbf{p}}_t ||_2^2 \\ \nonumber
 &+ \lambda_F|| \hat{\mathbf{F}}_{\theta_2}||_1 + \lambda_N||\hat{\mathbf{N}}_{\theta_3}||_1,%
\end{align}
where  $[\hat{\dot{\mathbf{q}}},\hat{\dot{\mathbf{p}}}]$ are known ground truth data. The first two components of the left hand side of Eq. (\ref{eqn.loss}) minimize the difference between the predicted and ground truth state time derivatives with a squared error loss. The last two components in Eq. (\ref{eqn.loss}) are the forcing and damping terms that are added to the loss function with an $L_1$ penalty when using pHNN. Using an $L_{1}$ penalty on these terms encourages the network to learn simpler models. We empirically found that this technique prevents the pHNN from learning spurious force and damping terms in unforced and undamped systems compared to an $L_2$ penalty. The regularization parameters $\lambda_{F}$ and $\lambda_{N}$ were determined via grid search (see Appendix A). 
We use 200 nodes per hidden layer and find that most activation functions, including $\tanh(\cdot)$, $\sin(\cdot)$ and $\cos(\cdot)$ yield comparable results.

We generate our data using an RK4 integrator given some initial conditions for each system. We use a small $\Delta t$ to evaluate the integral and ensure ground truth data is generated with rtol $=10^{-10}$. The gradients of the state at each integration step are computed using the underlying differential equation. Details about the sampling of the initial conditions are described independently for each system in section~\ref{sec:results}.

We note that in some settings it might be hard to obtain the ground truth state time-derivatives $[\mathbf{\dot{q}},\mathbf{\dot{p}}]$ for training. A natural way to address this problem is to embed an integrator into the training, similar to NeuralODE \cite{chen_neural_2018}. As such, we also run our study with an embedded RK-4 integrator. Our method is still the most performant when all the methods incorporate an embedded RK-4 integrator (see Appendix) and the loss function is $([\hat{q},\hat{p}]_{t+1} - [q,p]_{t+1})^2$ . In other words, our system can learn from either state time derivative data or directly from state data given an embedded integrator. We were motivated to use gradient data since HNN is trained in this way and we wanted a fair comparison. 

\subsection{Testing}
Once trained, each of the networks in Fig. \ref{fig.architecture} can approximate $[\dot{\mathbf{q}},\dot{\mathbf{p}}]$. As such, these networks can be used in a time integrator to evolve initial conditions in the test set. We refer to the integration for $t\in\left(0,T_{\max}\right)$ as the \textit{state rollout}. We measure the performance of the network by comparing the predicted state rollout with the ground truth. Specifically, we assess the networks performance by computing the mean squared error (MSE) of the predicted state variables and predicted energy (Hamiltonian) across the integration time,
\begin{align}
  \mathrm{MSE}_{\mathrm{state}} &= \frac{1}{N}\sum_{i=1}^N (\mathbf{q}_i-\hat{\mathbf{q}}_i)^2 + \frac{1}{N}\sum_{i=1}^N (\mathbf{p}_i - \hat{\mathbf{p}}_i)^2 \label{eq:stateMSE} \\
  \mathrm{MSE}_{\mathrm{energy}} &= \frac{1}{N} \sum_{i=1}^N \left(\mathcal{H}(\mathbf{q}_i,\mathbf{p}_i)-\mathcal{H}(\hat{\mathbf{q}}_i,\hat{\mathbf{p}}_i)\right)^2 \label{eq:energyMSE},
\end{align}
where $N = T_{\max}/\Delta t $, with $\Delta t$ the time step size. These terms are computed for multiple initial conditions during inference and averaged across them.

A  guide to the training  process  is outlined in Appendix B.

\section{\label{sec:results}Results}

We benchmark the performance of pHNNs against the other networks shown in Fig.~ \ref{fig.architecture}. We evaluate the methods over datasets that cover simple time-independent systems to complex chaotic damped and driven dynamical systems. The results are presented in order of increasing complexity from a model perspective.

\subsection{Simple Mass-Spring System}

We begin our analysis with a simple mass-spring system (harmonic oscillator), obeying Hooke's Law from classical physics with no force or damping. The  Hamiltonian that describes such a system reads:
\begin{equation}
\mathcal{H} = \frac{1}{2}k q^2 + \frac{p^2}{2m}, 
\end{equation}
where $k$ is the spring constant and $m$ denotes the mass. In this one-dimensional system the position and momentum are scalar functions of time.

\textbf{Training:} Without loss of generalization, we set $k=m=1$  for our experiments. We randomly sample 25 initial training conditions $[q_0,p_0]$ that satisfy $q_0^2+p_0^2 = r_0^2$ where $1 \leq r_0 \leq 4.5$ which corresponds to sampling initial conditions with energies in the range $[1, 4.5]$. We evolve  each initial state using a RK4 integrator with $\Delta t =0.05$ and $T_{\max} = 3.05$. 

\textbf{Testing:} We evaluate the performance of the NNs by sampling 25 random initial conditions in the same way as training. We investigate the simple harmonic oscillator system and show that learning a separate, regularized forcing term results in  better state and energy predictions in comparison to TDHNN and the Baseline NN. Learning a separate forcing term and regularizing pHNN keeps the time component independent of the Hamiltonian and therefore allows us to closely match the performance of the standard time-implicit HNN. 

\begin{figure}[h!]
\centering
\captionsetup{justification=centering}
	\begin{subfigure}[b]{0.4\textwidth}
		\centering
		\includegraphics[width=\textwidth, trim={0 0 0 12cm},clip]{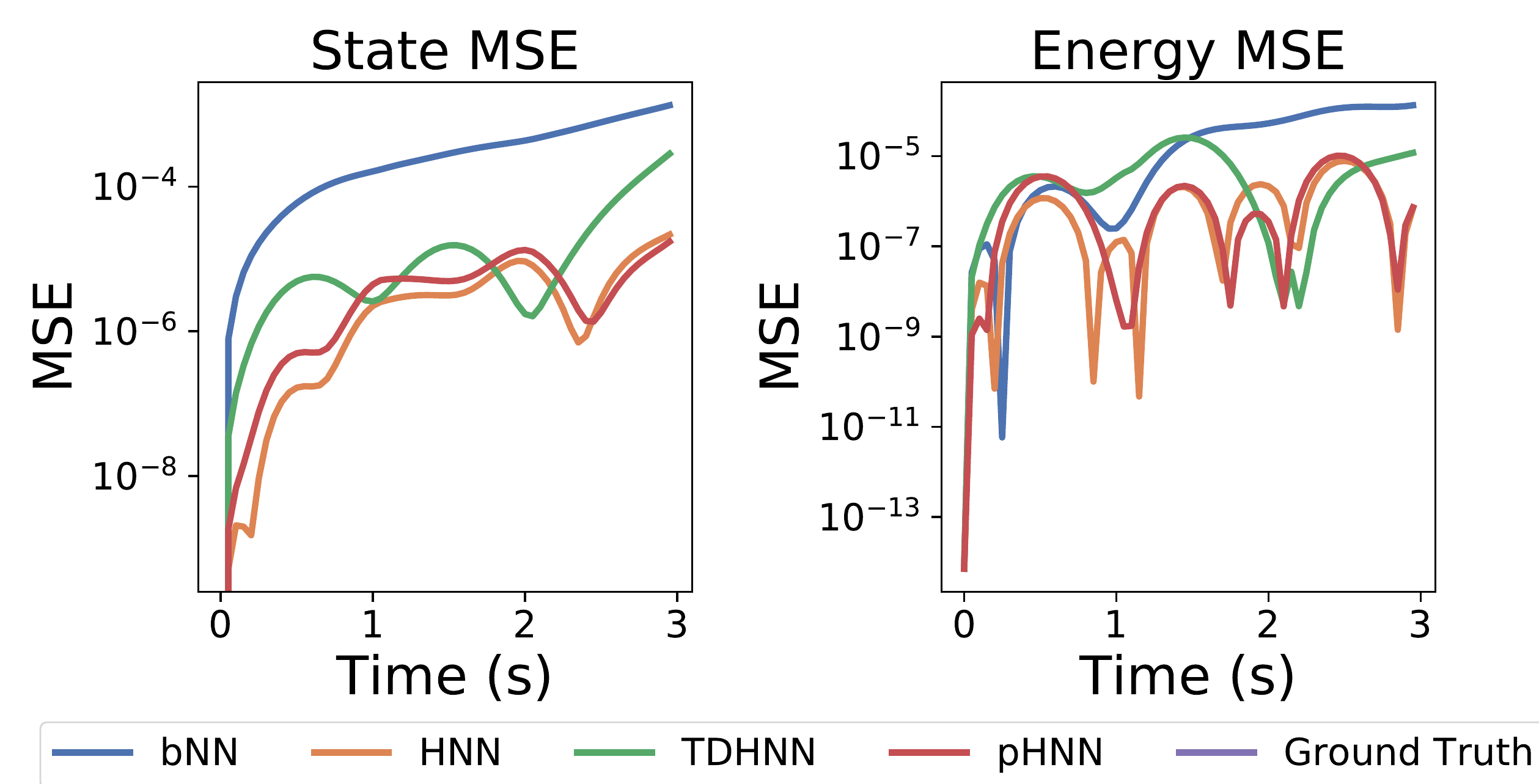}
		\caption{State and energy MSE as a function of time of an initial condition in the test set.}
	\end{subfigure}
	\begin{subfigure}[b]{0.48\textwidth}
		\centering
		\includegraphics[width=\textwidth]{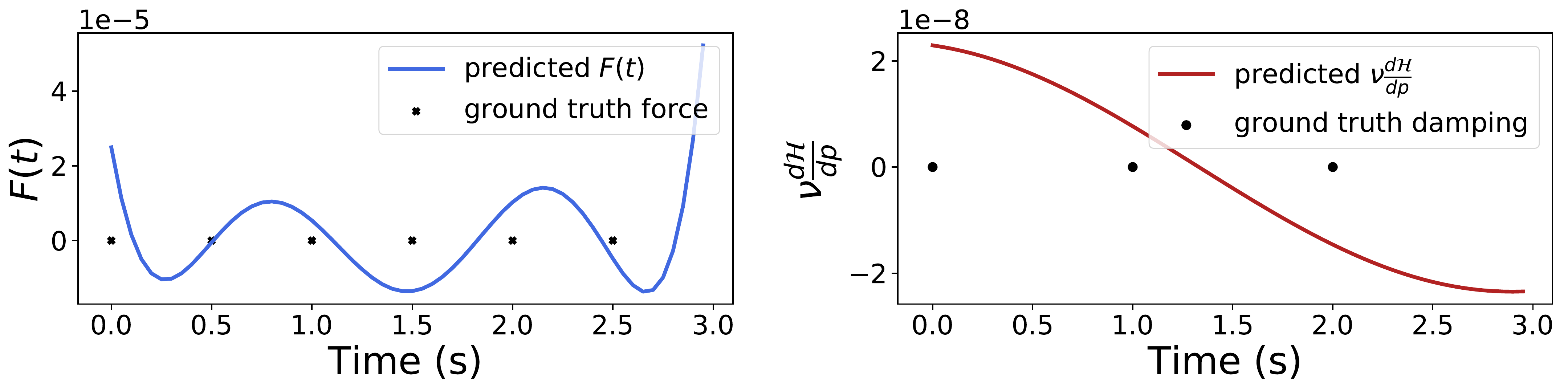}
		\caption{Learnt force and damping terms by pHNN}
	\end{subfigure}
	\begin{subfigure}[b]{0.48\textwidth}
	    \centering
		\includegraphics[width=\textwidth]{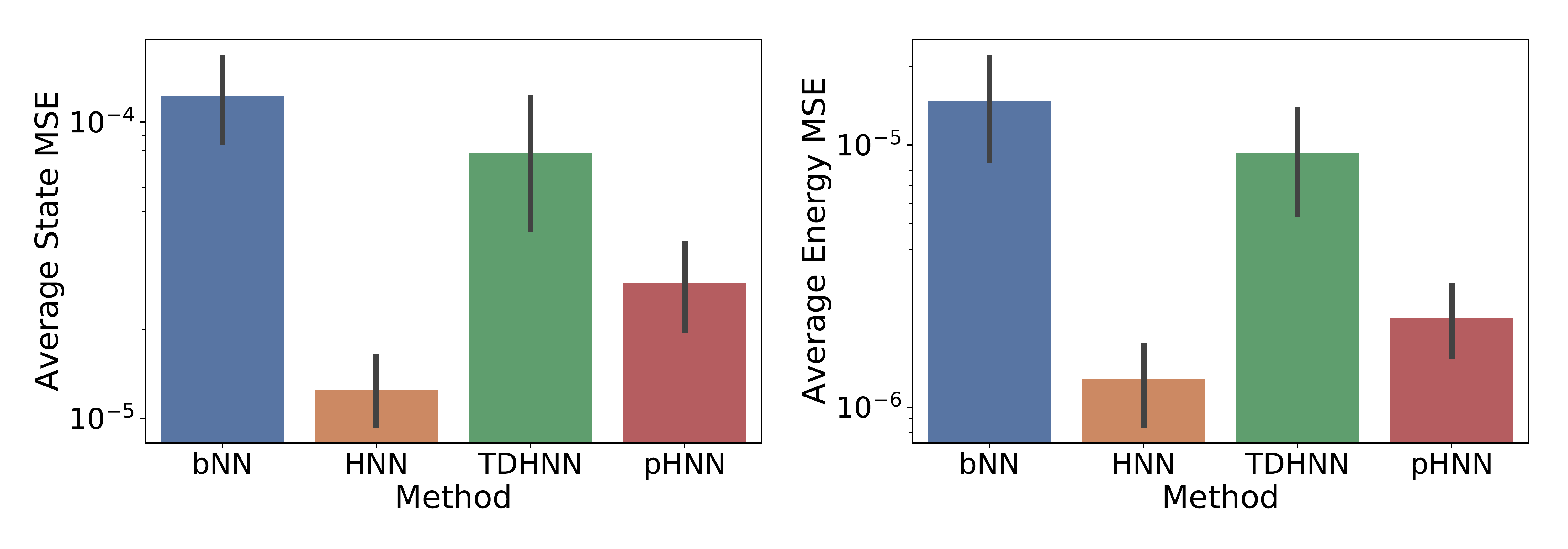}
		\caption{State and energy MSE averaged across 25 initial test states (error bars showing $\pm 1 \sigma$).}
	\end{subfigure}
\caption{The simple mass-spring system has no explicit time dependence. We see that the pHNN can almost recover the dynamics as well as the HNN. While the pHNN does learn a non-zero force and damping term, their contribution to $\frac{dp}{dt}$ is small.
}
\label{mspring}
\end{figure}

In particular, in Fig. \ref{mspring}(a) we show the state and energy MSE for an initial state from the testing set. Figure \ref{mspring}(b) presents the predicted force (blue solid line) and damping over time $\nu \frac{\partial H}{\partial p}$ (red solid curve), while black dots corresponds to ground truth observations. We observe that the error in the predictions is of the order of $10^{-5}$ for recovering the force function and of $10^{-8}$ for the damping term. In Fig. \ref{mspring}(c) we report the state and energy MSE averaged along all the testing initial states, where the black lines in the histograms represent the error bars of the statistics.

\subsection{Damped Mass-Spring System}

We extend the simple mass-spring system to include a damping term that reduces the initial energy of the system over time. The inclusion of this term violates energy conservation and therefore we cannot write a scalar Hamiltonian for such a system (see Appendix C for details). Knowing this a priori already gives us an indication that the HNN will perform poorly on such a system.

\textbf{Training:} We have 20 initial training conditions, with position and momentum uniformly sampled in $[-1,1]^2$. Each trajectory is evolved until $T_{\max} = 30.1$ with a $\Delta t = 0.1$. We fix the damping coefficient $\nu = 0.3$ without loss of generality.

\begin{figure}[h!]
\centering
\captionsetup{justification=centering}
	\begin{subfigure}[b]{0.48\textwidth}
		\centering
		\includegraphics[width=\textwidth]{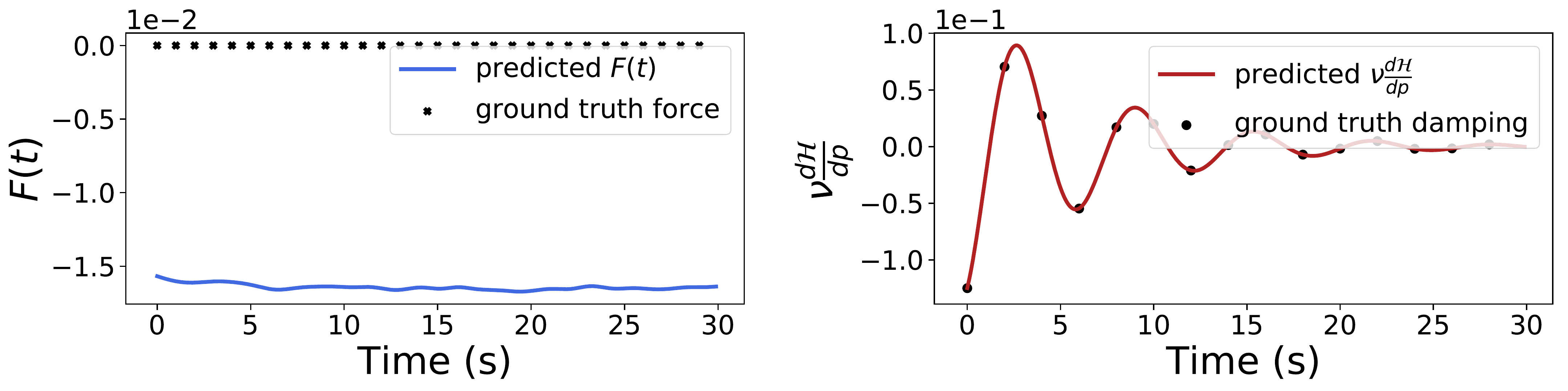}
		\caption{Learnt Force and Damping terms by pHNN}
	\end{subfigure}
	\begin{subfigure}[b]{0.48\textwidth}
	    \centering
		\includegraphics[width=\textwidth]{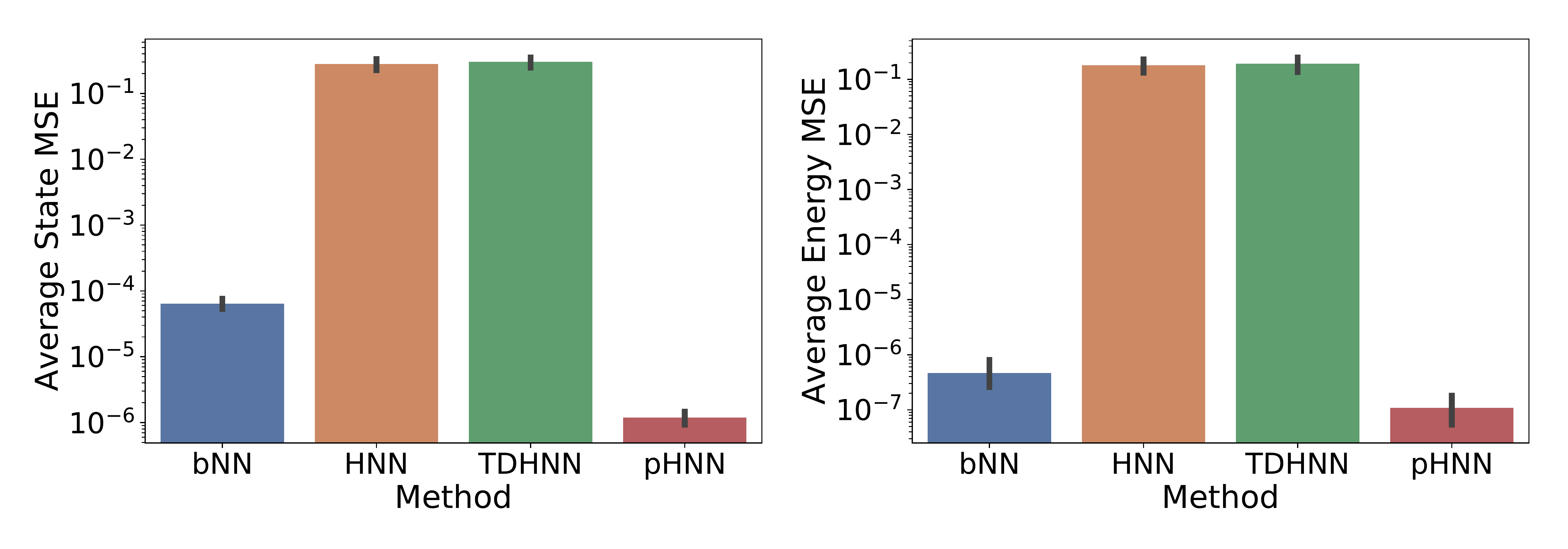}
		\caption{State and energy MSE averaged across 25 initial test states(error bars showing $\pm 1 \sigma$).}
	\end{subfigure}
\caption{Damped mass-spring setting: The baseline NN and pHNN recover the underlying dynamics well. pHNN is also able to accurately learn the damping coefficient since the predicted damping is indistinguishable from the ground truth.}
\label{damped}
\end{figure}

\textbf{Testing:} At inference, we compute the average rollout MSE of 25 unseen initial conditions sampled in the same manner as the training data.  Figure \ref{damped}(a) outlines  the predicted force and damping for an arbitrary  initial state, while in Fig. \ref{damped}(b) we report the average state and energy rollout MSE.

We observe in Fig.~\ref{damped} that both baseline NN and pHNN recover the dynamics well, whereas th HNN (as expected) and TDHNN struggle to learn the dynamics of the damped system. This failure happens because there is no direct way of writing a scalar Hamiltonian with damping and thus, both the HNN and TDHNN cannot learn the underlying dynamics. This observation indicates that a partial inductive bias is not enough to make a network robust at predicting the dynamics well. A second observation shows that the pHNN learns a non-zero oscillating force. That implies a possible leaking of the information from the predicted Hamiltonian into the predicted force and vice versa. 
This arises because there is an identifiability challenge inferring the damping and force exclusively from the state data we provide. In spite of this challenge, we find that the pHNN converges to forcing and damping terms consistent with the ground truth generating terms and sufficient to inform us about the underlying dynamics as well as to evolve initial states with small numerical error. 

\subsection{Forced Mass-Spring System}
We complete the investigation of the simple mass-spring problem by including a driven time-dependent force that controls the system. To understand the effect of the force we consider an undamped driven oscillator system. Typically, while we cannot write the Hamiltonian for a damped system, we can write one for a forced system.  We study two cases of forced mass-spring systems. The first has the following Hamiltonian form:
\begin{equation}
\label{eq:hamF1}
\mathcal{H} = \frac{1}{2}kq^2 + \frac{p^2}{2m} - qF_0\sin(\omega t).
\end{equation}
The second has a more complex force  described by the Hamiltonian:
\begin{equation}
\label{eq:hamF2}
\mathcal{H} = \frac{1}{2}kq^2 + \frac{p^2}{2m} - qF_0\sin(\omega t)\sin(2\omega t),
\end{equation}
where $F_0$ and $\omega$ is the amplitude and frequency of the external force term.
The forced mass-spring system is typically used to study resonance effects, e.g. in material science, and plays an important role in a wide range of applications including music, bridge design, and molecular excitation making it an important system to investigate.

\begin{figure}[h!]
\centering
\captionsetup{justification=centering}
	\begin{subfigure}[b]{0.48\textwidth}
		\centering
		\includegraphics[width=\textwidth]{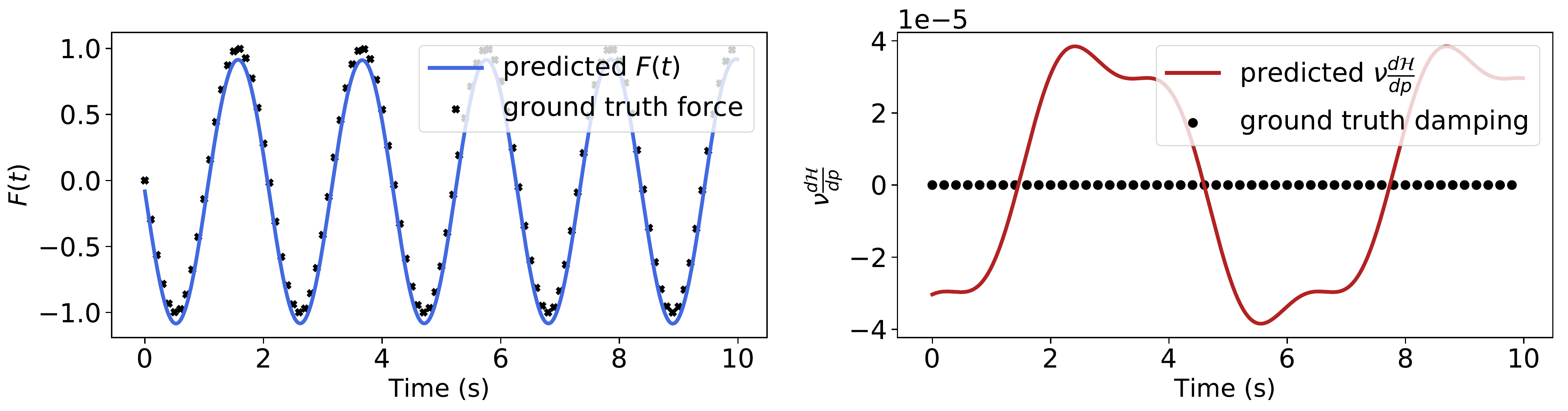}
		\caption{Learnt force and damping terms by pHNN.}
	\end{subfigure}
	\begin{subfigure}[b]{0.48\textwidth}
	    \centering
		\includegraphics[width=\textwidth]{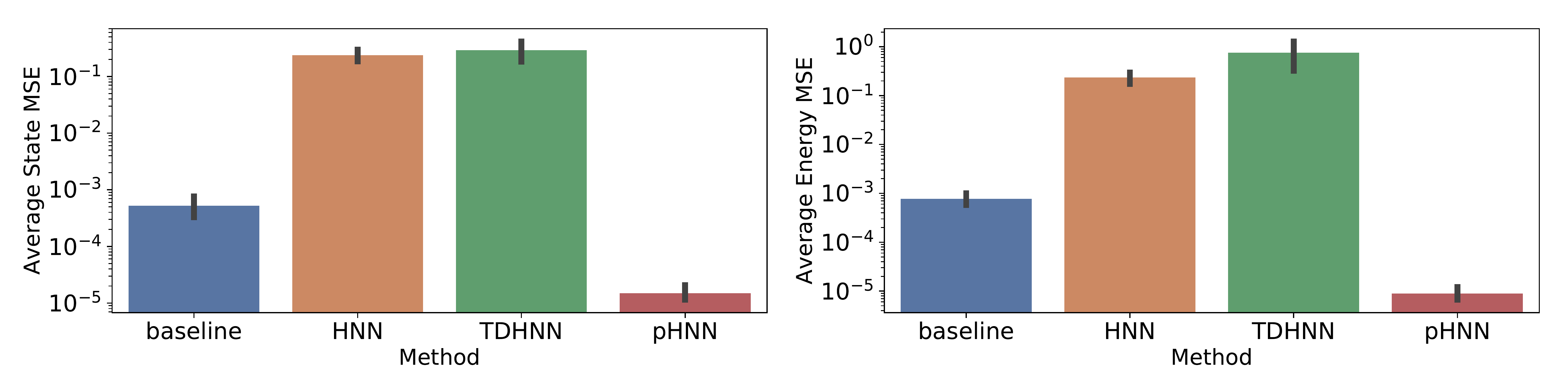}
		\caption{Rollout state and energy MSE averaged across 25 initial test states (error bars showing $\pm 1 \sigma$).}
	\end{subfigure}
\caption{Forced mass-spring of Eq. (\ref{eq:hamF1}): Standard  HNN cannot learn the underlying dynamics as it has no explicit-time dependence. pHNN shows the best performance as it explicitly learns a time-dependent force.}
\label{fig.fmspring1}
\end{figure}

\begin{figure}[h!]
\centering
\captionsetup{justification=centering}
	\begin{subfigure}[b]{0.48\textwidth}
		\centering
		\includegraphics[width=\textwidth]{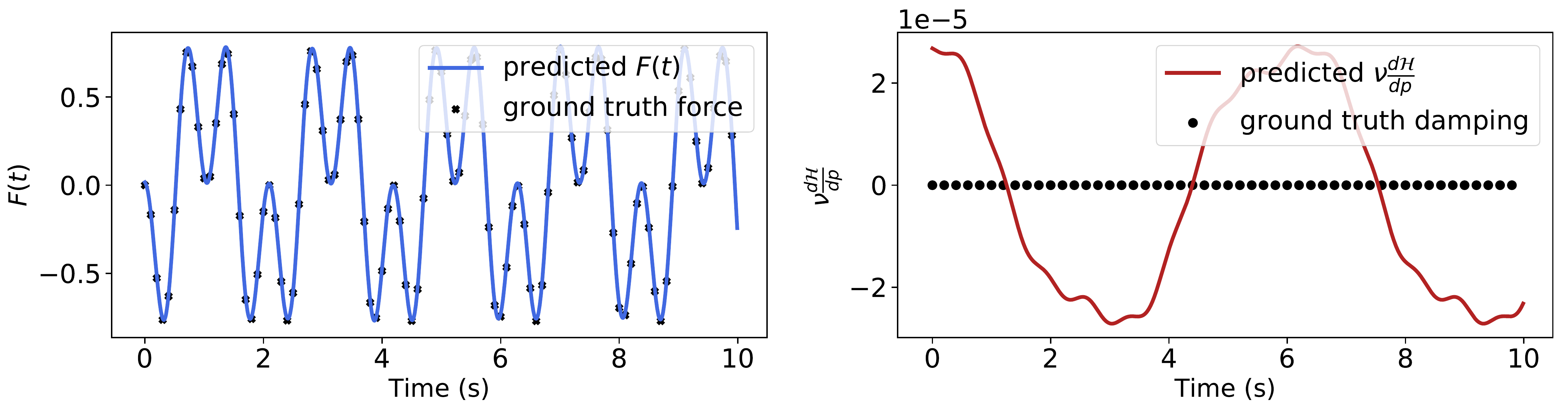}
		\caption{pHNN recovers force and damping.}
	\end{subfigure}
	\begin{subfigure}[b]{0.48\textwidth}
	    \centering
		\includegraphics[width=\textwidth]{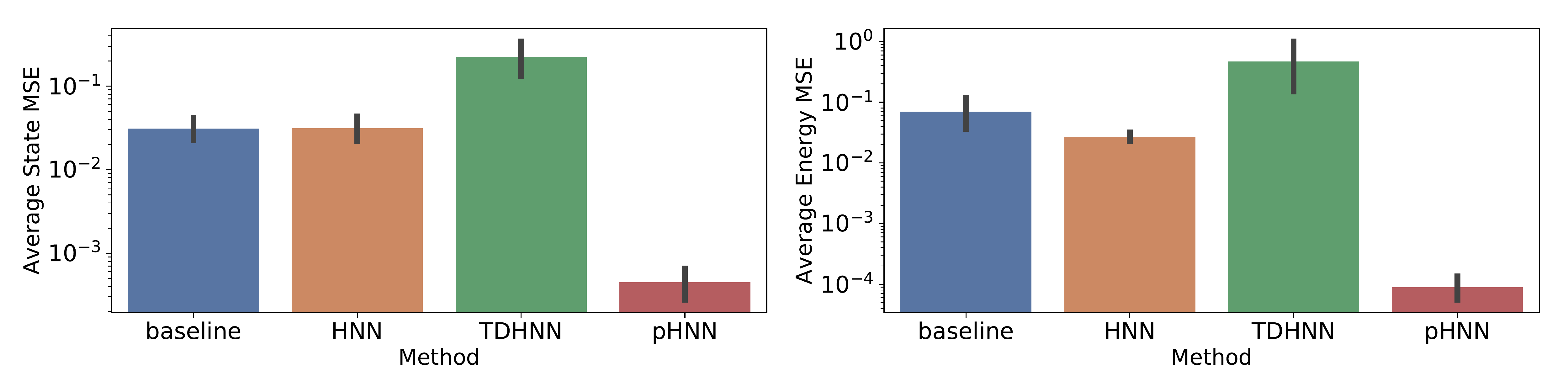}
		\caption{Rollout state and energy MSE averaged across 25 initial test states (error bars showing $\pm 1 \sigma$).}
	\end{subfigure}
\caption{Forced mass-spring system of Eq. (\ref{eq:hamF2}): pHNN is able to recover a non-harmonic force and  evolves testing initial states better than the other models.}
\label{fig.fmspring2}
\end{figure}

\begin{figure*}[ht!]
\centering
\includegraphics[width=0.8\textwidth, trim={0 3cm 0 0},clip]{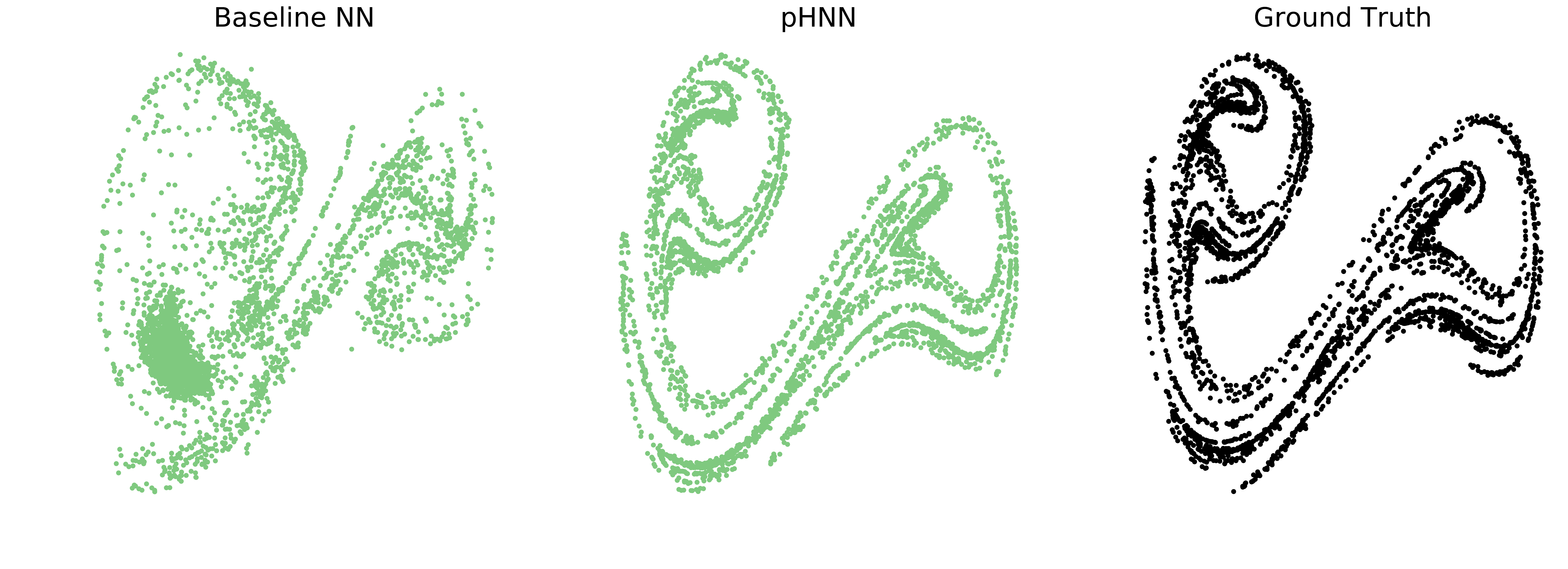}
\caption{Poincar\'e sections of a chaotic Duffing oscillator. Baseline NN (left) and pHNN (middle) are trained for 20000 iterations with 2000 data points. The left and middle images indicate the predicted Poincar\'e map for a initial state not used in networks optimization. The pHNN significantly outperforms the baseline NN at recovering the ground truth Poincar\'e section (right). 
}
\label{fig.chaos1}
\end{figure*}

\textbf{Training:} In both systems of Hamiltonian Eqs.  (\ref{eq:hamF1}) and (\ref{eq:hamF2}), we use 20 initial conditions, where the initial state $[q_0,p_0]$ is sampled such that $q_0^2 +p_0^2 =r_0^2$ where $1 \leq r_0 \leq 4.5$. For the force term we set $F_0=1$ and $\omega=3$, and without loss of generality we set $k=m=1$.  The states are rolled out to $T_{\max}=10.01$ at a $\Delta t = 0.01$.

\textbf{Testing:} At inference, we compute the rollout of 25 unseen initial conditions in the same range as the training data. Figures \ref{fig.fmspring1}a and  \ref{fig.fmspring2}a  demonstrate the predicted forces and dissipation terms for the systems of Eqs. (\ref{eq:hamF1})  and (\ref{eq:hamF2}), respectively. Accordingly, in Figs. \ref{fig.fmspring1}(b) and \ref{fig.fmspring2}(b) we
report the average state and energy rollout MSE for each system.

We study both systems to illustrate that while the baseline NN performs relatively well in comparison to the pHNN when a simple force is considered such as in the system of Eq. (\ref{eq:hamF1}), a more complex force significantly hurts its performance in terms of state/energy MSE shown by Fig. \ref{fig.fmspring2}(b). More importantly, we read in Figs. \ref{fig.fmspring1}(a) and \ref{fig.fmspring2}(a) that for both systems pHNN can recover the ground truth force quite precisely  while it learns very small spurious damping terms that do not significantly contribute to the dynamics; the contribution to $\frac{dp}{dt}$ term is of the order $10^{-5}$, which is practically negligible.

\subsection{Duffing Equation}
Another problem that we investigate is given by the Duffing equation, a nonlinear dynamical system that includes both forcing and damping. The unforced and undamped stationary Hamiltonian $\mathcal{H}_\text{stat}$ of the Duffing system is given by:
\begin{equation}
\mathcal{H}_\text{stat} = \frac{p^2}{2m}+ \alpha \frac{q^2}{2} + \beta \frac{q^4}{4}.
\end{equation}
Unlike the simple mass-spring system,  the Duffing equation has an additional quadratic function of $q$ that makes the system non-harmonic. The shape of the potential function can be tuned to be a double-well or a single  well based on the coefficients $\alpha$ and $\beta$. The general Duffing equation includes a time variant force and a damping term proportional to  ${\partial \mathcal{H}_\text{stat}}/{\partial q}$. Typically the Duffing nonlinear equation of motion is written as:
\begin{equation}
\ddot{q} = -\delta \dot{q} -\alpha q -\beta q^3 +\gamma \sin(\omega t). 
\end{equation}
Different combinations of parameters $\alpha,\beta,\delta,\gamma,\omega$  make the Duffing system either chaotic or non-chaotic. We study both regimes. The Duffing equation reveals numerous phenomena of practical importance including frequency hysteresis (e.g. in magnets), elasticity and chaos theory. 

\begin{figure}[h!]
\centering
\captionsetup{justification=centering}
	\begin{subfigure}[b]{0.48\textwidth}
		\centering
		\includegraphics[width=\textwidth]{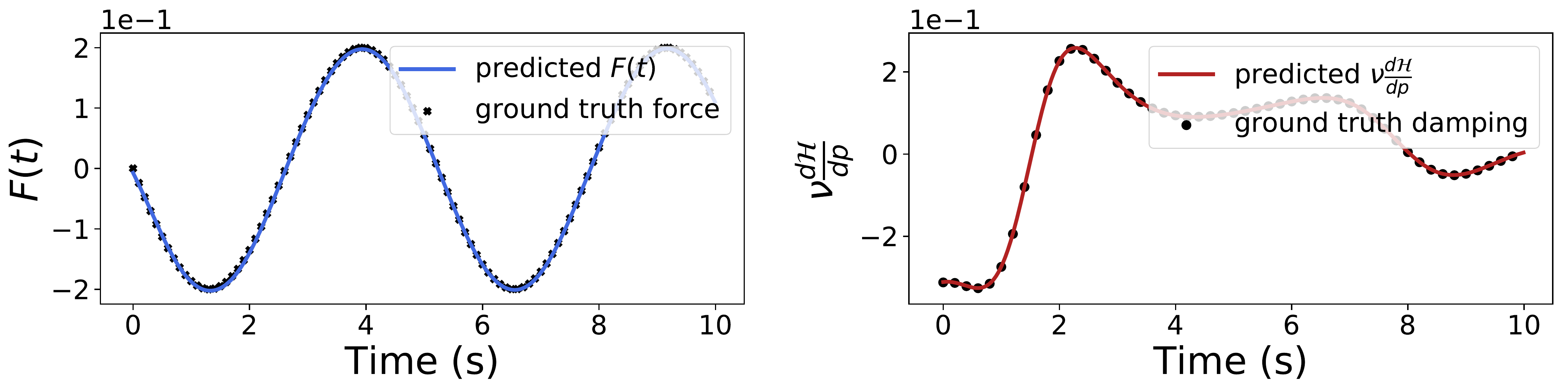}
		\caption{Learnt Force and Damping terms of pHNN}
	\end{subfigure}
	\begin{subfigure}[b]{0.48\textwidth}
	    \centering
		\includegraphics[width=\textwidth]{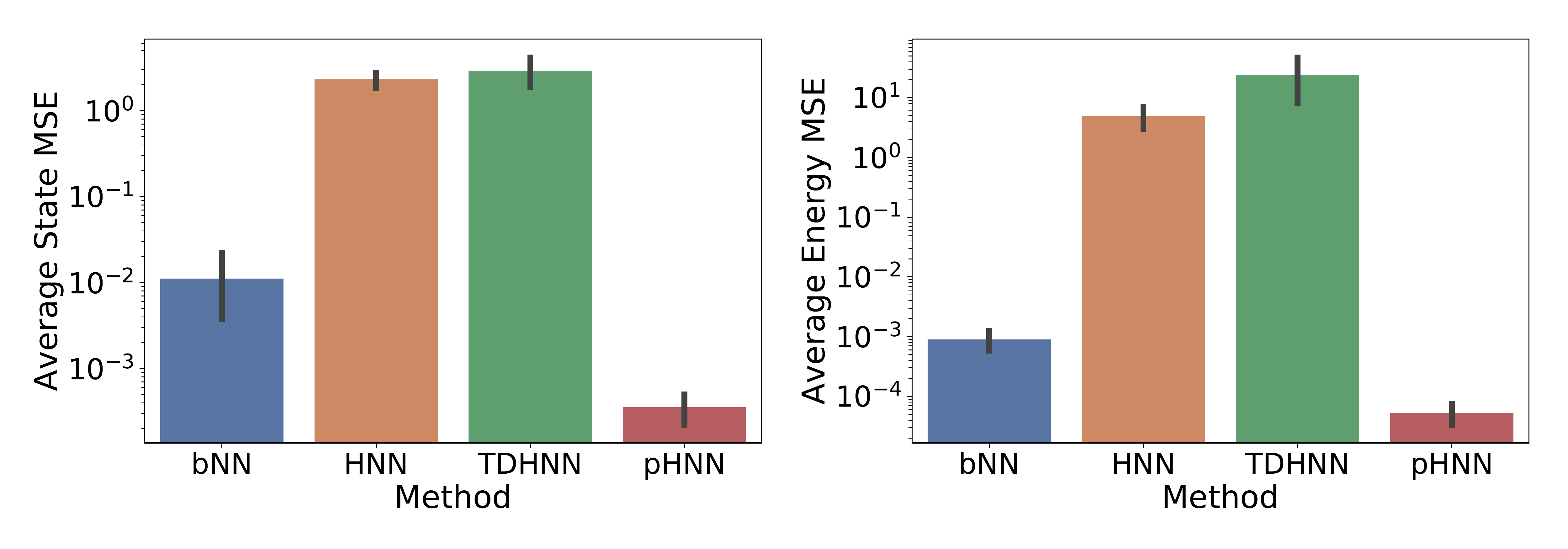}
		\caption{State and energy MSE averaged across 25 initial test states (error bars showing $\pm 1 \sigma$).}
	\end{subfigure}
\caption{Duffing system (non-chaotic): pHNN significantly outperforms the other methods and is able to extract the ground truth force and damping coefficient.}
\label{fig.duffing}
\end{figure}

\begin{figure}[h!]
\centering
\captionsetup{justification=centering}
	\begin{subfigure}[b]{0.48\textwidth}
		\centering
		\includegraphics[width=\textwidth]{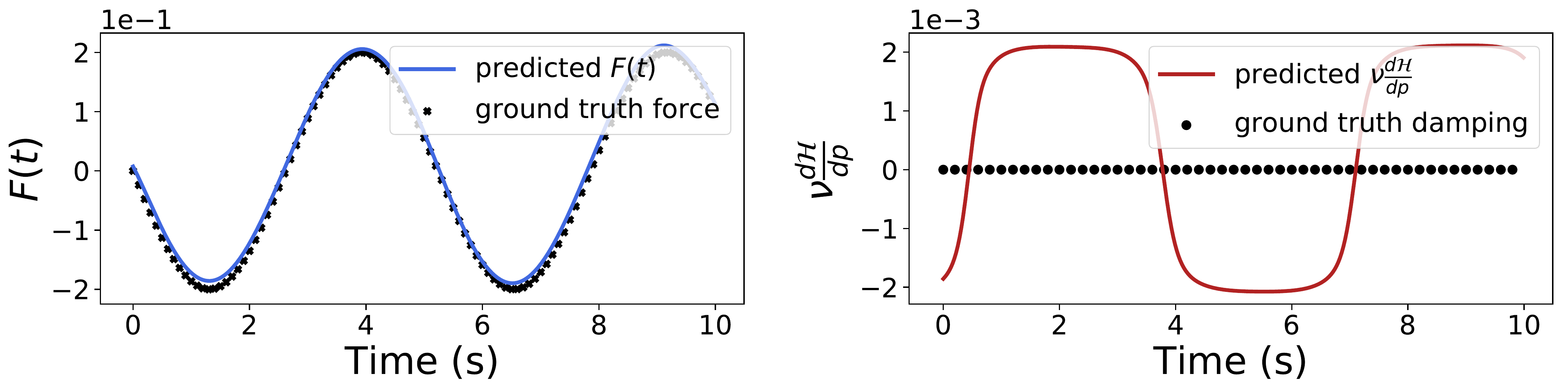}
		\caption{Learnt Force and damping terms of pHNN}
	\end{subfigure}
	\begin{subfigure}[b]{0.48\textwidth}
	    \centering
		\includegraphics[width=\textwidth]{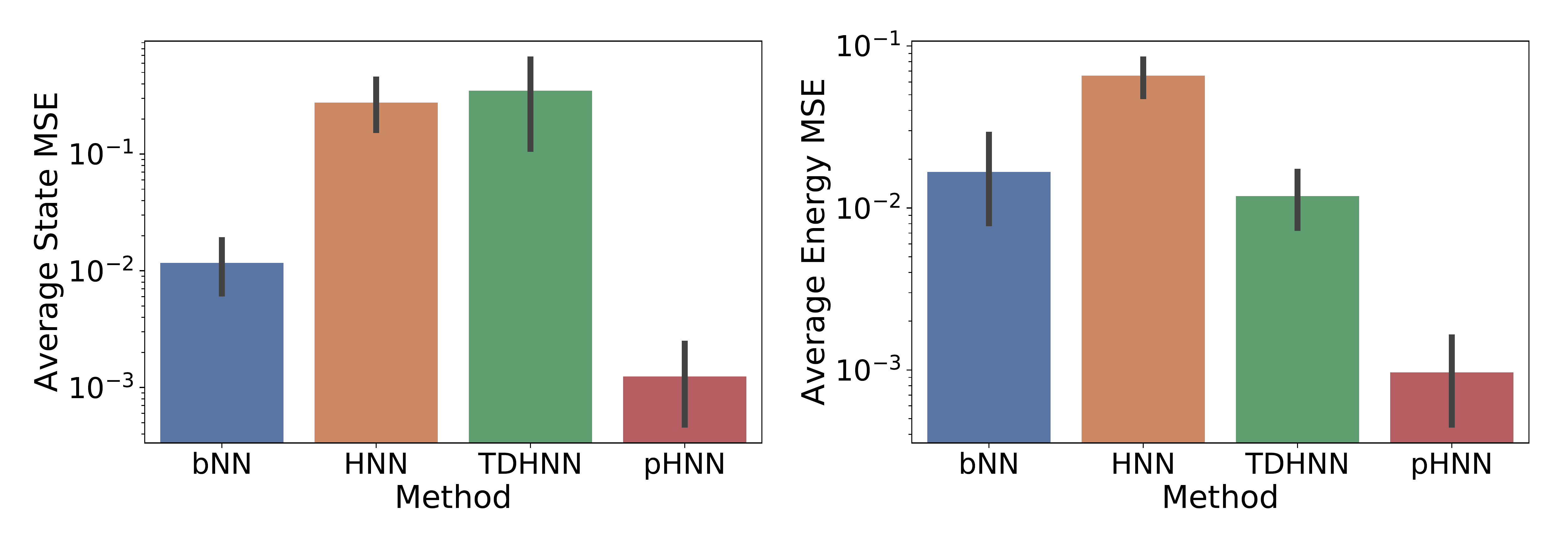}
		\caption{Rollout state and energy MSE averaged across 25 initial test states}
	\end{subfigure}
\caption{Learned dynamics of a relativistic Duffing system.} \label{fig:relativistic}
\end{figure}
\subsubsection{Non-Chaotic regime}

Given a set of initial parameters for the Duffing equation: $\alpha =-1,\beta=1,\delta=0.3,\gamma=0.2,\omega=1.2$ we can obtain training data in a  non-chaotic regime of the Duffing system. 

\textbf{Training:} We uniformly sample initial conditions in $[-1,1]^2$ and use 25 initial conditions for training, rolled out to $T_{\max}=10.01$ with $\Delta t =0.01$. 

\begin{figure*}[htb]
\centering
\includegraphics[width=0.9\textwidth]{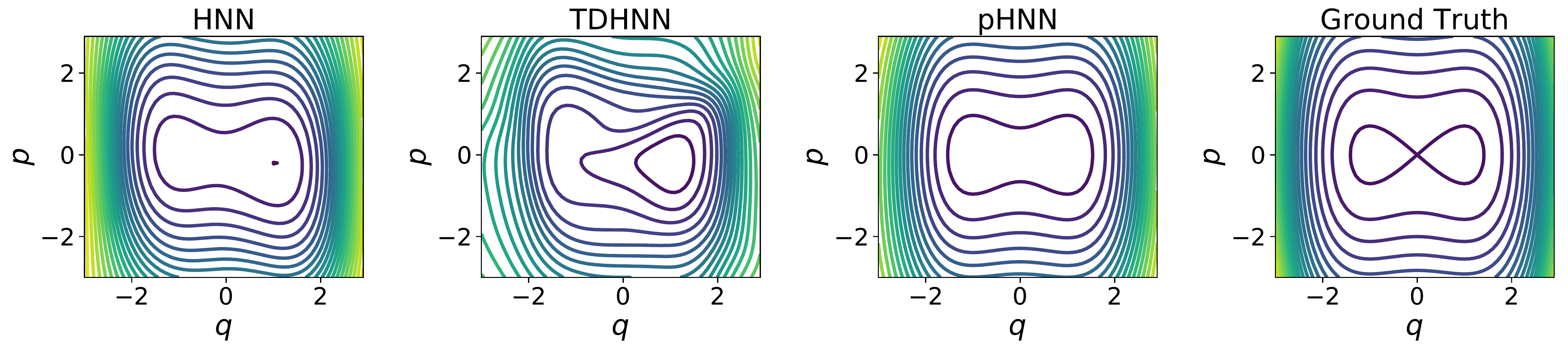}
\caption{Non-chaotic Duffing setting: Learnt $\mathcal{H}_\text{stat}$ in $q-p$ plane (phase space)  across different network architectures. HNN and TDHNN learn distorted Hamiltonians that strongly depend on the input time-variable, whereas pHNN is able to recover a non-distorted Hamiltonian.
}
\label{duffing_ham}
\end{figure*}

\textbf{Testing:} We integrate 25 unseen initial conditions at inference using the same $T_{\max}$ and $\Delta t$ used to generate the training set. We evaluate all the neural network models on this testing set and present the results in Fig. \ref{fig.duffing}. We observe in  Fig. \ref{fig.duffing}(a) that pHNN accurately recovers the underlying force and damping. Moreover, Fig. \ref{fig.duffing}(b) indicates the pHNN outperforms the other models used in this study.
We further assess the network performance by inspecting the predicted Hamiltonian. In  Fig. \ref{duffing_ham} we outline the learnt $\mathcal{H}_\text{stat}$ as a function of $q$ and $p$ that comprise the phase space. 
We observe that pHNN can learn the functional form of $\mathcal{H}_\text{stat}$, outperforming the other architectures.

\subsubsection{Chaotic regime}
The choice of the  parameters: $\alpha =1,\beta=1,\delta=0.1,\gamma=0.39,\omega=1.4$, yields chaotic behavior in the Duffing system. Chaotic trajectories are highly sensitive to initial conditions and thus, it is much more difficult to learn from a chaotic system than from a non-chaotic.

\textbf{Training:} 20 initial conditions, sampled uniformly in $[-1,1]^2$ each rolled out for one period $T=2\pi/\omega$ where $\Delta t = T/100$ resulting in 2000 training points.

\textbf{Testing:} We test our system by assessing whether it is visually able to recover the ground truth Poincar\'e section of an initial condition and we focus our attention on baseline NN and pHNN since they are the most performant in the non-chaotic regime. The Poincar\'e map (or section) of a trajectory is measured by plotting the position and momentum values at regular intervals governed by the period of the forcing term. For example, a simple mass-spring system will generate a single point in phase space when measured at regular intervals, while a chaotic system generates a more complex map. To visually assess the performance of our network through a Poincar\'e map, we test the system on a single initial condition, not used in the training set, rolled out to $T_{\max} = 18000$ with the same $\Delta t$ as in the training phase. In order to integrate our system to such a large $T_{\max}$ for this example we work under the assumption that we have explicit knowledge of the period of the force, and as such, we normalize the time variable with the period. We emphasize that while prior knowledge of $\omega$ assists the training, it is only used to extract the Poincar\'e map. This is necessary as the models are not explicitly trained on time steps beyond $2\pi/\omega$. The results are demonstrated in Fig. \ref{fig.chaos1}. In our study we find that
the visual similarity between the section generated by pHNN is much closer to the ground truth than baseline. A simple
quantitative measure of this similarity can be computed using the MSE between the 2D histogram plots. The pHNN results in 1.61 and bNN in 4.05. The outcome suggests that the pHNN can indeed be used to model chaos even after being trained with only a few data points from a chaotic trajectory. We believe this is a valuable result, since it implies that the pHNN can be used to model chaotic behavior.

\subsection{Relativistic System}
In the final experiment presented in this study, we go beyond non-relativistic classical mechanics and explore a different form of Hamiltonians. In particular, we
 investigate the motion of a driven relativistic particle in a nonlinear double well potential, which is mathematically represented by the Duffing equation in a relativistic framework. The Hamiltonian under consideration is:
\begin{equation}
\mathcal{H} =  c\sqrt{p^2 +m_0^2c^2} + \frac{\alpha}{2}q^2 +\frac{\beta}{4}q^4 - q\gamma\sin(\omega t),
\end{equation}
where   $c$ is the speed of light that typically is set to 1. For simplicity, we also set the rest mass $m_0=1$, though our framework naturally accounts for other values.

\textbf{Training.} We train on 25 initial conditions, uniformly sampled in $[0,2]^2$. We consider $T_{\max} = 20.01$, $\Delta t = 0.01$, and the  parameters  $\alpha =1, \beta =1,\delta =0,\gamma = 0.2,\omega = 1.2$.

\textbf{Testing.} Using the same parameters as training, we roll out 25 unseen initial conditions and present the results in Fig. \ref{fig:relativistic}. We observe that pHNN is able to recover the underlying force and  outperforms the other architectures in generating the temporal state of previously unknown initial conditions. This experiment is evidence that pHNN can discover the dynamics of temporal systems independent of the form of the underlying Hamiltonian and consistently outperforms the other architectures investigated in this study.

\section{Discussion}

We have shown that pHNN outperforms other approaches in learning complex physical systems (details shown in Appendix  D and E), as well as being able to recover the underlying stationary Hamiltonian, the external time varying force, and the damping term of non-autonomous systems. One challenge in achieving this result is fine-tuning the $\lambda_F$ and $\lambda_N$ regularisation coefficients for the force and damping. In the Duffing setting where we have both terms, it is possible to learn a shifted force i.e. $F = F_0 + \epsilon$. This is possible because the state-vectors $\mathbf{q},\mathbf{p}$ do not provide enough information to simultaneously identify both the Hamiltonian and the force resulting in a leak of information between $\mathcal{H}$ and $F$. 
Nevertheless, we find that a reasonable force and damping term are generally learnt which are sufficient to reveal the underlying dynamics.


While it may be argued that pHNN is constrained to learn and predict within the training time horizon, we believe our method is still versatile at informing us of periodic forcing, since we can inspect the force over time and, essentially, learn the period of the underlying force. This, in turn, can readily be used to renormalize the time variable at periodic intervals to integrate the system beyond the training time as we showed in the Poincar\'e map of Fig. \ref{fig.chaos1}.

We also run the entire set of experiments on noisy input state vectors where the noise is sampled from $\mathcal{N}(0,\sigma)$ with $\sigma \in [0.01,0.1,0.5]$. The details of the results can be found in Appendix F. We find that even with the addition of noise to the input state-vector, pHNN outperforms other methods.

We also carry out a simple study on a 2 body coupled spring system where one of the masses is forced by a cosine varying signal. The results indicate that pHNN performs the best and has potential to scale to larger domains (see Appendix).

\section{Conclusion}

We  have  shown  that  learning  the  dynamics  of  time-dependent non-autonomous systems can be achieved with pHNN, a versatile neural network embedded with the Port-Hamiltonian formulation. Our experimental investigation demonstrates that pHNN outperforms extensions of existing methods in numerous settings. Specifically, we outlined that the proposed network not only learns the underlying dynamics of a simple mass-spring system, achieving comparable performance to the efficient HNN architecture, but it extends to more complex nonlinear forced and damped physical systems.
Furthermore, using pHNN we were able to show, with minimal training data, the ability to  recover the Poincar\'e section of a chaotic driven system. Unlike existing methods, pHNN is able to identify systems where minimal state data is available and reveal the functional form of the controlling force, damping, and underlying stationary Hamiltonian. Collectively, these results form a strong basis for further advances in learning complex systems including, but not limited to, chemical bond forces, robotic motion, and more general controlled dynamics without explicit knowledge of the force and damping. 


\begin{thebibliography}{29}%
\makeatletter
\providecommand \@ifxundefined [1]{%
 \@ifx{#1\undefined}
}%
\providecommand \@ifnum [1]{%
 \ifnum #1\expandafter \@firstoftwo
 \else \expandafter \@secondoftwo
 \fi
}%
\providecommand \@ifx [1]{%
 \ifx #1\expandafter \@firstoftwo
 \else \expandafter \@secondoftwo
 \fi
}%
\providecommand \natexlab [1]{#1}%
\providecommand \enquote  [1]{``#1''}%
\providecommand \bibnamefont  [1]{#1}%
\providecommand \bibfnamefont [1]{#1}%
\providecommand \citenamefont [1]{#1}%
\providecommand \href@noop [0]{\@secondoftwo}%
\providecommand \href [0]{\begingroup \@sanitize@url \@href}%
\providecommand \@href[1]{\@@startlink{#1}\@@href}%
\providecommand \@@href[1]{\endgroup#1\@@endlink}%
\providecommand \@sanitize@url [0]{\catcode `\\12\catcode `\$12\catcode
  `\&12\catcode `\#12\catcode `\^12\catcode `\_12\catcode `\%12\relax}%
\providecommand \@@startlink[1]{}%
\providecommand \@@endlink[0]{}%
\providecommand \url  [0]{\begingroup\@sanitize@url \@url }%
\providecommand \@url [1]{\endgroup\@href {#1}{\urlprefix }}%
\providecommand \urlprefix  [0]{URL }%
\providecommand \Eprint [0]{\href }%
\providecommand \doibase [0]{https://doi.org/}%
\providecommand \selectlanguage [0]{\@gobble}%
\providecommand \bibinfo  [0]{\@secondoftwo}%
\providecommand \bibfield  [0]{\@secondoftwo}%
\providecommand \translation [1]{[#1]}%
\providecommand \BibitemOpen [0]{}%
\providecommand \bibitemStop [0]{}%
\providecommand \bibitemNoStop [0]{.\EOS\space}%
\providecommand \EOS [0]{\spacefactor3000\relax}%
\providecommand \BibitemShut  [1]{\csname bibitem#1\endcsname}%
\let\auto@bib@innerbib\@empty
\bibitem [{\citenamefont {Hornik}\ \emph {et~al.}(1989)\citenamefont {Hornik},
  \citenamefont {Stinchcombe},\ and\ \citenamefont
  {White}}]{hornik_multilayer_1989}%
  \BibitemOpen
  \bibfield  {author} {\bibinfo {author} {\bibfnamefont {K.}~\bibnamefont
  {Hornik}}, \bibinfo {author} {\bibfnamefont {M.}~\bibnamefont
  {Stinchcombe}},\ and\ \bibinfo {author} {\bibfnamefont {H.}~\bibnamefont
  {White}},\ }\bibfield  {title} {{\bibinfo {title}
  {Multilayer feedforward networks are universal approximators}},\ }\href
  {https://doi.org/10.1016/0893-6080(89)90020-8} {\bibfield  {journal}
  {\bibinfo  {journal} {Neural Networks}\ }\textbf {\bibinfo {volume} {2}},\
  \bibinfo {pages} {359} (\bibinfo {year} {1989})}\BibitemShut {NoStop}%
\bibitem [{\citenamefont {He}\ \emph {et~al.}(2018)\citenamefont {He},
  \citenamefont {Gkioxari}, \citenamefont {Dollár},\ and\ \citenamefont
  {Girshick}}]{he_mask_2018}%
  \BibitemOpen
  \bibfield  {author} {\bibinfo {author} {\bibfnamefont {K.}~\bibnamefont
  {He}}, \bibinfo {author} {\bibfnamefont {G.}~\bibnamefont {Gkioxari}},
  \bibinfo {author} {\bibfnamefont {P.}~\bibnamefont {Dollár}},\ and\ \bibinfo
  {author} {\bibfnamefont {R.}~\bibnamefont {Girshick}},\ }\bibfield  {title}
  {\bibinfo {title} {Mask {R}-{CNN}},\ }\href {http://arxiv.org/abs/1703.06870}
  {\bibfield  {journal} {\bibinfo  {journal} {arXiv:1703.06870 [cs]}\ }
  (\bibinfo {year} {2018})},\ \bibinfo {note} {arXiv: 1703.06870}\BibitemShut
  {NoStop}%
\bibitem [{\citenamefont {Devlin}\ \emph {et~al.}(2019)\citenamefont {Devlin},
  \citenamefont {Chang}, \citenamefont {Lee},\ and\ \citenamefont
  {Toutanova}}]{devlin_bert_2019}%
  \BibitemOpen
  \bibfield  {author} {\bibinfo {author} {\bibfnamefont {J.}~\bibnamefont
  {Devlin}}, \bibinfo {author} {\bibfnamefont {M.-W.}\ \bibnamefont {Chang}},
  \bibinfo {author} {\bibfnamefont {K.}~\bibnamefont {Lee}},\ and\ \bibinfo
  {author} {\bibfnamefont {K.}~\bibnamefont {Toutanova}},\ }\bibfield  {title}
  {\bibinfo {title} {{BERT}: {Pre}-training of {Deep} {Bidirectional}
  {Transformers} for {Language} {Understanding}},\ }\href
  {http://arxiv.org/abs/1810.04805} {\bibfield  {journal} {\bibinfo  {journal}
  {arXiv:1810.04805 [cs]}\ } (\bibinfo {year} {2019})},\ \bibinfo {note}
  {arXiv: 1810.04805}\BibitemShut {NoStop}%
\bibitem [{\citenamefont {Toussaint}\ \emph {et~al.}(2018)\citenamefont
  {Toussaint}, \citenamefont {Allen}, \citenamefont {Smith},\ and\
  \citenamefont {Tenenbaum}}]{toussaint_differentiable_2018}%
  \BibitemOpen
  \bibfield  {author} {\bibinfo {author} {\bibfnamefont {M.}~\bibnamefont
  {Toussaint}}, \bibinfo {author} {\bibfnamefont {K.}~\bibnamefont {Allen}},
  \bibinfo {author} {\bibfnamefont {K.}~\bibnamefont {Smith}},\ and\ \bibinfo
  {author} {\bibfnamefont {J.}~\bibnamefont {Tenenbaum}},\ }\bibfield  {title}
  {{\bibinfo {title} {Differentiable {Physics} and {Stable}
  {Modes} for {Tool}-{Use} and {Manipulation} {Planning}}},\ }in\ \href
  {https://doi.org/10.15607/RSS.2018.XIV.044} {{\emph
  {\bibinfo {booktitle} {Robotics: {Science} and {Systems} {XIV}}}}}\ (\bibinfo
   {publisher} {Robotics: Science and Systems Foundation},\ \bibinfo {year}
  {2018})\BibitemShut {NoStop}%
\bibitem [{\citenamefont {Yao}\ \emph {et~al.}(2018)\citenamefont {Yao},
  \citenamefont {Herr}, \citenamefont {Toth}, \citenamefont {Mckintyre},\ and\
  \citenamefont {Parkhill}}]{yao_tensormol-01_2018}%
  \BibitemOpen
  \bibfield  {author} {\bibinfo {author} {\bibfnamefont {K.}~\bibnamefont
  {Yao}}, \bibinfo {author} {\bibfnamefont {J.~E.}\ \bibnamefont {Herr}},
  \bibinfo {author} {\bibfnamefont {D.}~\bibnamefont {Toth}}, \bibinfo {author}
  {\bibfnamefont {R.}~\bibnamefont {Mckintyre}},\ and\ \bibinfo {author}
  {\bibfnamefont {J.}~\bibnamefont {Parkhill}},\ }\bibfield  {title}
  {{\bibinfo {title} {The {TensorMol}-0.1 model chemistry:
  a neural network augmented with long-range physics}},\ }\href
  {https://doi.org/10.1039/C7SC04934J} {\bibfield  {journal} {\bibinfo
  {journal} {Chemical Science}\ }\textbf {\bibinfo {volume} {9}},\ \bibinfo
  {pages} {2261} (\bibinfo {year} {2018})}\BibitemShut {NoStop}%
\bibitem [{\citenamefont {Greydanus}\ \emph {et~al.}(2019)\citenamefont
  {Greydanus}, \citenamefont {Dzamba},\ and\ \citenamefont
  {Yosinski}}]{greydanus_hamiltonian_2019}%
  \BibitemOpen
  \bibfield  {author} {\bibinfo {author} {\bibfnamefont {S.}~\bibnamefont
  {Greydanus}}, \bibinfo {author} {\bibfnamefont {M.}~\bibnamefont {Dzamba}},\
  and\ \bibinfo {author} {\bibfnamefont {J.}~\bibnamefont {Yosinski}},\
  }\bibfield  {title} {\bibinfo {title} {Hamiltonian {Neural} {Networks}},\
  }in\ \href {http://papers.nips.cc/paper/9672-hamiltonian-neural-networks.pdf}
  {\emph {\bibinfo {booktitle} {Advances in {Neural} {Information} {Processing}
  {Systems} 32}}},\ \bibinfo {editor} {edited by\ \bibinfo {editor}
  {\bibfnamefont {H.}~\bibnamefont {Wallach}}, \bibinfo {editor} {\bibfnamefont
  {H.}~\bibnamefont {Larochelle}}, \bibinfo {editor} {\bibfnamefont
  {A.}~\bibnamefont {Beygelzimer}}, \bibinfo {editor} {\bibfnamefont {F.~d.}\
  \bibnamefont {Alché-Buc}}, \bibinfo {editor} {\bibfnamefont
  {E.}~\bibnamefont {Fox}},\ and\ \bibinfo {editor} {\bibfnamefont
  {R.}~\bibnamefont {Garnett}}}\ (\bibinfo  {publisher} {Curran Associates,
  Inc.},\ \bibinfo {year} {2019})\ pp.\ \bibinfo {pages}
  {15379--15389}\BibitemShut {NoStop}%
\bibitem [{\citenamefont {Pukrittayakamee}\ \emph {et~al.}(2009)\citenamefont
  {Pukrittayakamee}, \citenamefont {Malshe}, \citenamefont {Hagan},
  \citenamefont {Raff}, \citenamefont {Narulkar}, \citenamefont {Bukkapatnum},\
  and\ \citenamefont {Komanduri}}]{pukrittayakamee_simultaneous_2009}%
  \BibitemOpen
  \bibfield  {author} {\bibinfo {author} {\bibfnamefont {A.}~\bibnamefont
  {Pukrittayakamee}}, \bibinfo {author} {\bibfnamefont {M.}~\bibnamefont
  {Malshe}}, \bibinfo {author} {\bibfnamefont {M.}~\bibnamefont {Hagan}},
  \bibinfo {author} {\bibfnamefont {L.~M.}\ \bibnamefont {Raff}}, \bibinfo
  {author} {\bibfnamefont {R.}~\bibnamefont {Narulkar}}, \bibinfo {author}
  {\bibfnamefont {S.}~\bibnamefont {Bukkapatnum}},\ and\ \bibinfo {author}
  {\bibfnamefont {R.}~\bibnamefont {Komanduri}},\ }\bibfield  {title}
  {{\bibinfo {title} {Simultaneous fitting of a
  potential-energy surface and its corresponding force fields using feedforward
  neural networks}},\ }\href {https://doi.org/10.1063/1.3095491} {\bibfield
  {journal} {\bibinfo  {journal} {The Journal of Chemical Physics}\ }\textbf
  {\bibinfo {volume} {130}},\ \bibinfo {pages} {134101} (\bibinfo {year}
  {2009})}\BibitemShut {NoStop}%
\bibitem [{\citenamefont {Mattheakis}\ \emph {et~al.}(2020)\citenamefont
  {Mattheakis}, \citenamefont {Sondak}, \citenamefont {Dogra},\ and\
  \citenamefont {Protopapas}}]{mattheakis_hamiltonian_2020}%
  \BibitemOpen
  \bibfield  {author} {\bibinfo {author} {\bibfnamefont {M.}~\bibnamefont
  {Mattheakis}}, \bibinfo {author} {\bibfnamefont {D.}~\bibnamefont {Sondak}},
  \bibinfo {author} {\bibfnamefont {A.~S.}\ \bibnamefont {Dogra}},\ and\
  \bibinfo {author} {\bibfnamefont {P.}~\bibnamefont {Protopapas}},\ }\bibfield
   {title} {\bibinfo {title} {Hamiltonian {Neural} {Networks} for solving
  differential equations},\ }\href {http://arxiv.org/abs/2001.11107} {\bibfield
   {journal} {\bibinfo  {journal} {arXiv:2001.11107 [physics]}\ } (\bibinfo
  {year} {2020})},\ \bibinfo {note} {arXiv: 2001.11107}\BibitemShut {NoStop}%
\bibitem [{\citenamefont {Cranmer}\ \emph {et~al.}(2020)\citenamefont
  {Cranmer}, \citenamefont {Greydanus}, \citenamefont {Hoyer}, \citenamefont
  {Battaglia}, \citenamefont {Spergel},\ and\ \citenamefont
  {Ho}}]{cranmer_lagrangian_2020}%
  \BibitemOpen
  \bibfield  {author} {\bibinfo {author} {\bibfnamefont {M.}~\bibnamefont
  {Cranmer}}, \bibinfo {author} {\bibfnamefont {S.}~\bibnamefont {Greydanus}},
  \bibinfo {author} {\bibfnamefont {S.}~\bibnamefont {Hoyer}}, \bibinfo
  {author} {\bibfnamefont {P.}~\bibnamefont {Battaglia}}, \bibinfo {author}
  {\bibfnamefont {D.}~\bibnamefont {Spergel}},\ and\ \bibinfo {author}
  {\bibfnamefont {S.}~\bibnamefont {Ho}},\ }\bibfield  {title} {\bibinfo
  {title} {Lagrangian {Neural} {Networks}},\ }\href
  {http://arxiv.org/abs/2003.04630} {\bibfield  {journal} {\bibinfo  {journal}
  {arXiv:2003.04630 [physics, stat]}\ } (\bibinfo {year} {2020})},\ \bibinfo
  {note} {arXiv: 2003.04630}\BibitemShut {NoStop}%
\bibitem [{\citenamefont {Lutter}\ \emph {et~al.}(2019)\citenamefont {Lutter},
  \citenamefont {Ritter},\ and\ \citenamefont {Peters}}]{lutter_deep_2019}%
  \BibitemOpen
  \bibfield  {author} {\bibinfo {author} {\bibfnamefont {M.}~\bibnamefont
  {Lutter}}, \bibinfo {author} {\bibfnamefont {C.}~\bibnamefont {Ritter}},\
  and\ \bibinfo {author} {\bibfnamefont {J.}~\bibnamefont {Peters}},\
  }\bibfield  {title} {\bibinfo {title} {Deep {Lagrangian} {Networks}: {Using}
  {Physics} as {Model} {Prior} for {Deep} {Learning}},\ }\href
  {http://arxiv.org/abs/1907.04490} {\bibfield  {journal} {\bibinfo  {journal}
  {arXiv:1907.04490 [cs, eess, stat]}\ } (\bibinfo {year} {2019})},\ \bibinfo
  {note} {arXiv: 1907.04490}\BibitemShut {NoStop}%
\bibitem [{\citenamefont {Chen}\ \emph {et~al.}(2018)\citenamefont {Chen},
  \citenamefont {Rubanova}, \citenamefont {Bettencourt},\ and\ \citenamefont
  {Duvenaud}}]{chen_neural_2018}%
  \BibitemOpen
  \bibfield  {author} {\bibinfo {author} {\bibfnamefont {R.~T.~Q.}\
  \bibnamefont {Chen}}, \bibinfo {author} {\bibfnamefont {Y.}~\bibnamefont
  {Rubanova}}, \bibinfo {author} {\bibfnamefont {J.}~\bibnamefont
  {Bettencourt}},\ and\ \bibinfo {author} {\bibfnamefont {D.~K.}\ \bibnamefont
  {Duvenaud}},\ }\bibfield  {title} {\bibinfo {title} {Neural {Ordinary}
  {Differential} {Equations}},\ }in\ \href
  {http://papers.nips.cc/paper/7892-neural-ordinary-differential-equations.pdf}
  {\emph {\bibinfo {booktitle} {Advances in {Neural} {Information} {Processing}
  {Systems} 31}}},\ \bibinfo {editor} {edited by\ \bibinfo {editor}
  {\bibfnamefont {S.}~\bibnamefont {Bengio}}, \bibinfo {editor} {\bibfnamefont
  {H.}~\bibnamefont {Wallach}}, \bibinfo {editor} {\bibfnamefont
  {H.}~\bibnamefont {Larochelle}}, \bibinfo {editor} {\bibfnamefont
  {K.}~\bibnamefont {Grauman}}, \bibinfo {editor} {\bibfnamefont
  {N.}~\bibnamefont {Cesa-Bianchi}},\ and\ \bibinfo {editor} {\bibfnamefont
  {R.}~\bibnamefont {Garnett}}}\ (\bibinfo  {publisher} {Curran Associates,
  Inc.},\ \bibinfo {year} {2018})\ pp.\ \bibinfo {pages}
  {6571--6583}\BibitemShut {NoStop}%
\bibitem [{\citenamefont {Raissi}\ \emph {et~al.}(2017)\citenamefont {Raissi},
  \citenamefont {Perdikaris},\ and\ \citenamefont
  {Karniadakis}}]{raissi_physics_2017}%
  \BibitemOpen
  \bibfield  {author} {\bibinfo {author} {\bibfnamefont {M.}~\bibnamefont
  {Raissi}}, \bibinfo {author} {\bibfnamefont {P.}~\bibnamefont {Perdikaris}},\
  and\ \bibinfo {author} {\bibfnamefont {G.~E.}\ \bibnamefont {Karniadakis}},\
  }\bibfield  {title} {\bibinfo {title} {Physics {Informed} {Deep} {Learning}
  ({Part} {I}): {Data}-driven {Solutions} of {Nonlinear} {Partial}
  {Differential} {Equations}},\ }\href {http://arxiv.org/abs/1711.10561}
  {\bibfield  {journal} {\bibinfo  {journal} {arXiv:1711.10561 [cs, math,
  stat]}\ } (\bibinfo {year} {2017})},\ \bibinfo {note} {arXiv:
  1711.10561}\BibitemShut {NoStop}%
\bibitem [{\citenamefont {Choudhary}\ \emph {et~al.}(2020)\citenamefont
  {Choudhary}, \citenamefont {Lindner}, \citenamefont {Holliday}, \citenamefont
  {Miller}, \citenamefont {Sinha},\ and\ \citenamefont {Ditto}}]{HNN_PRE2020}%
  \BibitemOpen
  \bibfield  {author} {\bibinfo {author} {\bibfnamefont {A.}~\bibnamefont
  {Choudhary}}, \bibinfo {author} {\bibfnamefont {J.~F.}\ \bibnamefont
  {Lindner}}, \bibinfo {author} {\bibfnamefont {E.~G.}\ \bibnamefont
  {Holliday}}, \bibinfo {author} {\bibfnamefont {S.~T.}\ \bibnamefont
  {Miller}}, \bibinfo {author} {\bibfnamefont {S.}~\bibnamefont {Sinha}},\ and\
  \bibinfo {author} {\bibfnamefont {W.~L.}\ \bibnamefont {Ditto}},\ }\bibfield
  {title} {\bibinfo {title} {Physics-enhanced neural networks learn order and
  chaos},\ }\href {https://doi.org/10.1103/PhysRevE.101.062207} {\bibfield
  {journal} {\bibinfo  {journal} {Phys. Rev. E}\ }\textbf {\bibinfo {volume}
  {101}},\ \bibinfo {pages} {062207} (\bibinfo {year} {2020})}\BibitemShut
  {NoStop}%
\bibitem [{\citenamefont {Toth}\ \emph {et~al.}(2020)\citenamefont {Toth},
  \citenamefont {Rezende}, \citenamefont {Jaegle}, \citenamefont {Racaniere},
  \citenamefont {Botev},\ and\ \citenamefont {Higgins}}]{HNN_iclr2020}%
  \BibitemOpen
  \bibfield  {author} {\bibinfo {author} {\bibfnamefont {P.}~\bibnamefont
  {Toth}}, \bibinfo {author} {\bibfnamefont {D.~J.}\ \bibnamefont {Rezende}},
  \bibinfo {author} {\bibfnamefont {A.}~\bibnamefont {Jaegle}}, \bibinfo
  {author} {\bibfnamefont {S.}~\bibnamefont {Racaniere}}, \bibinfo {author}
  {\bibfnamefont {A.}~\bibnamefont {Botev}},\ and\ \bibinfo {author}
  {\bibfnamefont {I.}~\bibnamefont {Higgins}},\ }\bibfield  {title} {\bibinfo
  {title} {Hamiltonian generative networks},\ }in\ \href@noop {} {\emph
  {\bibinfo {booktitle} {International Conference on Learning
  Representations}}}\ (\bibinfo {year} {2020})\BibitemShut {NoStop}%
\bibitem [{\citenamefont {Battaglia}\ \emph {et~al.}(2016)\citenamefont
  {Battaglia}, \citenamefont {Pascanu}, \citenamefont {Lai}, \citenamefont
  {Rezende},\ and\ \citenamefont {Kavukcuoglu}}]{battaglia_interaction_2016}%
  \BibitemOpen
  \bibfield  {author} {\bibinfo {author} {\bibfnamefont {P.~W.}\ \bibnamefont
  {Battaglia}}, \bibinfo {author} {\bibfnamefont {R.}~\bibnamefont {Pascanu}},
  \bibinfo {author} {\bibfnamefont {M.}~\bibnamefont {Lai}}, \bibinfo {author}
  {\bibfnamefont {D.}~\bibnamefont {Rezende}},\ and\ \bibinfo {author}
  {\bibfnamefont {K.}~\bibnamefont {Kavukcuoglu}},\ }\bibfield  {title}
  {\bibinfo {title} {Interaction {Networks} for {Learning} about {Objects},
  {Relations} and {Physics}},\ }\href {http://arxiv.org/abs/1612.00222}
  {\bibfield  {journal} {\bibinfo  {journal} {arXiv:1612.00222 [cs]}\ }
  (\bibinfo {year} {2016})},\ \bibinfo {note} {arXiv: 1612.00222}\BibitemShut
  {NoStop}%
\bibitem [{\citenamefont {Sanchez-Gonzalez}\ \emph {et~al.}(2019)\citenamefont
  {Sanchez-Gonzalez}, \citenamefont {Bapst}, \citenamefont {Cranmer},\ and\
  \citenamefont {Battaglia}}]{sanchez-gonzalez_hamiltonian_2019}%
  \BibitemOpen
  \bibfield  {author} {\bibinfo {author} {\bibfnamefont {A.}~\bibnamefont
  {Sanchez-Gonzalez}}, \bibinfo {author} {\bibfnamefont {V.}~\bibnamefont
  {Bapst}}, \bibinfo {author} {\bibfnamefont {K.}~\bibnamefont {Cranmer}},\
  and\ \bibinfo {author} {\bibfnamefont {P.}~\bibnamefont {Battaglia}},\
  }\bibfield  {title} {\bibinfo {title} {Hamiltonian {Graph} {Networks} with
  {ODE} {Integrators}},\ }\href {http://arxiv.org/abs/1909.12790} {\bibfield
  {journal} {\bibinfo  {journal} {arXiv:1909.12790 [physics]}\ } (\bibinfo
  {year} {2019})},\ \bibinfo {note} {arXiv: 1909.12790}\BibitemShut {NoStop}%
\bibitem [{\citenamefont {van~der
  Schaft}(2007)}]{sanz-sole_port-hamiltonian_2007}%
  \BibitemOpen
  \bibfield  {author} {\bibinfo {author} {\bibfnamefont {A.}~\bibnamefont
  {van~der Schaft}},\ }\bibfield  {title} {{\bibinfo
  {title} {Port-{Hamiltonian} systems: an introductory survey}},\ }in\ \href
  {https://doi.org/10.4171/022-3/65} {{\emph {\bibinfo
  {booktitle} {Proceedings of the {International} {Congress} of
  {Mathematicians} {Madrid}, {August} 22–30, 2006}}}},\ \bibinfo {editor}
  {edited by\ \bibinfo {editor} {\bibfnamefont {M.}~\bibnamefont {Sanz-Solé}},
  \bibinfo {editor} {\bibfnamefont {J.}~\bibnamefont {Soria}}, \bibinfo
  {editor} {\bibfnamefont {J.~L.}\ \bibnamefont {Varona}},\ and\ \bibinfo
  {editor} {\bibfnamefont {J.}~\bibnamefont {Verdera}}}\ (\bibinfo  {publisher}
  {European Mathematical Society Publishing House},\ \bibinfo {address}
  {Zuerich, Switzerland},\ \bibinfo {year} {2007})\ pp.\ \bibinfo {pages}
  {1339--1365}\BibitemShut {NoStop}%
\bibitem [{\citenamefont {Ortega}\ \emph {et~al.}(2002)\citenamefont {Ortega},
  \citenamefont {{van der Schaft}}, \citenamefont {Maschke},\ and\
  \citenamefont {Escobar}}]{ORTEGA2002585}%
  \BibitemOpen
  \bibfield  {author} {\bibinfo {author} {\bibfnamefont {R.}~\bibnamefont
  {Ortega}}, \bibinfo {author} {\bibfnamefont {A.}~\bibnamefont {{van der
  Schaft}}}, \bibinfo {author} {\bibfnamefont {B.}~\bibnamefont {Maschke}},\
  and\ \bibinfo {author} {\bibfnamefont {G.}~\bibnamefont {Escobar}},\
  }\bibfield  {title} {\bibinfo {title} {Interconnection and damping assignment
  passivity-based control of port-controlled hamiltonian systems},\ }\href
  {https://doi.org/https://doi.org/10.1016/S0005-1098(01)00278-3} {\bibfield
  {journal} {\bibinfo  {journal} {Automatica}\ }\textbf {\bibinfo {volume}
  {38}},\ \bibinfo {pages} {585 } (\bibinfo {year} {2002})}\BibitemShut
  {NoStop}%
\bibitem [{\citenamefont {Acosta}\ \emph {et~al.}(2005)\citenamefont {Acosta},
  \citenamefont {Ortega}, \citenamefont {Astolfi},\ and\ \citenamefont
  {Mahindrakar}}]{acosta_interconnection_2005}%
  \BibitemOpen
  \bibfield  {author} {\bibinfo {author} {\bibfnamefont {J.~A.}\ \bibnamefont
  {Acosta}}, \bibinfo {author} {\bibfnamefont {R.}~\bibnamefont {Ortega}},
  \bibinfo {author} {\bibfnamefont {A.}~\bibnamefont {Astolfi}},\ and\ \bibinfo
  {author} {\bibfnamefont {A.~D.}\ \bibnamefont {Mahindrakar}},\ }\bibfield
  {title} {\bibinfo {title} {Interconnection and damping assignment
  passivity-based control of mechanical systems with underactuation degree
  one},\ }\href {https://doi.org/10.1109/TAC.2005.860292} {\bibfield  {journal}
  {\bibinfo  {journal} {IEEE Transactions on Automatic Control}\ }\textbf
  {\bibinfo {volume} {50}},\ \bibinfo {pages} {1936} (\bibinfo {year}
  {2005})},\ \bibinfo {note} {conference Name: IEEE Transactions on Automatic
  Control}\BibitemShut {NoStop}%
\bibitem [{\citenamefont {Zheng}\ \emph {et~al.}(2018)\citenamefont {Zheng},
  \citenamefont {Yuan},\ and\ \citenamefont {Huang}}]{zheng_time-varying_2018}%
  \BibitemOpen
  \bibfield  {author} {\bibinfo {author} {\bibfnamefont {M.}~\bibnamefont
  {Zheng}}, \bibinfo {author} {\bibfnamefont {T.}~\bibnamefont {Yuan}},\ and\
  \bibinfo {author} {\bibfnamefont {T.}~\bibnamefont {Huang}},\ }\href
  {https://doi.org/https://doi.org/10.1155/2018/8134230} {\bibinfo {title} {Time-{Varying} {Impedance} {Control} of {Port}
  {Hamiltonian} {System} with a {New} {Energy}-{Storing} {Tank}}} (\bibinfo
  {year} {2018}),\ \bibinfo {note} {iSSN: 1076-2787 Pages: e8134230 Publisher:
  Hindawi Volume: 2018}\BibitemShut {NoStop}%
\bibitem [{\citenamefont {Cherifi}(2020)}]{cherifi_overview_2020}%
  \BibitemOpen
  \bibfield  {author} {\bibinfo {author} {\bibfnamefont {K.}~\bibnamefont
  {Cherifi}},\ }\bibfield  {title} {{\bibinfo {title} {An
  overview on recent machine learning techniques for {Port} {Hamiltonian}
  systems}},\ }\href {https://doi.org/10.1016/j.physd.2020.132620} {\bibfield
  {journal} {\bibinfo  {journal} {Physica D: Nonlinear Phenomena}\ }\textbf
  {\bibinfo {volume} {411}},\ \bibinfo {pages} {132620} (\bibinfo {year}
  {2020})}\BibitemShut {NoStop}%
\bibitem [{\citenamefont {Zhong}\ \emph {et~al.}(2020)\citenamefont {Zhong},
  \citenamefont {Dey},\ and\ \citenamefont
  {Chakraborty}}]{zhong_dissipative_2020}%
  \BibitemOpen
  \bibfield  {author} {\bibinfo {author} {\bibfnamefont {Y.~D.}\ \bibnamefont
  {Zhong}}, \bibinfo {author} {\bibfnamefont {B.}~\bibnamefont {Dey}},\ and\
  \bibinfo {author} {\bibfnamefont {A.}~\bibnamefont {Chakraborty}},\
  }\bibfield  {title} {\bibinfo {title} {Dissipative {SymODEN}: {Encoding}
  {Hamiltonian} {Dynamics} with {Dissipation} and {Control} into {Deep}
  {Learning}},\ }\href {http://arxiv.org/abs/2002.08860} {\bibfield  {journal}
  {\bibinfo  {journal} {arXiv:2002.08860 [cs, eess, stat]}\ } (\bibinfo {year}
  {2020})},\ \bibinfo {note} {arXiv: 2002.08860}\BibitemShut {NoStop}%
\bibitem [{\citenamefont {Raissi}\ \emph {et~al.}(2019)\citenamefont {Raissi},
  \citenamefont {Perdikaris},\ and\ \citenamefont
  {Karniadakis}}]{raissi_physics-informed_2019}%
  \BibitemOpen
  \bibfield  {author} {\bibinfo {author} {\bibfnamefont {M.}~\bibnamefont
  {Raissi}}, \bibinfo {author} {\bibfnamefont {P.}~\bibnamefont {Perdikaris}},\
  and\ \bibinfo {author} {\bibfnamefont {G.~E.}\ \bibnamefont {Karniadakis}},\
  }\bibfield  {title} {{\bibinfo {title} {Physics-informed
  neural networks: {A} deep learning framework for solving forward and inverse
  problems involving nonlinear partial differential equations}},\ }\href
  {https://doi.org/10.1016/j.jcp.2018.10.045} {\bibfield  {journal} {\bibinfo
  {journal} {Journal of Computational Physics}\ }\textbf {\bibinfo {volume}
  {378}},\ \bibinfo {pages} {686} (\bibinfo {year} {2019})}\BibitemShut
  {NoStop}%
\bibitem [{\citenamefont {Maclaurin}\ \emph {et~al.}()\citenamefont
  {Maclaurin}, \citenamefont {Duvenaud},\ and\ \citenamefont
  {Adams}}]{maclaurin_autograd_nodate}%
  \BibitemOpen
  \bibfield  {author} {\bibinfo {author} {\bibfnamefont {D.}~\bibnamefont
  {Maclaurin}}, \bibinfo {author} {\bibfnamefont {D.}~\bibnamefont
  {Duvenaud}},\ and\ \bibinfo {author} {\bibfnamefont {R.~P.}\ \bibnamefont
  {Adams}},\ }\bibfield  {title} {{\bibinfo {title}
  {Autograd {Eﬀortless} {Gradients} in {Numpy}}},\ }\href@noop {} {\ ,\
  \bibinfo {pages} {3}}\BibitemShut {NoStop}%
\bibitem [{\citenamefont {Dupont}\ \emph {et~al.}(2019)\citenamefont {Dupont},
  \citenamefont {Doucet},\ and\ \citenamefont {Teh}}]{dupont_augmented_2019}%
  \BibitemOpen
  \bibfield  {author} {\bibinfo {author} {\bibfnamefont {E.}~\bibnamefont
  {Dupont}}, \bibinfo {author} {\bibfnamefont {A.}~\bibnamefont {Doucet}},\
  and\ \bibinfo {author} {\bibfnamefont {Y.~W.}\ \bibnamefont {Teh}},\
  }\bibfield  {title} {\bibinfo {title} {Augmented {Neural} {ODEs}},\ }\href
  {http://arxiv.org/abs/1904.01681} {\bibfield  {journal} {\bibinfo  {journal}
  {arXiv:1904.01681 [cs, stat]}\ } (\bibinfo {year} {2019})},\ \bibinfo {note}
  {arXiv: 1904.01681}\BibitemShut {NoStop}%
\bibitem [{\citenamefont {Zhu}\ \emph {et~al.}(2020)\citenamefont {Zhu},
  \citenamefont {Jin},\ and\ \citenamefont {Tang}}]{zhu_deep_2020}%
  \BibitemOpen
  \bibfield  {author} {\bibinfo {author} {\bibfnamefont {A.}~\bibnamefont
  {Zhu}}, \bibinfo {author} {\bibfnamefont {P.}~\bibnamefont {Jin}},\ and\
  \bibinfo {author} {\bibfnamefont {Y.}~\bibnamefont {Tang}},\ }\bibfield
  {title} {\bibinfo {title} {Deep {Hamiltonian} networks based on symplectic
  integrators},\ }\href {http://arxiv.org/abs/2004.13830} {\bibfield  {journal}
  {\bibinfo  {journal} {arXiv:2004.13830 [cs, math]}\ } (\bibinfo {year}
  {2020})},\ \bibinfo {note} {arXiv: 2004.13830}\BibitemShut {NoStop}%
\bibitem [{\citenamefont {Sanchez-Gonzalez}\ \emph {et~al.}(2018)\citenamefont
  {Sanchez-Gonzalez}, \citenamefont {Heess}, \citenamefont {Springenberg},
  \citenamefont {Merel}, \citenamefont {Riedmiller}, \citenamefont {Hadsell},\
  and\ \citenamefont {Battaglia}}]{sanchez-gonzalez_graph_2018}%
  \BibitemOpen
  \bibfield  {author} {\bibinfo {author} {\bibfnamefont {A.}~\bibnamefont
  {Sanchez-Gonzalez}}, \bibinfo {author} {\bibfnamefont {N.}~\bibnamefont
  {Heess}}, \bibinfo {author} {\bibfnamefont {J.~T.}\ \bibnamefont
  {Springenberg}}, \bibinfo {author} {\bibfnamefont {J.}~\bibnamefont {Merel}},
  \bibinfo {author} {\bibfnamefont {M.}~\bibnamefont {Riedmiller}}, \bibinfo
  {author} {\bibfnamefont {R.}~\bibnamefont {Hadsell}},\ and\ \bibinfo {author}
  {\bibfnamefont {P.}~\bibnamefont {Battaglia}},\ }\bibfield  {title} {\bibinfo
  {title} {Graph networks as learnable physics engines for inference and
  control},\ }\href {http://arxiv.org/abs/1806.01242} {\bibfield  {journal}
  {\bibinfo  {journal} {arXiv:1806.01242 [cs, stat]}\ } (\bibinfo {year}
  {2018})},\ \bibinfo {note} {arXiv: 1806.01242}\BibitemShut {NoStop}%
\bibitem [{\citenamefont {Sanchez-Gonzalez}\ \emph {et~al.}(2020)\citenamefont
  {Sanchez-Gonzalez}, \citenamefont {Godwin}, \citenamefont {Pfaff},
  \citenamefont {Ying}, \citenamefont {Leskovec},\ and\ \citenamefont
  {Battaglia}}]{sanchez-gonzalez_learning_2020}%
  \BibitemOpen
  \bibfield  {author} {\bibinfo {author} {\bibfnamefont {A.}~\bibnamefont
  {Sanchez-Gonzalez}}, \bibinfo {author} {\bibfnamefont {J.}~\bibnamefont
  {Godwin}}, \bibinfo {author} {\bibfnamefont {T.}~\bibnamefont {Pfaff}},
  \bibinfo {author} {\bibfnamefont {R.}~\bibnamefont {Ying}}, \bibinfo {author}
  {\bibfnamefont {J.}~\bibnamefont {Leskovec}},\ and\ \bibinfo {author}
  {\bibfnamefont {P.~W.}\ \bibnamefont {Battaglia}},\ }\bibfield  {title}
  {\bibinfo {title} {Learning to {Simulate} {Complex} {Physics} with {Graph}
  {Networks}},\ }\href {http://arxiv.org/abs/2002.09405} {\bibfield  {journal}
  {\bibinfo  {journal} {arXiv:2002.09405 [physics, stat]}\ } (\bibinfo {year}
  {2020})},\ \bibinfo {note} {arXiv: 2002.09405}\BibitemShut {NoStop}%
\bibitem [{\citenamefont {Finzi}\ \emph {et~al.}(2020)\citenamefont {Finzi},
  \citenamefont {Stanton}, \citenamefont {Izmailov},\ and\ \citenamefont
  {Wilson}}]{finzi_generalizing_2020}%
  \BibitemOpen
  \bibfield  {author} {\bibinfo {author} {\bibfnamefont {M.}~\bibnamefont
  {Finzi}}, \bibinfo {author} {\bibfnamefont {S.}~\bibnamefont {Stanton}},
  \bibinfo {author} {\bibfnamefont {P.}~\bibnamefont {Izmailov}},\ and\
  \bibinfo {author} {\bibfnamefont {A.~G.}\ \bibnamefont {Wilson}},\ }\bibfield
   {title} {\bibinfo {title} {Generalizing {Convolutional} {Neural} {Networks}
  for {Equivariance} to {Lie} {Groups} on {Arbitrary} {Continuous} {Data}},\
  }\href {http://arxiv.org/abs/2002.12880} {\bibfield  {journal} {\bibinfo
  {journal} {arXiv:2002.12880 [cs, stat]}\ } (\bibinfo {year} {2020})},\
  \bibinfo {note} {arXiv: 2002.12880}\BibitemShut {NoStop}%
\end{thebibliography}
%

\clearpage

\end{document}


\appendix
\section{Hyperparameters}

\begin{figure}[!htb]
\centering
\includegraphics[width=.45\textwidth, height=5cm]{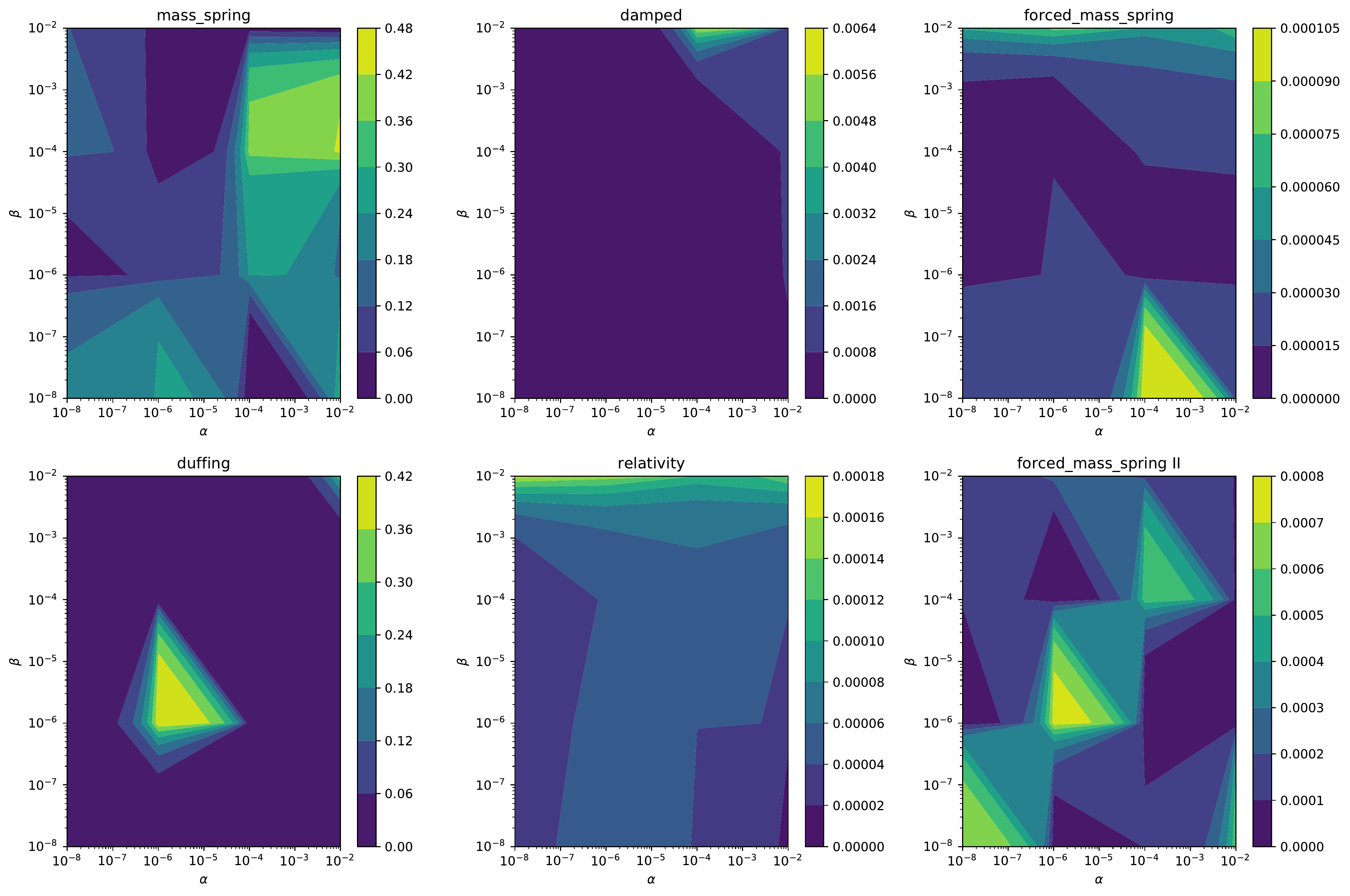}
\caption{Hyperparameter Optimization for pHNN. We plot, for each system, the validation loss as a function of the $\lambda_F$ and $\lambda_N$ parameters from the loss in eqn. 5}
\end{figure}

The choice of the regularization coefficients $\lambda_F$ and $\lambda_{N}$ was made via a grid search across the log-space $[10^{-2},10^{-4},10^{-6},10^{-8}]$, for each hyper-parameter, that generates the lowest validation loss.

\section{Pipeline}
The detailed procedure of the pipeline for the pHNN model is as follows:
\begin{enumerate}
\item Obtain ground truth state variable data $[\mathbf{q},\mathbf{p},t]$ and time derivatives $[\dot{\mathbf{q}},\dot{\mathbf{p}}]$ from trajectories of a certain system.
\item Provide state-variable information to pHNN which learns a Hamiltonian, force and damping term to predict $[\hat{\dot{\mathbf{q}}},\hat{\dot{\mathbf{p}}}]$.
\item Optimize pHNN by minimizing the loss function of Eq. (5).
\item Once trained, use the pHNN in a scientific integrator to evolve a set of random initial conditions in the test set.
\item Assess the performance by calculating the MSE functions of Eqs. (6) and (7).
\end{enumerate}

\section{Damped Hamiltonians}
Why can we not including damping into the Hamiltonian? Let us take the following damped system where $\delta$ is the damping coefficient:
\begin{equation}
\ddot{\mathbf{q}} = -\mathbf{q} - \delta\dot{\mathbf{q}}
\end{equation}
where we know $\mathbf{p}=m\dot{\mathbf{q}}$ which implies $\dot{\mathbf{q}} = m^{-1}\mathbf{p}$.

Then, the integral of the right hand side with respect to q will give us:

\begin{equation}
\frac{\mathbf{q}^T \mathbf{q}}{2} + \delta \mathbf{q}^T \dot{\mathbf{q}}
\end{equation}

The equation above looks like a modified potential function which can be combined with a kinetic energy term to give a Hamiltonian s.t.:

\begin{equation}
\mathcal{H} =\frac{ \mathbf{p}^T \mathbf{p}}{2m} + \frac{\mathbf{q}^T \mathbf{q}}{2} + \delta \mathbf{q}^T \dot{\mathbf{q}}
\end{equation}

However, although we can recover the differential equation for $\ddot{\mathbf{q}}$ by $-\frac{\partial\mathcal{H}}{\mathrm{d}\mathbf{q}}  =-\mathbf{q} - \delta\dot{\mathbf{q}} = \ddot{\mathbf{q}}$, we violate the rule that $\dot{\mathbf{q}} = m^{-1}\mathbf{p}$ since $ \frac{\partial\mathcal{H}}{\mathrm{d}\mathbf{p}} =  \dot{\mathbf{q}} +\delta \mathbf{q} \neq \dot{\mathbf{q}} $.


\clearpage
\section{Results}

For each system we illustrate the rollout of a single initial condition in the test set and record the MSE as a function of time.

\begin{figure}[!htb]
\centering
\captionsetup{justification=centering}
\begin{subfigure}[b]{0.48\textwidth}
\includegraphics[width=\textwidth]{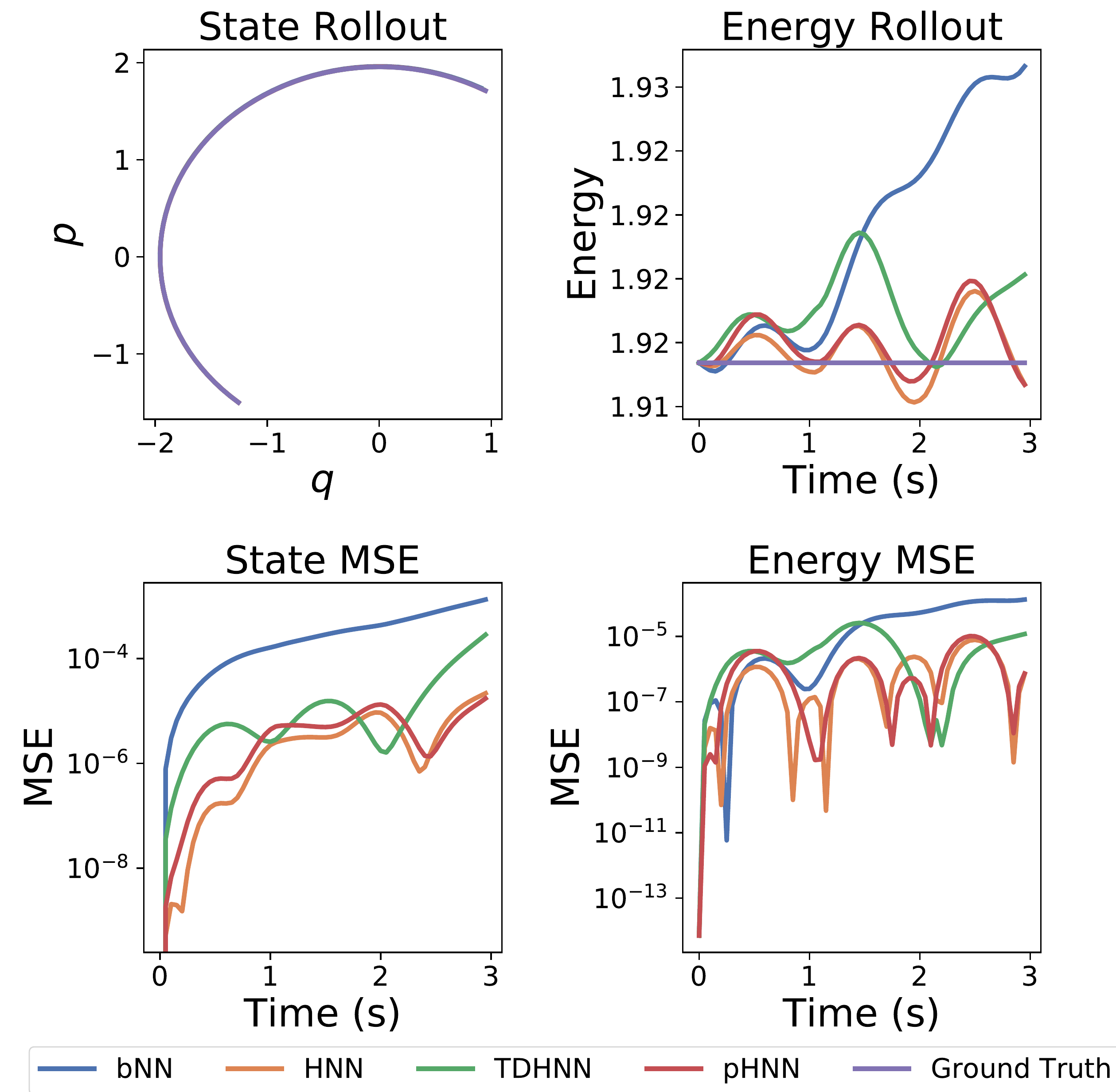}
\caption{State and energy rollout of an initial condition from the test set}
\end{subfigure}
\begin{subfigure}[b]{0.48\textwidth}
\includegraphics[width=\textwidth]{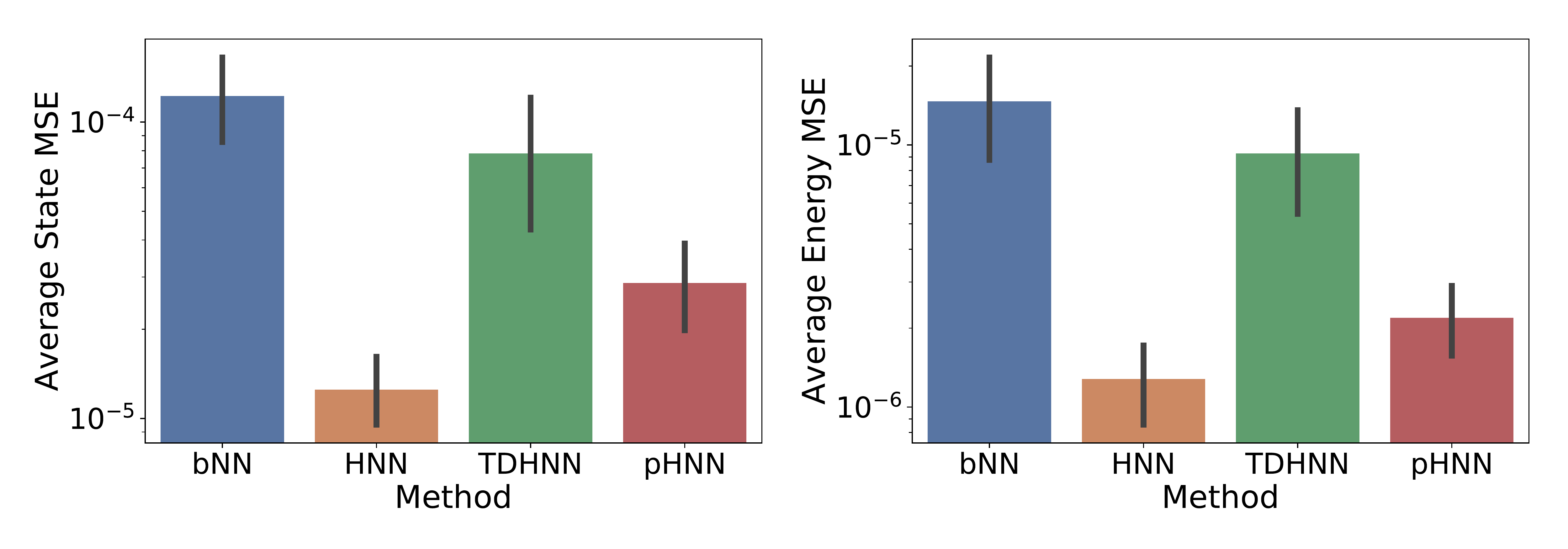}
\caption{The average state and energy MSE across 25 test points}
\end{subfigure}
\begin{subfigure}[b]{0.48\textwidth}
\includegraphics[width=\textwidth]{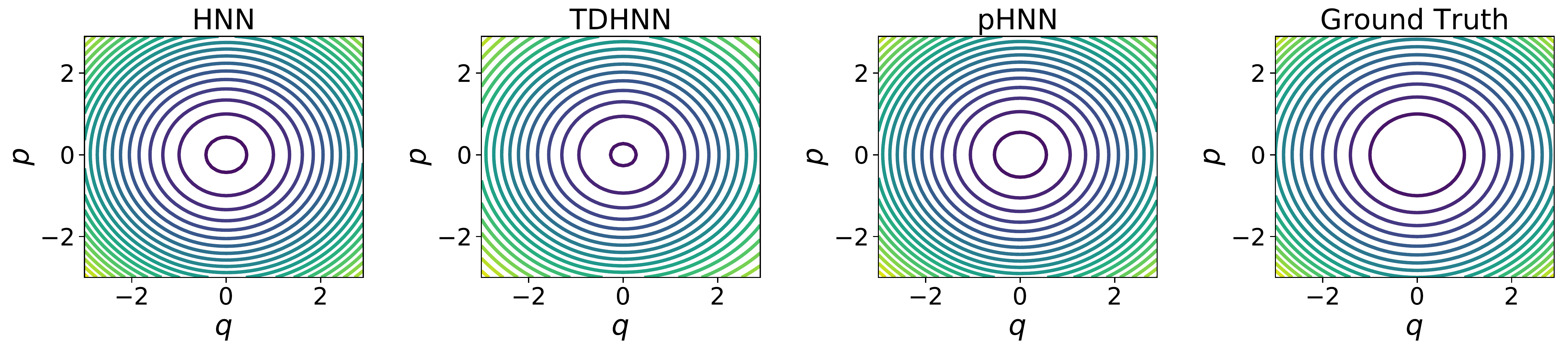}
\caption{The learnt Hamiltonian across methods}
\end{subfigure}
\begin{subfigure}[b]{0.48\textwidth}
\includegraphics[width=\textwidth]{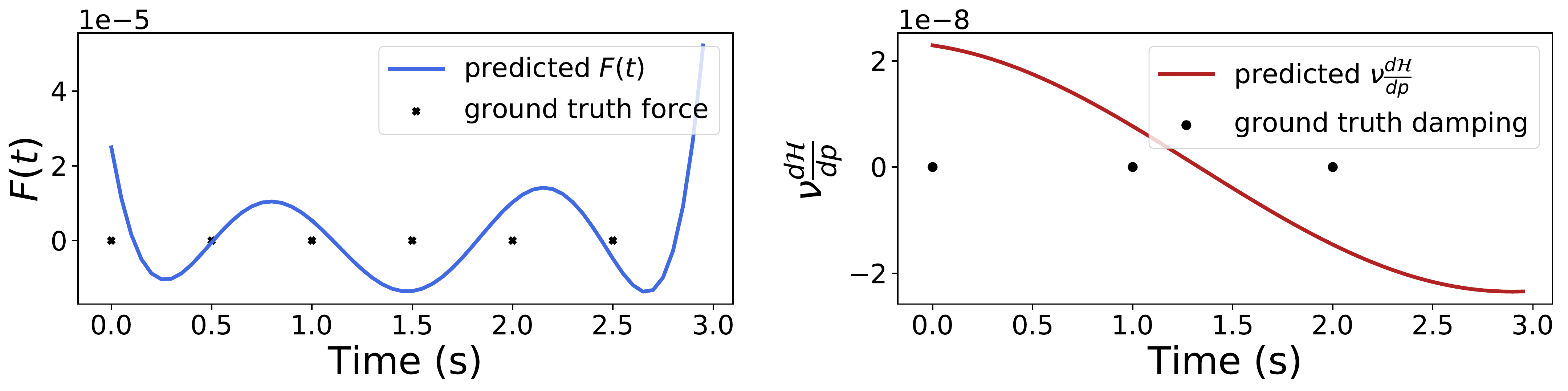}
\caption{The learnt force and damping of pHNN}
\end{subfigure}
\caption{Mass-Spring System}
\end{figure}
\begin{figure}[!htb]
\centering
\captionsetup{justification=centering}
\begin{subfigure}[b]{0.48\textwidth}
\includegraphics[width=\textwidth]{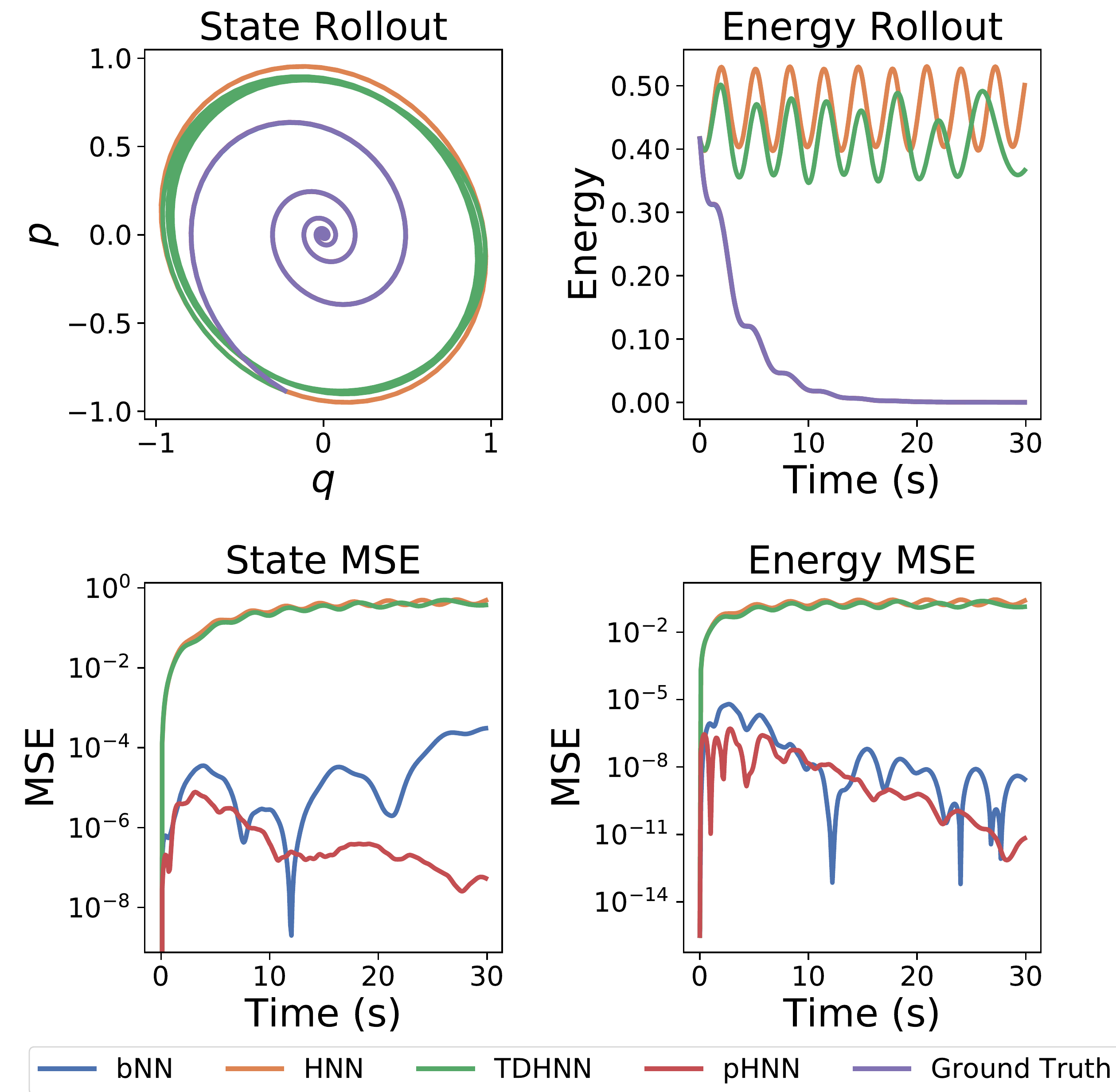}
\caption{State and energy rollout of an initial condition from the test set}
\end{subfigure}
\begin{subfigure}[b]{0.48\textwidth}
\includegraphics[width=\textwidth]{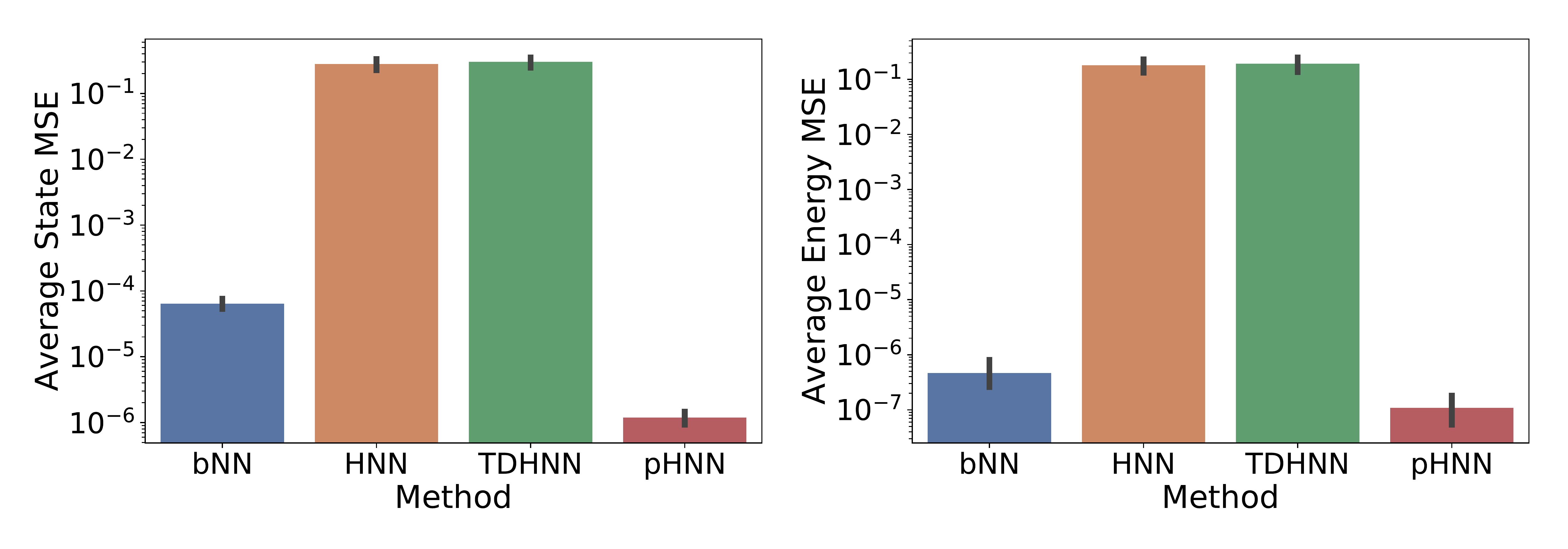}
\caption{The average state and energy MSE across 25 test points}
\end{subfigure}
\begin{subfigure}[b]{0.48\textwidth}
\includegraphics[width=\textwidth]{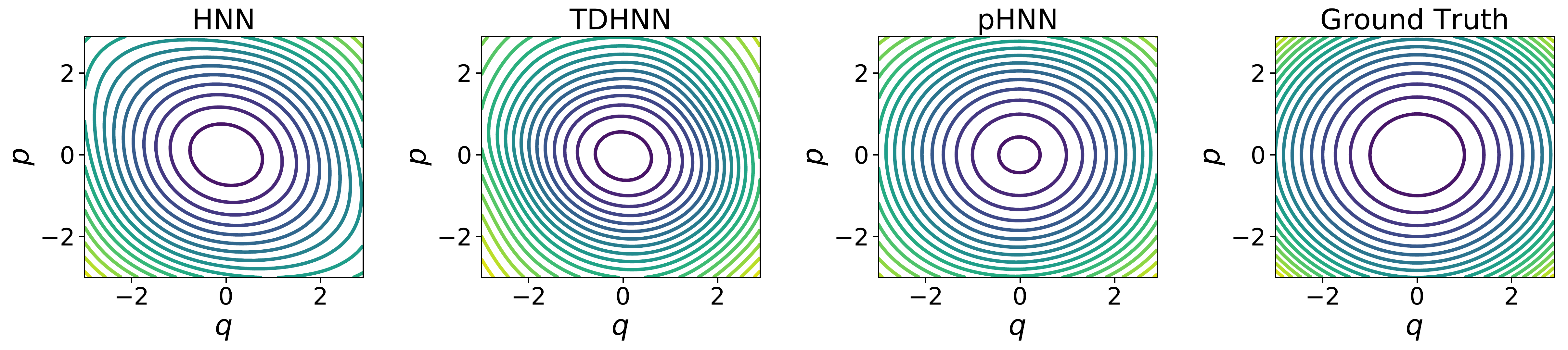}
\caption{The learnt Hamiltonian across methods}
\end{subfigure}
\begin{subfigure}[b]{0.48\textwidth}
\includegraphics[width=\textwidth]{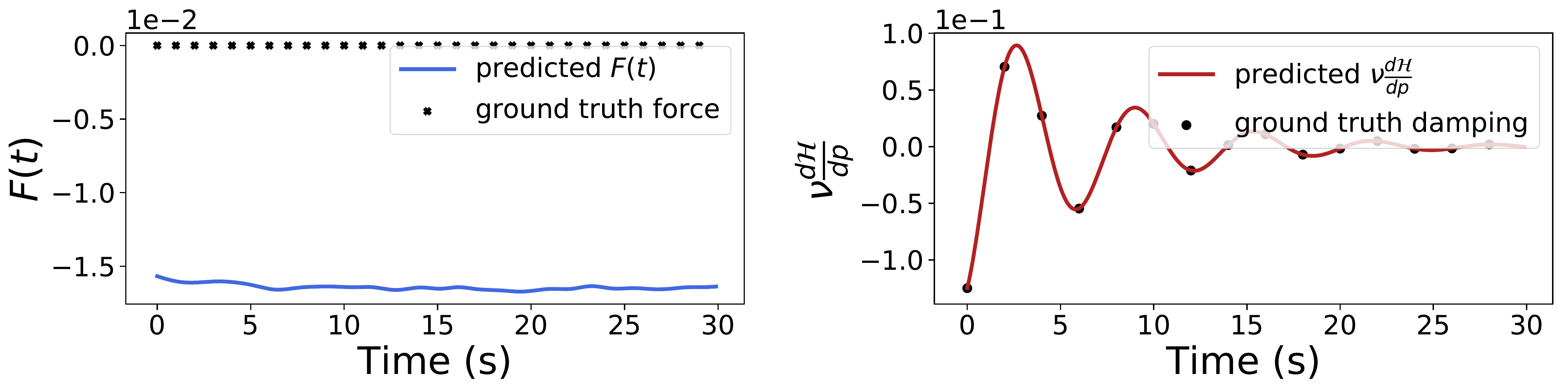}
\caption{The learnt force and damping of pHNN}
\end{subfigure}
\caption{Damped Mass-Spring System}
\end{figure}
\begin{figure}[!htb]
\centering
\captionsetup{justification=centering}
\begin{subfigure}[b]{0.48\textwidth}
\includegraphics[width=\textwidth]{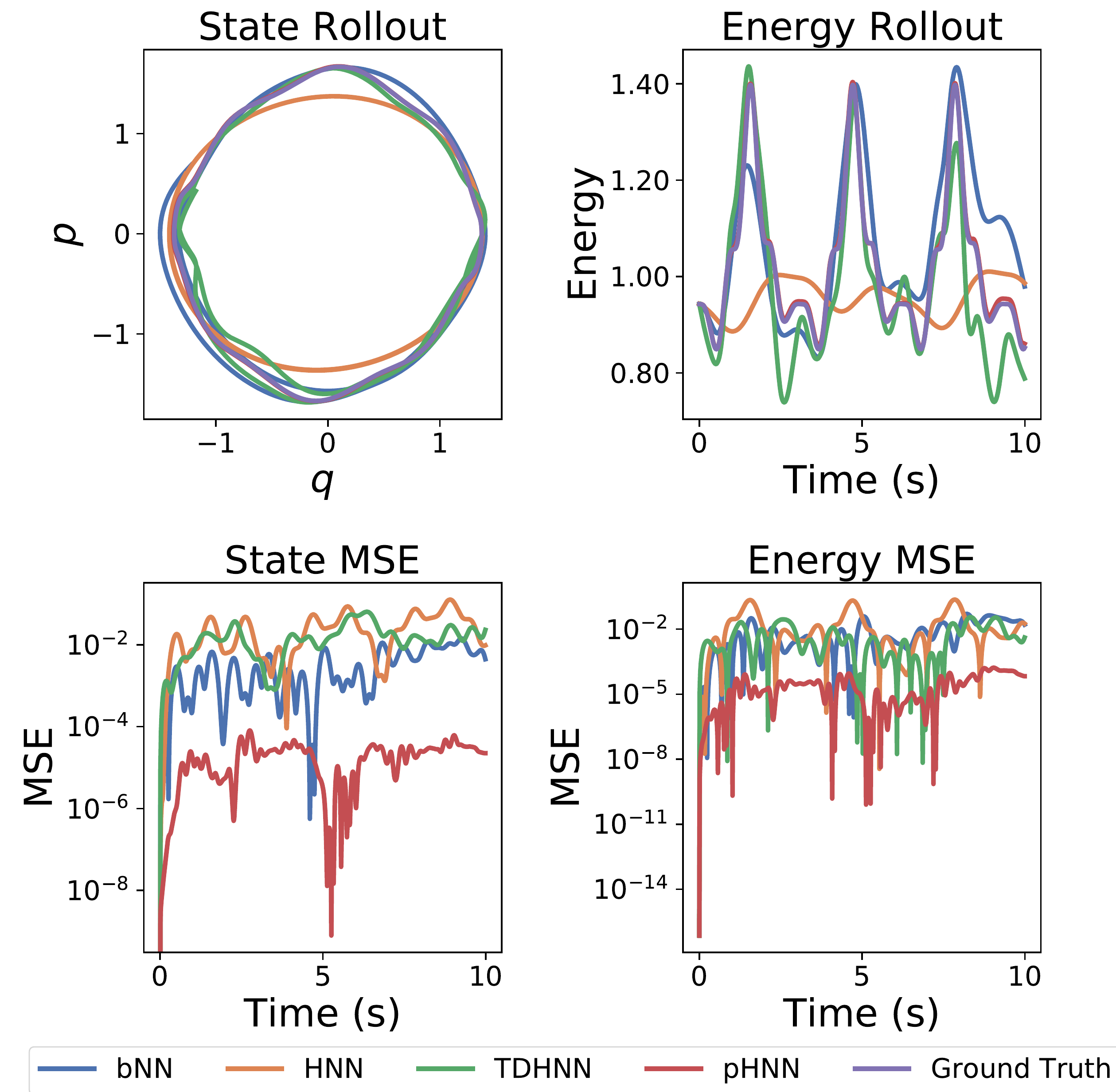}
\caption{State and energy rollout of an initial condition from the test set}
\end{subfigure}
\begin{subfigure}[b]{0.48\textwidth}
\includegraphics[width=\textwidth]{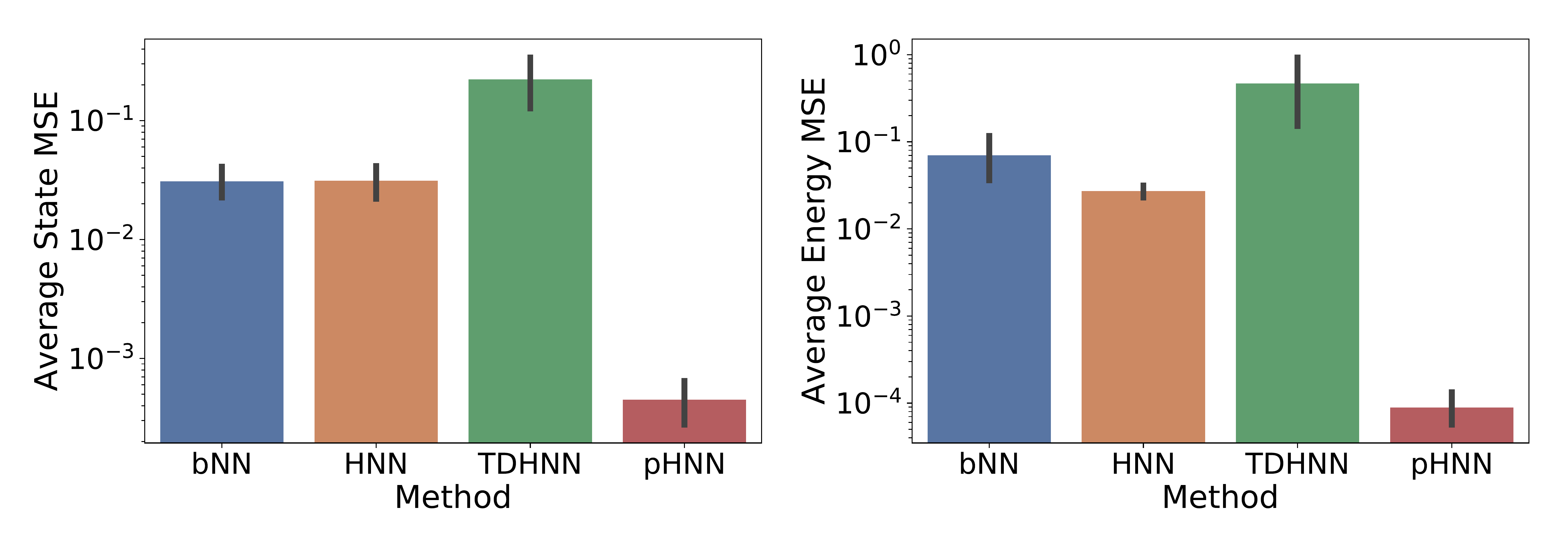}
\caption{The average state and energy MSE across 25 test points}
\end{subfigure}
\begin{subfigure}[b]{0.48\textwidth}
\includegraphics[width=\textwidth]{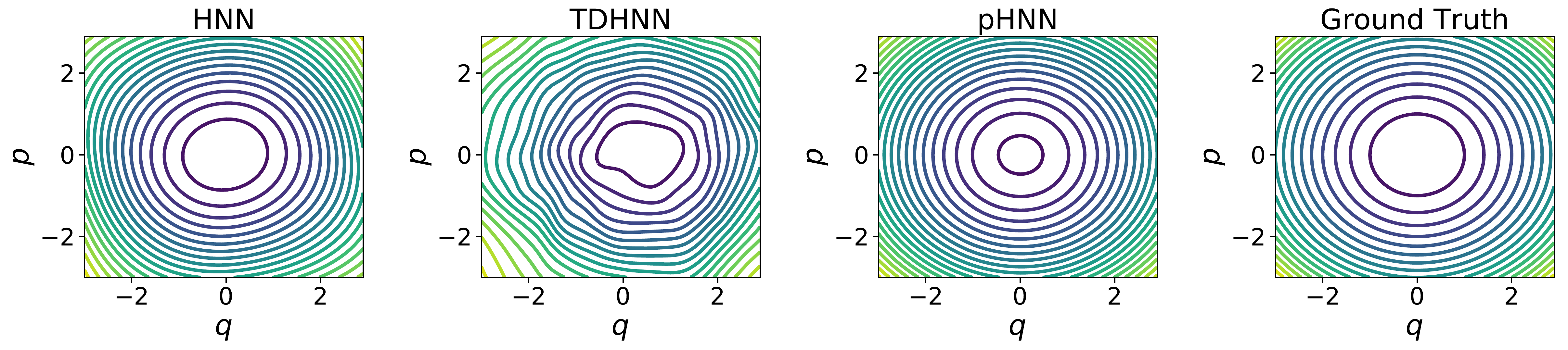}
\caption{The learnt Hamiltonian across methods}
\end{subfigure}
\begin{subfigure}[b]{0.48\textwidth}
\includegraphics[width=\textwidth]{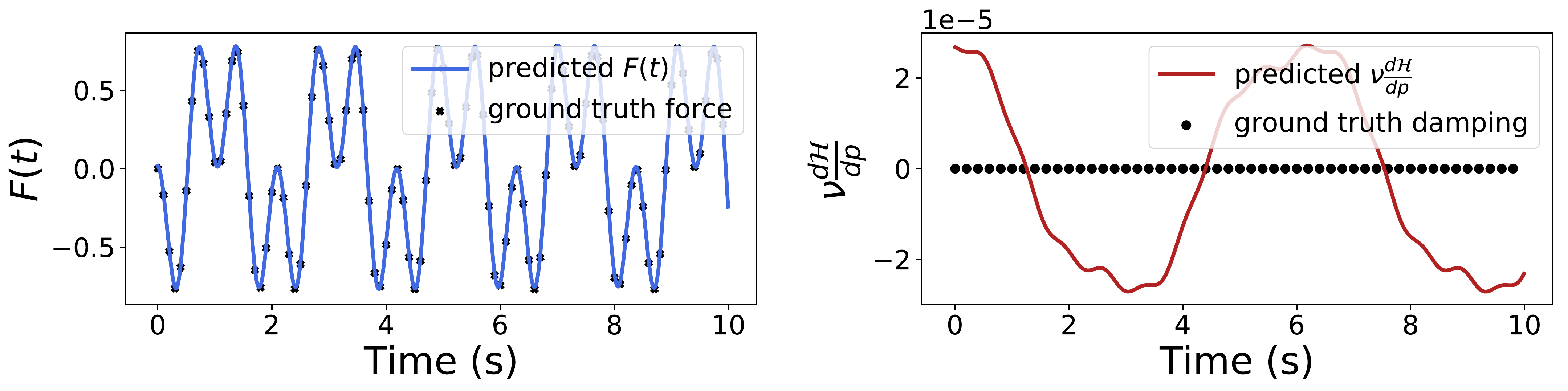}
\caption{The learnt force and damping of pHNN}
\end{subfigure}
\caption{Forced Mass-Spring System (I)}
\end{figure}
\begin{figure}[!htb]
\centering
\captionsetup{justification=centering}
\begin{subfigure}[b]{0.48\textwidth}
\includegraphics[width=\textwidth]{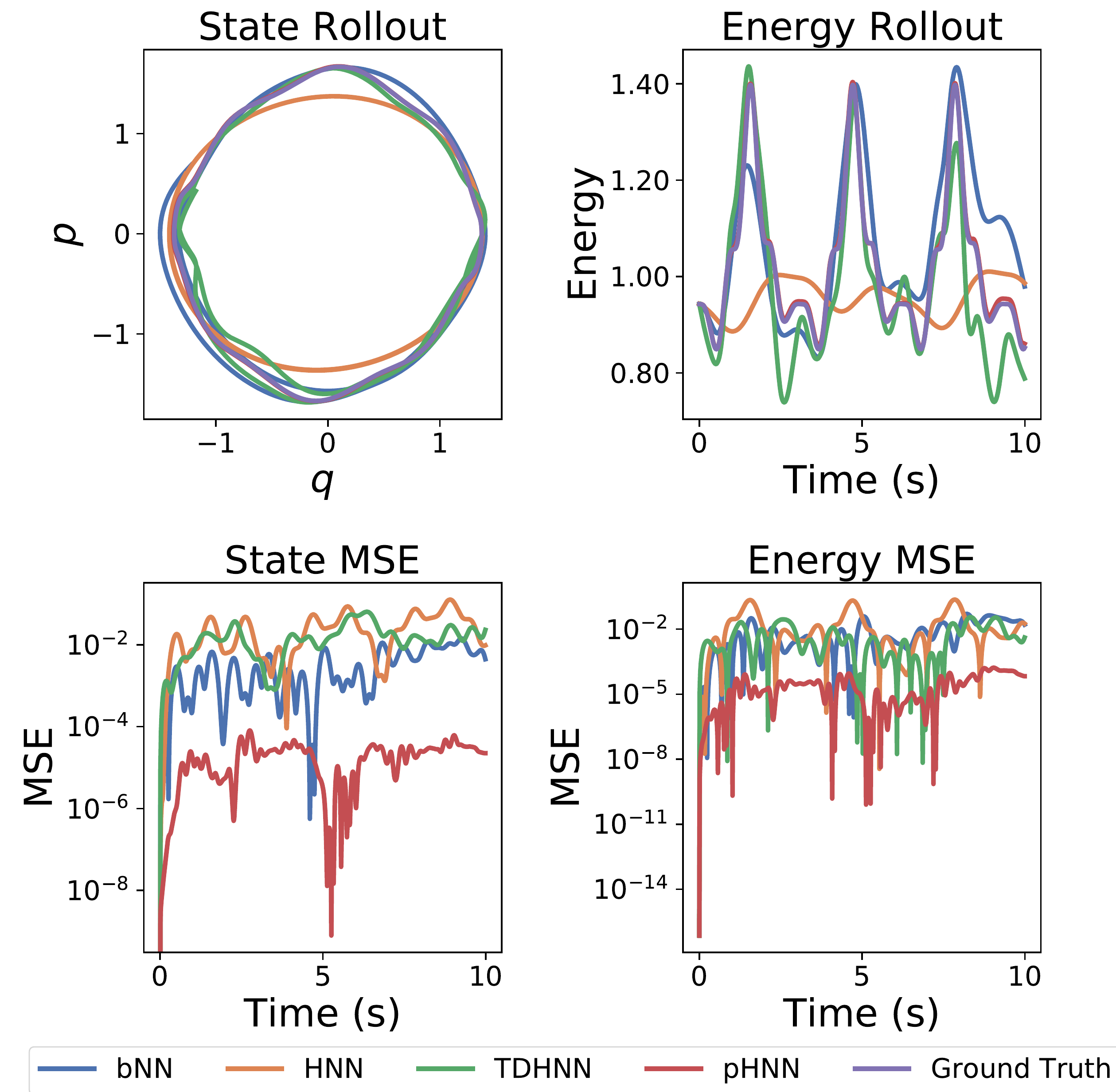}
\caption{State and energy rollout of an initial condition from the test set}
\end{subfigure}
\begin{subfigure}[b]{0.48\textwidth}
\includegraphics[width=\textwidth]{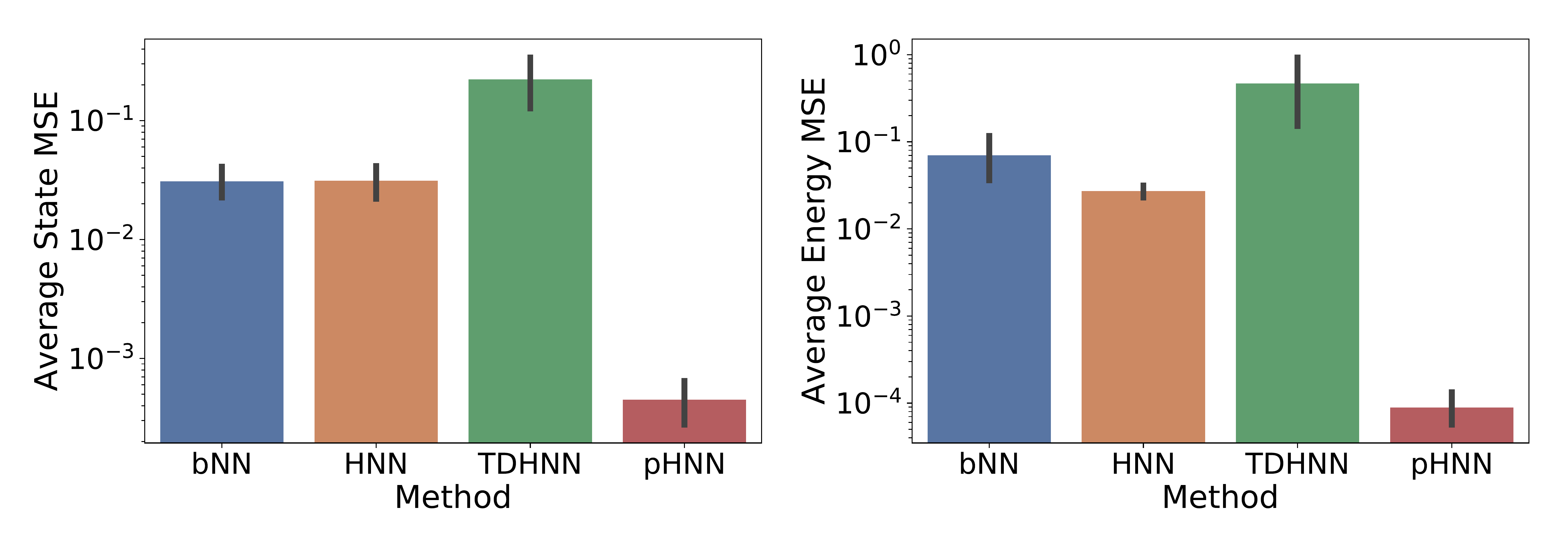}
\caption{The average state and energy MSE across 25 test points}
\end{subfigure}
\begin{subfigure}[b]{0.48\textwidth}
\includegraphics[width=\textwidth]{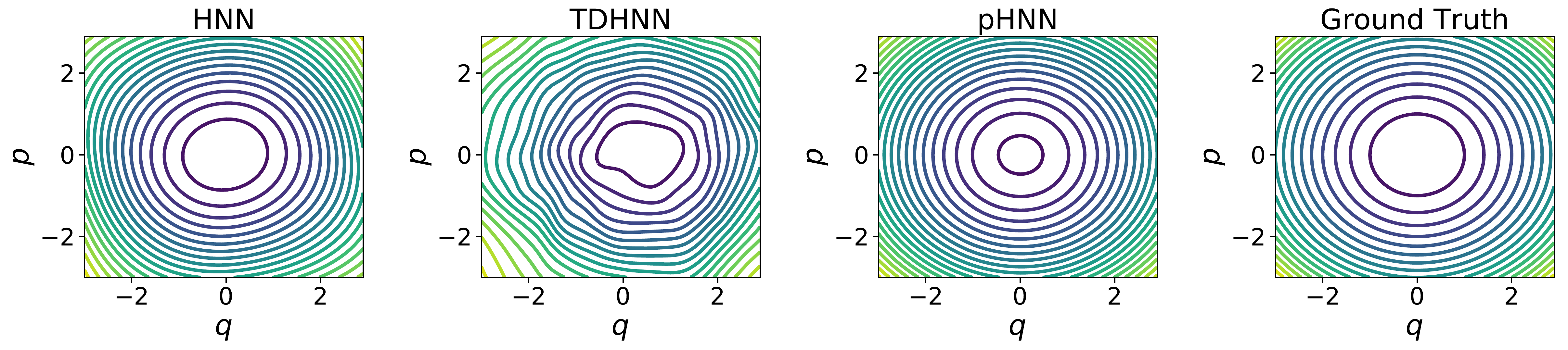}
\caption{The learnt Hamiltonian across methods}
\end{subfigure}
\begin{subfigure}[b]{0.48\textwidth}
\includegraphics[width=\textwidth]{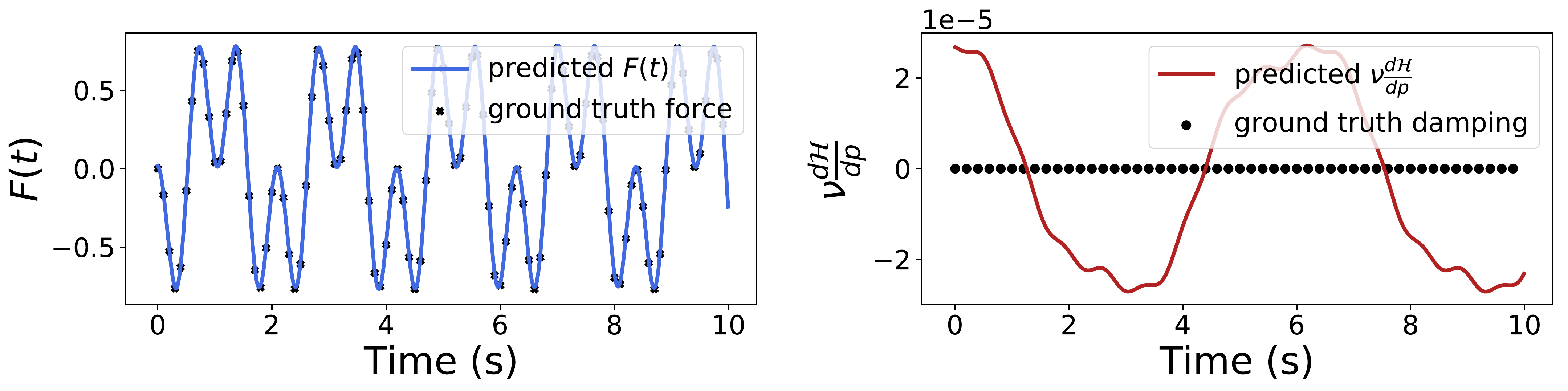}
\caption{The learnt force and damping of pHNN}
\end{subfigure}
\caption{Forced Mass-Spring System (II)}
\end{figure}
\begin{figure}[!htb]
\centering
\captionsetup{justification=centering}
\begin{subfigure}[b]{0.48\textwidth}
\includegraphics[width=\textwidth]{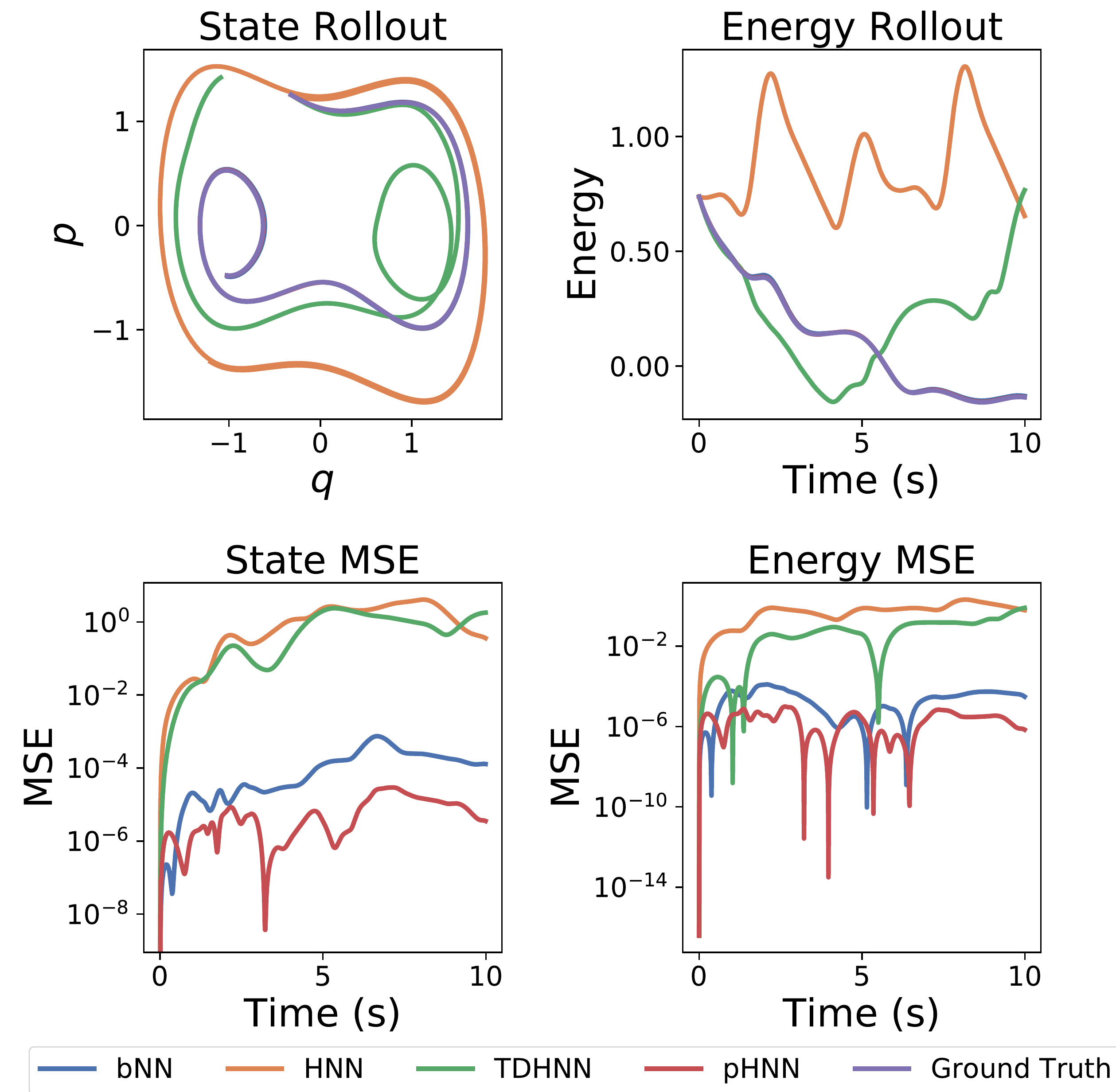}
\caption{State and energy rollout of an initial condition from the test set}
\end{subfigure}
\begin{subfigure}[b]{0.48\textwidth}
\includegraphics[width=\textwidth]{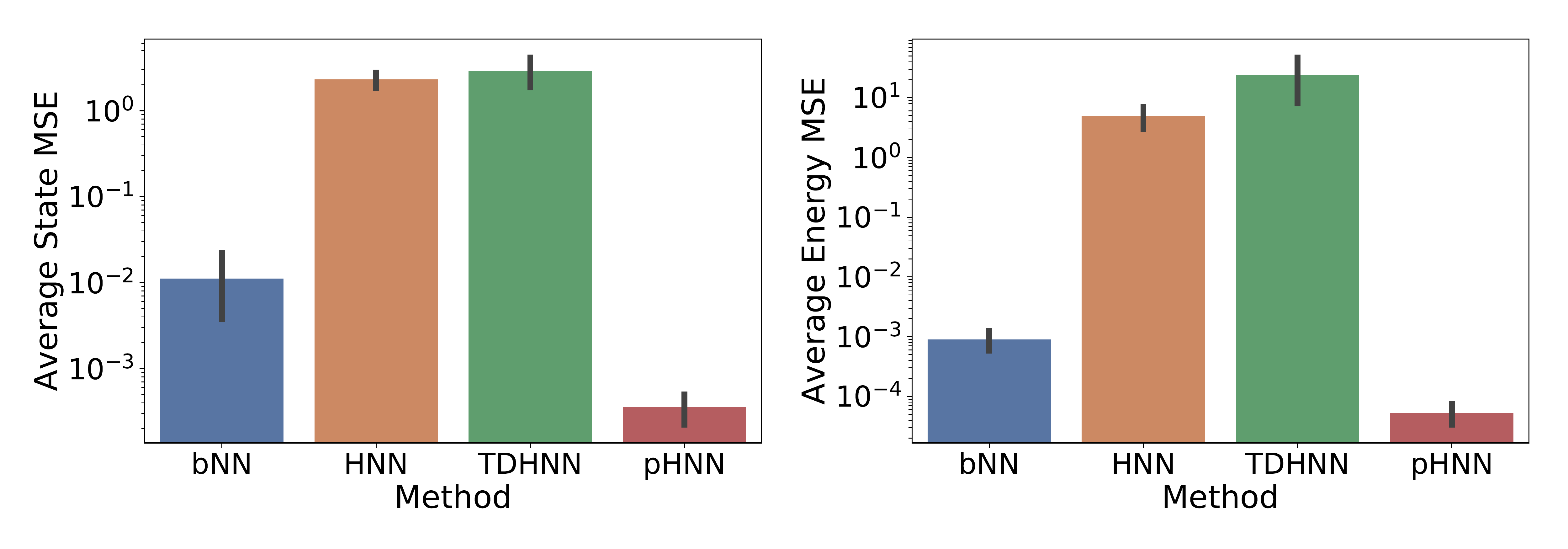}
\caption{The average state and energy MSE across 25 test points}
\end{subfigure}
\begin{subfigure}[b]{0.48\textwidth}
\includegraphics[width=\textwidth]{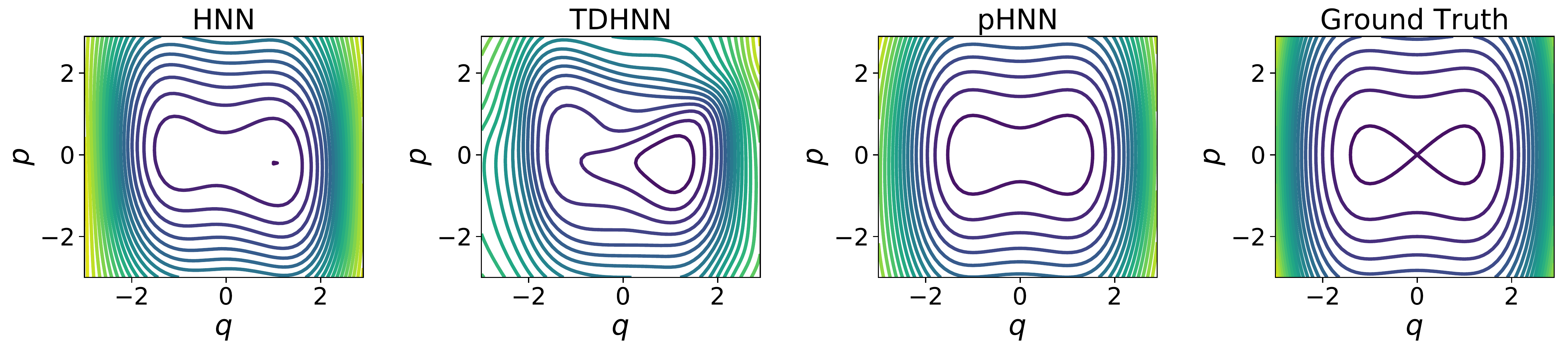}
\caption{The learnt Hamiltonian across methods}
\end{subfigure}
\begin{subfigure}[b]{0.48\textwidth}
\includegraphics[width=\textwidth]{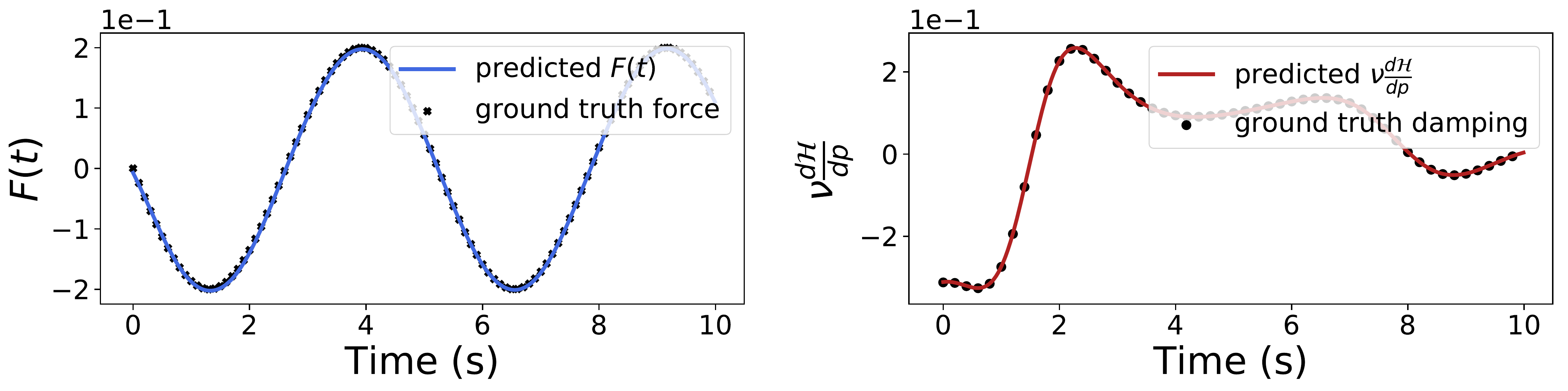}
\caption{The learnt force and damping of pHNN}
\end{subfigure}
\caption{Non-Chaotic Duffing Equation}
\end{figure}
\begin{figure}[!htb]
\centering
\captionsetup{justification=centering}
\begin{subfigure}[b]{0.48\textwidth}
\includegraphics[width=\textwidth]{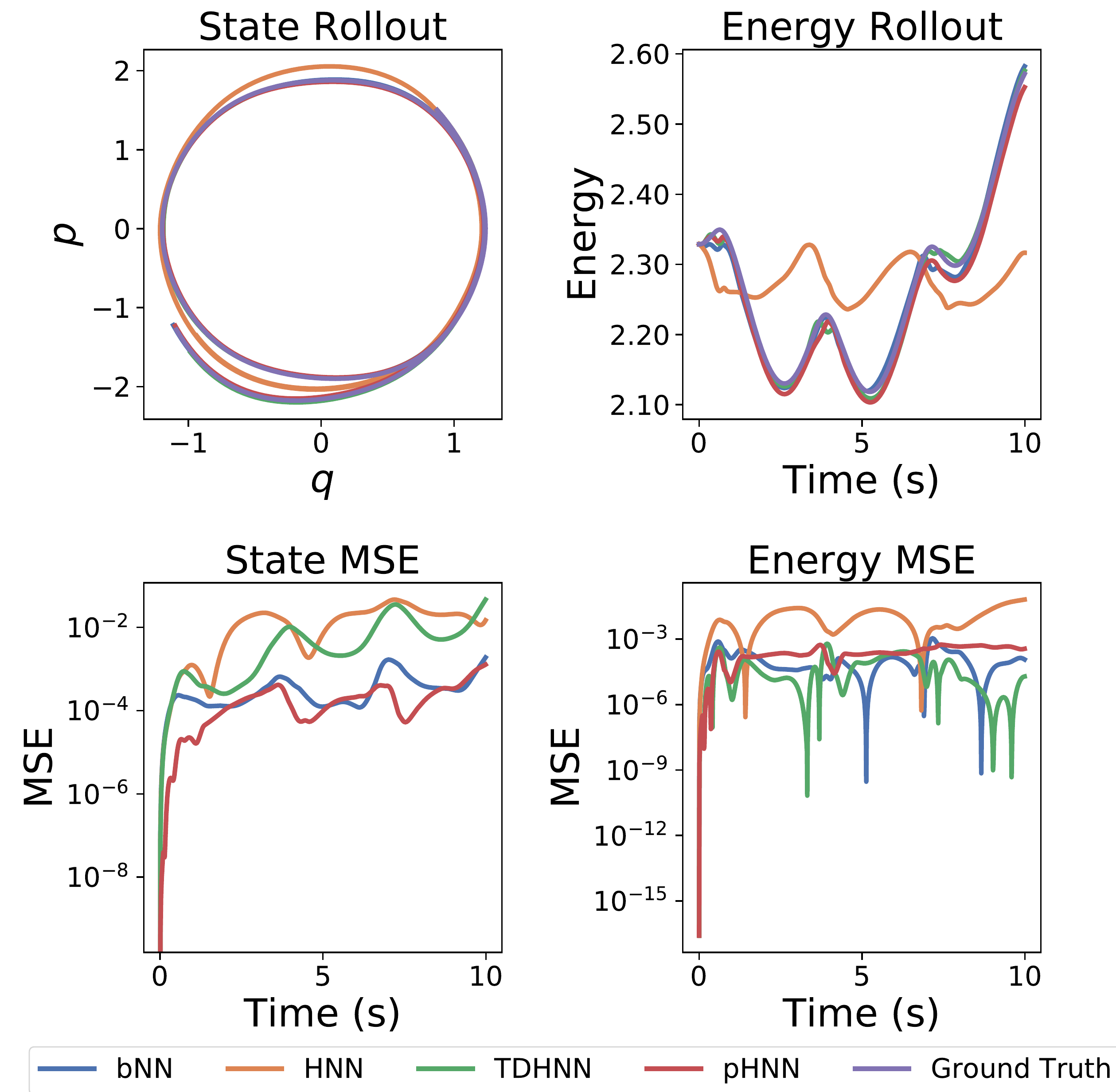}
\caption{State and energy rollout of an initial condition from the test set}
\end{subfigure}
\begin{subfigure}[b]{0.48\textwidth}
\includegraphics[width=\textwidth]{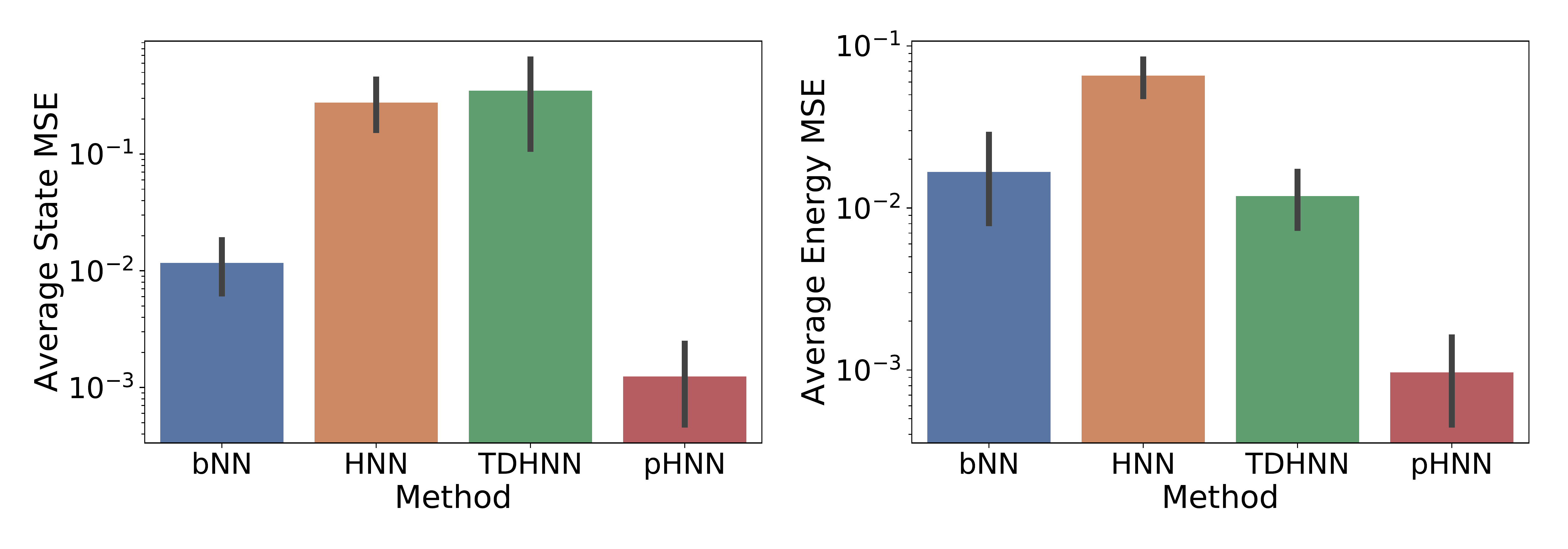}
\caption{The average state and energy MSE across 25 test points}
\end{subfigure}
\begin{subfigure}[b]{0.48\textwidth}
\includegraphics[width=\textwidth]{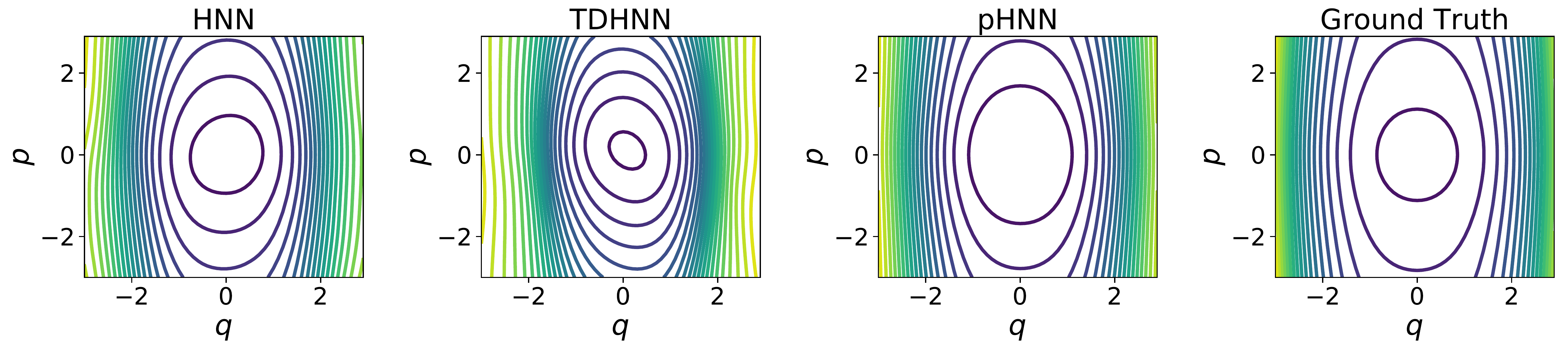}
\caption{The learnt Hamiltonian across methods}
\end{subfigure}
\begin{subfigure}[b]{0.48\textwidth}
\includegraphics[width=\textwidth]{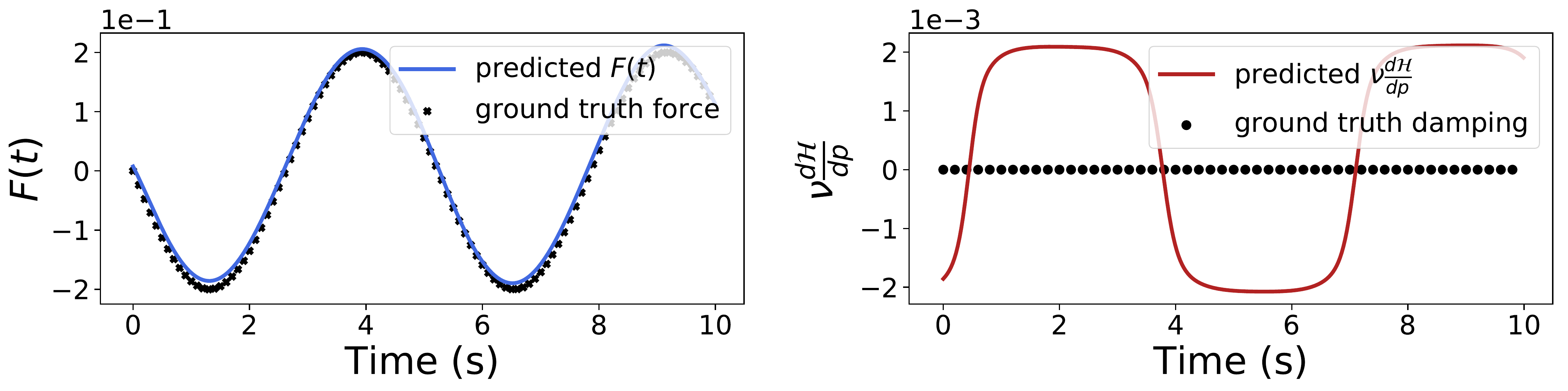}
\caption{The learnt force and damping of pHNN}
\end{subfigure}
\caption{Relativistic Duffing}
\end{figure}

\clearpage

\section{Tabularized Results}

\begin{table*}[htbp]
\label{tab.table1}
\centering
\caption{State and Energy MSE for each method averaged across 25 initial test points} \label{tab:1}
\resizebox{\textwidth}{!}{%
    \begin{tabular}{|l|c|c|c|c|c|c|c|c|}
    \toprule
          & \multicolumn{2}{c|}{\textbf{Baseline}} & \multicolumn{2}{c|}{\textbf{HNN}} & \multicolumn{2}{c|}{\textbf{TDHNN}} & \multicolumn{2}{c|}{\textbf{pHNN}} \\
\cmidrule{2-9}          & \textbf{State} & \textbf{Energy} & \textbf{State} & \textbf{Energy} & \textbf{State} & \textbf{Energy} & \textbf{State} & \textbf{Energy} \\
\cmidrule{2-9}    \textbf{Mass Spring} & 1.22E-4 $\pm$ 1.06E-4 & 1.56E-5 $\pm$ 1.84 E-5 & \textbf{1.33E-5 $\pm$ 8.05E-6} & \textbf{1.30E-6 $\pm$ 1.31E-6} & 7.84E-5$\pm$1.02E-4 & 9.26E-6 $\pm$ 1.13E-5 & 2.91E-5 $\pm$ 2.57E-5 & 2.33E-6 $\pm$ 2.41E-6 \\
    \textbf{Damped Mass-Spring} & 6.4E-5 $\pm$ 2.904E-5 & 4.68E-7 $\pm$ 7.51E-7 & 2.81E-1$\pm$ 1.37E-1 & 1.81E-1 $\pm$ 1.37E-1 & 3.01E-1$\pm$1.43E-1 & 1.92E-1$\pm$1.42E-1 & \textbf{1.74E-6$\pm$7.26E-7} & \textbf{1.08E-7$\pm$1.73E-7} \\
    \textbf{Forced (I)} & 5.21E-4 $\pm$ 5.74E-4 & 7.77E-4 $\pm$ 5.52E-4 & 2.36E-1 $\pm$ 1.43E-1 & 2.34E-1 $\pm$ 1.70E-1 & 2.91E-1 $\pm$ 3.06E-1 & 7.61E-1 $\pm$ 1.23 & \textbf{1.53E-5 $\pm$ 1.16E-5} & \textbf{9.21E-6 $\pm$ 7.11E-6} \\
    \textbf{Forced (II)} & 3.14E-2 $\pm$ 2.54E-2 & 7.00E-2 $\pm$ 1.14E-1 & 3.11E-2 $\pm$ 2.67E-2 & 2.70E-2 $\pm$ 1.12E-2 & 2.22E-1 $\pm$ 2.82E-1 & 4.68E-1 $\pm$ 1.06 & \textbf{4.51E-4 $\pm$ 4.85E-4} & \textbf{8.91E-5 $\pm$ 1.01E-4} \\
    \textbf{Duffing} & 1.11E-2 $\pm$ 2.43E-2 & 8.93E-4 $\pm$ 8.75E-4 & 2.31 $\pm$ 1.21 & 4.90 $\pm$ 5.30 & 2.91 $\pm$ 3.01 & 2.43E1 $\pm$ 5.11E1 & \textbf{3.50E-4 $\pm$3.54E-4} & \textbf{5.32E-5 $\pm$ 5.38E-5} \\
    \textbf{Relativistic Duffing} & 5.85E-2 $\pm$ 7.05E-2 & 3.61E-2 $\pm$ 4.76E-2 & 1.39 $\pm$ 1.66 & 2.37E-1 $\pm$ 1.55E-1 & 5.9E-1 $\pm$ 1.44 & 4.39E-2 $\pm$ 5.34E-2 & \textbf{4.96E-3 $\pm$ 8.61E-3} & \textbf{1.28E-3$\pm$1.45E-3} \\
    \bottomrule
\end{tabular}
}%
\end{table*}

\begin{table*}[htbp]
\label{tab.table12}
\centering
\caption{State and Energy MSE for each method embedded with an RK4 integrator averaged across 25 initial test points} \label{tab:1}
\resizebox{\textwidth}{!}{%
\begin{tabular}{|l|c|c|c|c|c|c|c|c|}
    \toprule
          & \multicolumn{2}{c|}{\textbf{Baseline}} & \multicolumn{2}{c|}{\textbf{HNN}} & \multicolumn{2}{c|}{\textbf{TDHNN}} & \multicolumn{2}{c|}{\textbf{pHNN}} \\
\cmidrule{2-9}          & \textbf{State} & \textbf{Energy} & \textbf{State} & \textbf{Energy} & \textbf{State} & \textbf{Energy} & \textbf{State} & \textbf{Energy} \\
\cmidrule{2-9}    \textbf{Mass Spring} & 1.18E-4 ± 1.01E-4 & 1.75E-5 ± 1.90E-5 & \textbf{1.96E-5 ± 1.98E-5} & \textbf{1.77E-6 ± 1.79E-6} & 2.78E-5 ± 2.95E-5 & 1.56E-5 ± 1.79E-5 & 2.78E-5 ± 2.65E-5 & 2.69E-6 ± 2.76E-6 \\
    \textbf{Damped Mass-Spring} & \multicolumn{1}{l|}{\textbf{1.95E-4 ± 8.77E-5}} & \multicolumn{1}{l|}{\textbf{8.11E-7 ± 7.49E-7}} & \multicolumn{1}{l|}{2.80E-1 ± 1.49E-1} & \multicolumn{1}{l|}{1.81E-1 ± 1.36E-1} & \multicolumn{1}{l|}{3.06E-1 ± 1.46E-1} & \multicolumn{1}{l|}{1.94E-1 ± 1.43E-1} & \multicolumn{1}{l|}{2.70E-3 ± 1.34E-3} & \multicolumn{1}{l|}{7.37E-4 ± 5.85E-4} \\
    \textbf{Forced (I)} & 6.82E-4 ± 8.30E-4 & 1.31E-3 ± 4.25E-3 & 1.84E-1 ± 1.15E-1 & 2.27E-1 ± 1.66E-1 & 7.62E-4 ± 5.39E-4 & 8.23E-4 ± 6.17E-4 & \textbf{3.45E-5 ± 2.32E-5} & \textbf{8.54E-6 ± 8.27E-6} \\
    \textbf{Forced (II)} & 2.49E-2 ± 1.29e-2 & 3.52E-2 ± 1.12e-2 & 2.80E-2 ± 2.18E-2 & 2.76E-2 ± 1.13E-2 & 5.01E-2 ± 5.11E-2 & 8.02E-2 ± 5.83E-2 & \textbf{1.33E-3 ± 1.64E-3} & \textbf{1.63E-4 ± 1.78E-4} \\
    \textbf{Duffing} & 5.41E-2 ± 1.47E-1 & 2.41E-3 ± 2.55E-3 & 2.29E0 ± 1.19E0 & 4.54E0 ± 4.58E0 & 1.84E0 ± 1.02E0 & 5.11E0 ± 5.58E0 & \textbf{3.57E-3 ± 2.99E-3} & \textbf{7.10E-4 ± 1.17E-3} \\
    \textbf{Relativistic Duffing} & 1.70E-2 ± 1.79E-2 & 2.16E-2 ± 3.04E-2 & 1.91E-1 ± 1.59E-1 & 5.83E-2 ± 4.19E-2 & 3.18E-2 ± 2.20E-2 & 4.83E-2 ± 4.62E-2 & \textbf{1.18E-3 ± 1.95E-3} & \textbf{6.09E-4 ± 1.08E-3} \\
    \bottomrule
    \end{tabular}%
    }%
\end{table*}%


\section{Results with Noise}
During the training of HNN, the authors add gaussian noise with a standard deviation $\sigma=0.1$ to the input state vector data. The reason this is done is to ensure the model is robustly trained. We run a set of experiments to test the robustness to this 'noisy' input. 
\begin{figure}[h!]
\centering
\captionsetup{justification=centering}
\begin{subfigure}[b]{0.42\textwidth}
\includegraphics[width=\textwidth]{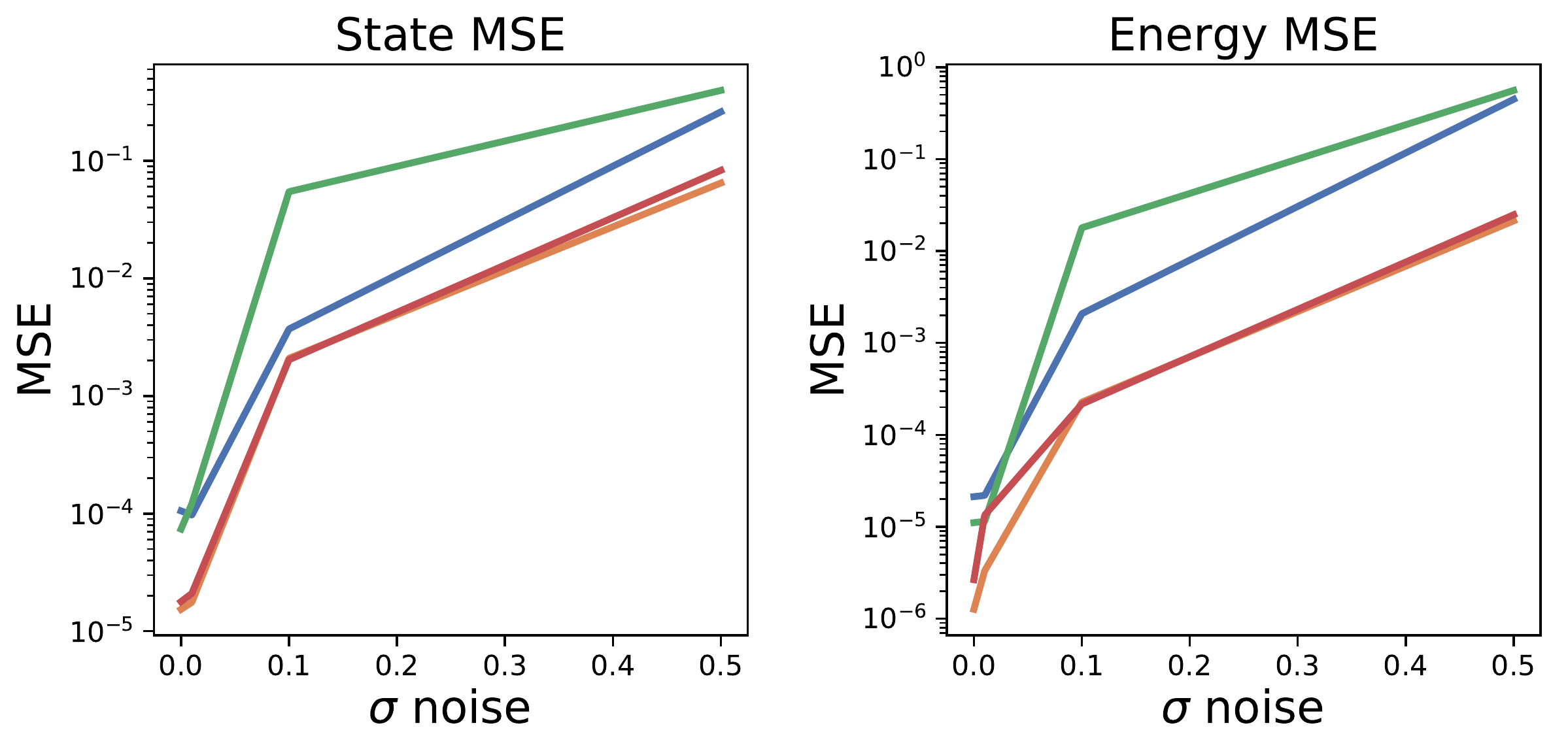}
\caption{mass spring}
\end{subfigure}
\begin{subfigure}[b]{0.42\textwidth}
\includegraphics[width=\textwidth]{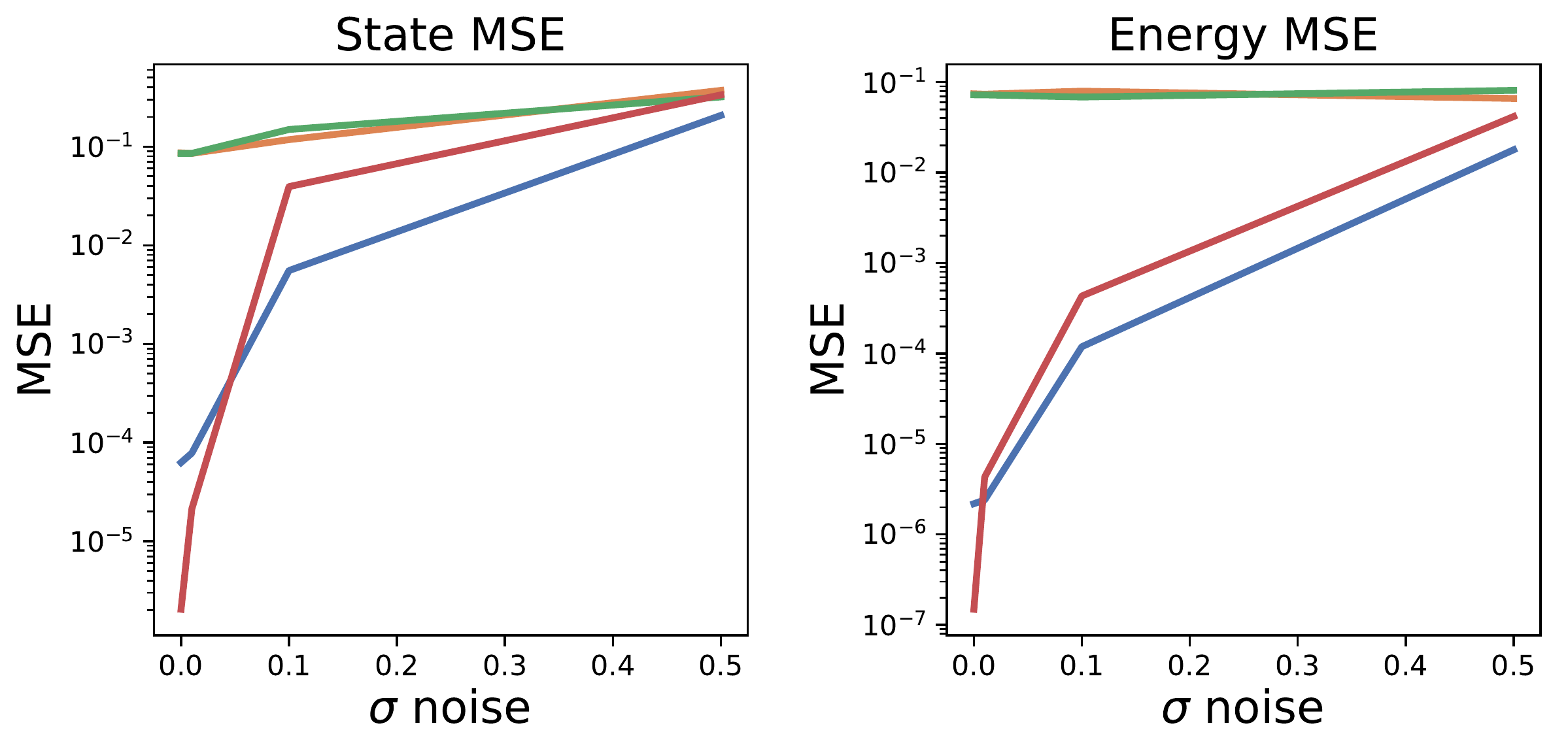}
\caption{damped}
\end{subfigure}
\begin{subfigure}[b]{0.42\textwidth}
\includegraphics[width=\textwidth]{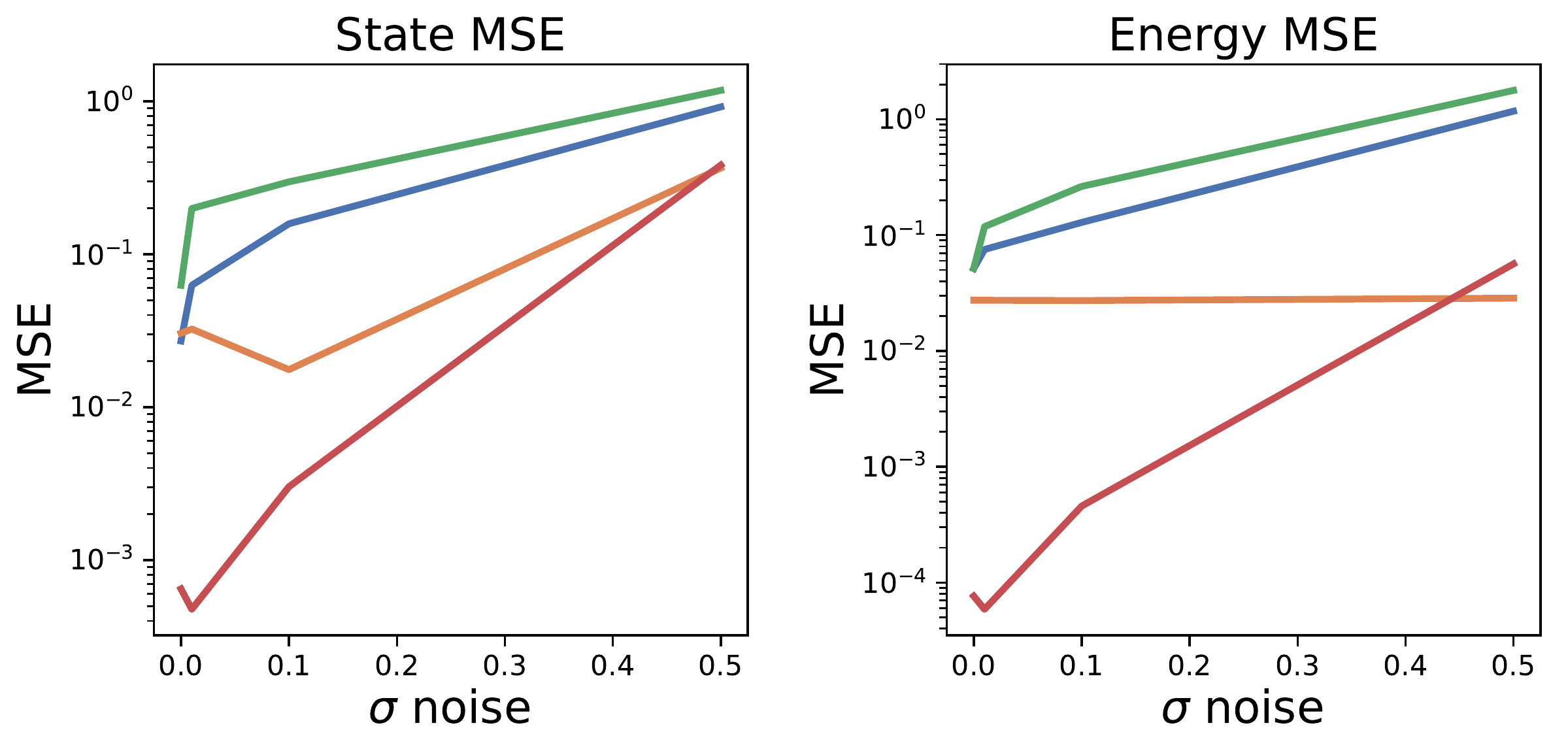}
\caption{forced mass spring (I)}
\end{subfigure}
\begin{subfigure}[b]{0.42\textwidth}
\includegraphics[width=\textwidth]{figures/forced_mass_spring_noise_scaling.pdf}
\caption{forced mass spring (II)}
\end{subfigure}
\begin{subfigure}[b]{0.42\textwidth}
\includegraphics[width=\textwidth]{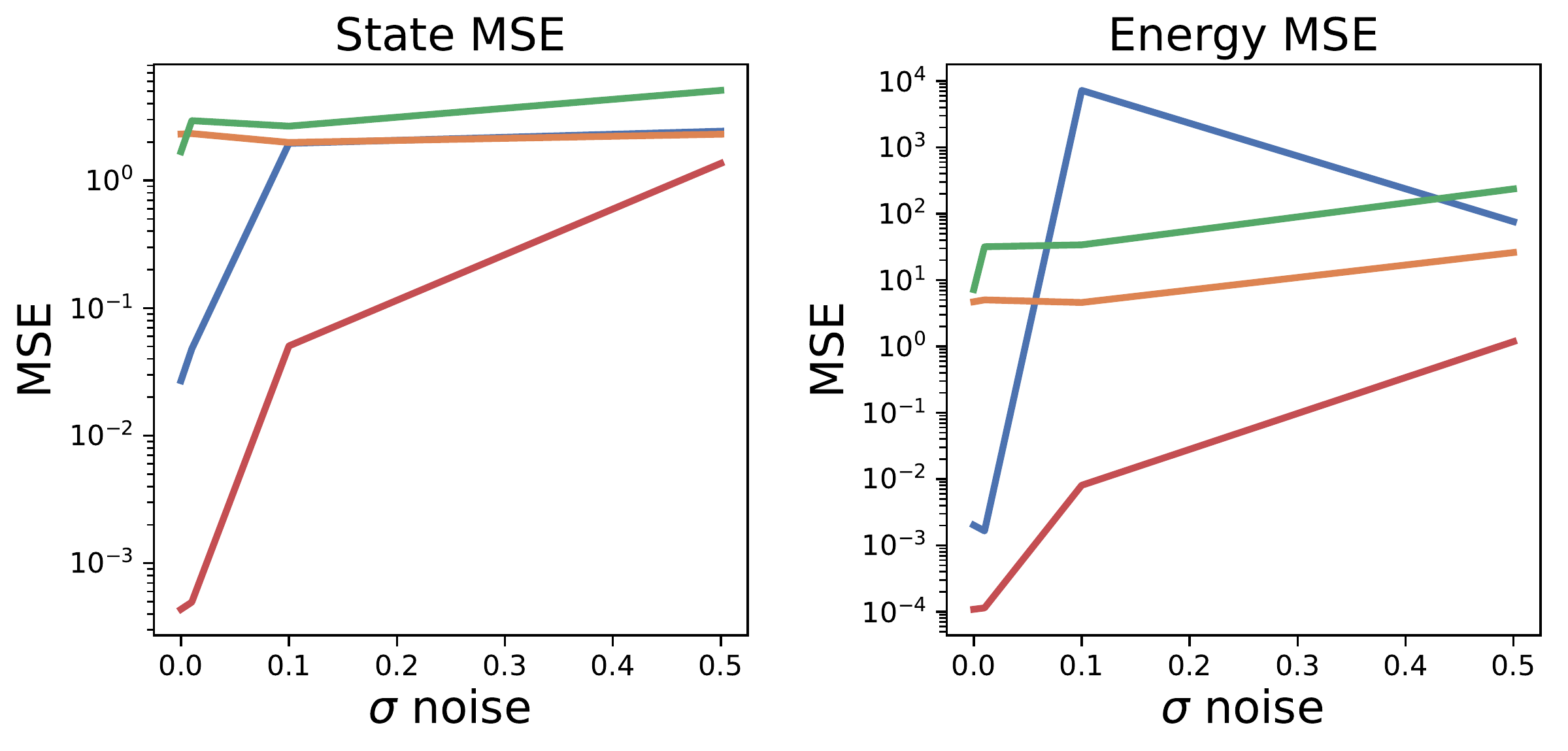}
\caption{Duffing}
\end{subfigure}
\begin{subfigure}[b]{0.42\textwidth}
\includegraphics[width=\textwidth]{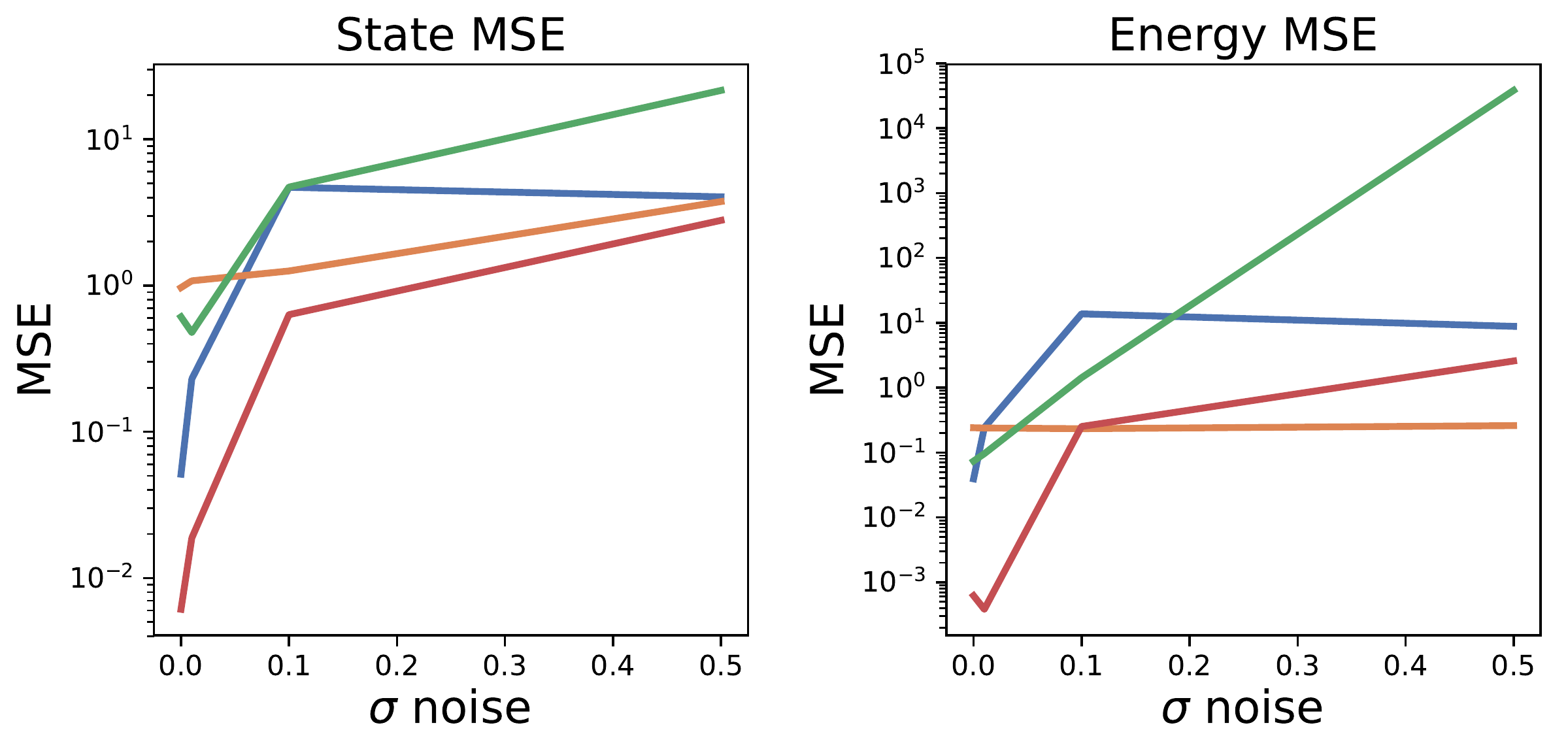}
\caption{relativistic Duffing}
\end{subfigure}
\caption{Results across systems when a gaussian noise with $\sigma$ is added to the position and momentum data}
\end{figure}

\section{2 body coupled spring system}

We run an additional study of a 2-body system with masses $m$ coupled by springs with constants $k$. The system is on a horizontal plane and consists of two masses, each attached to fixed walls on the left and right respectively and coupled to each other by another spring. A driving force of $\cos(t)$ is applied to the first mass. The Hamiltonian for this system is:

$$
\mathcal{H} = \frac{p_1^2}{2m} + \frac{p_2^2}{2m} + kq_1^2 + kq_2^2 - kq_1q_2 -q_1\cos(t) $$

The initial conditions are sampled such that $\mathbf{q} \in [-0.5,0.5]^2$ and $\mathbf{p} \in [-0.2,0.2]^2$. $\Delta t = 0.1$ and $T_{\max} = 5$. Only 25 initial conditions are used to train the system.

\begin{figure}
    \centering
    \includegraphics[width=0.4\textwidth]{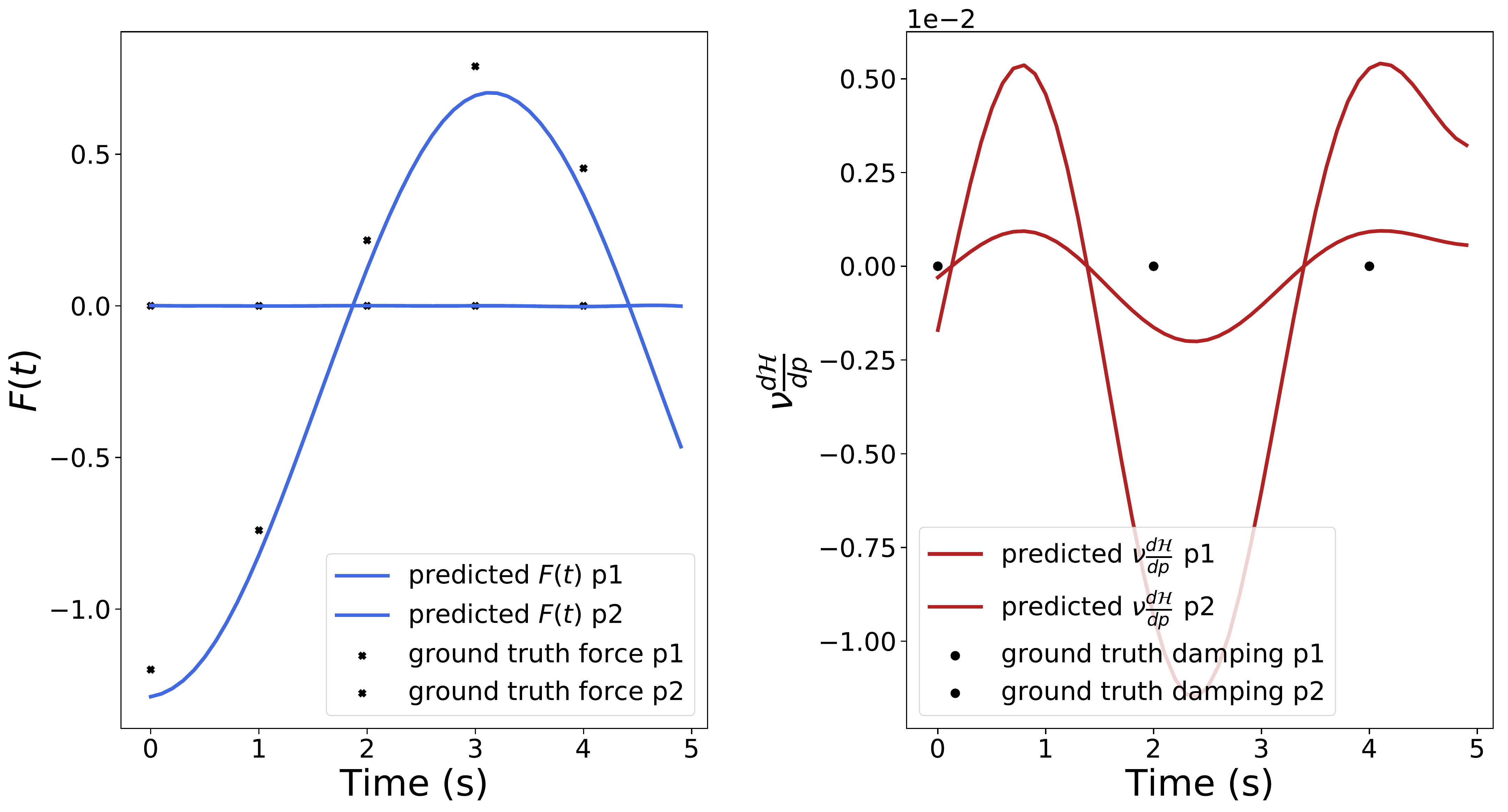}
    \includegraphics[width=0.4\textwidth]{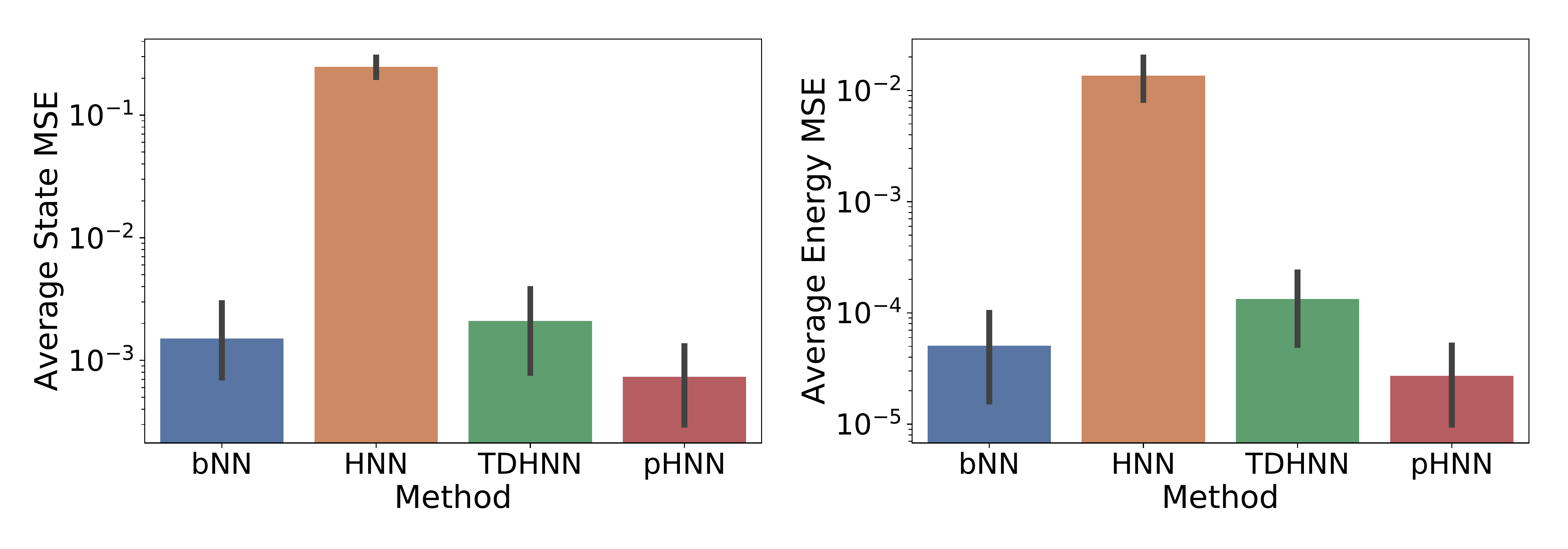}
    \caption{spring coupled results: pHNN performs the best with bNN performing similarly. Results are reported for an embedded integrator.}
    \label{fig:springc}
\end{figure}